\documentclass[%
 reprint,
 amsmath,amssymb,
prb,
]{revtex4-2}

\usepackage{lipsum}
\usepackage{bm}
\usepackage{amssymb}
\usepackage{amstext}
\usepackage{amsmath}
\usepackage{mathrsfs}
\usepackage{graphicx}
\usepackage{color}
\usepackage{amsthm}
\usepackage{floatrow}
\usepackage{array}

\usepackage{listings}
\usepackage{booktabs}
\usepackage{hyperref}

\begin{document}
\preprint{APS/123-QED}
\title{Statistical Physics of Deep Neural Networks: Initialization toward Optimal Channels}
\thanks{Correspondence should be addressed to P.S. and Y.T.}%

  \author{Kangyu Weng}
\email{wengky20@mails.tsinghua.edu.cn}
 \altaffiliation[]{Tsien Excellence in Engineering
Program, Xingjian College, Tsinghua University, Beijing, 100084, China.}
 \author{Aohua Cheng}
\email{aohuacheng18@gmail.com}
 \altaffiliation[]{Tsien Excellence in Engineering
Program, Xingjian College, Tsinghua University, Beijing, 100084, China.}
  \author{Ziyang Zhang}
\email{zhangziyang11@huawei.com}
 \altaffiliation[]{Laboratory of Advanced Computing and Storage, Central Research Institute, 2012 Laboratories, Huawei Technologies Co. Ltd., Beijing, 100084, China.}
\author{Pei Sun}%
 \email{peisun@tsinghua.edu.cn}
 \altaffiliation[]{Department of Psychology \& Tsinghua Brain and Intelligence Lab, Tsinghua University, Beijing, 100084, China.}
 \author{Yang Tian}
\email{tiany20@mails.tsinghua.edu.cn}
 \altaffiliation[]{Department of Psychology \& Tsinghua Laboratory of Brain and Intelligence, Tsinghua University, Beijing, 100084, China.}
 \altaffiliation[Also at]{Laboratory of Advanced Computing and Storage, Central Research Institute, 2012 Laboratories, Huawei Technologies Co. Ltd., Beijing, 100084, China.}



\begin{abstract}
In deep learning, neural networks serve as noisy channels between input data and its representation. This perspective naturally relates deep learning with the pursuit of constructing channels with optimal performance in information transmission and representation. While considerable efforts are concentrated on realizing optimal channel properties during network optimization, we study a frequently overlooked possibility that neural networks can be initialized toward optimal channels. Our theory, consistent with experimental validation, identifies primary mechanics underlying this unknown possibility and suggests intrinsic connections between statistical physics and deep learning. Unlike the conventional theories that characterize neural networks applying the classic mean-filed approximation, we offer analytic proof that this extensively applied simplification scheme is not valid in studying neural networks as information channels. To fill this gap, we develop a corrected mean-field framework applicable for characterizing the limiting behaviors of information propagation in neural networks without strong assumptions on inputs. Based on it, we propose an analytic theory to prove that mutual information maximization is realized between inputs and propagated signals when neural networks are initialized at dynamic isometry, a case where information transmits via norm-preserving mappings. These theoretical predictions are validated by experiments on real neural networks, suggesting the robustness of our theory against finite-size effects. Finally, we analyze our findings with information bottleneck theory to confirm the precise relations among dynamic isometry, mutual information maximization, and optimal channel properties in deep learning. Our work may lay a cornerstone for promoting deep learning in terms of network initialization and suggest general statistical physics mechanisms underlying diverse deep learning techniques.     

\end{abstract}

\maketitle
\section{Introduction}
\subsection{Neural networks are information channels} In deep learning, neural networks attempt to identify an optimal latent representation (e.g., a low-dimensional feature space) of the data such that subsequent learning tasks can be solved more efficiently \cite{bengio2013representation}. Below, we not only review latest advances in deep learning but also suggest a general perspective to unfiy them.

Let us consider a sample set $\mathbf{X}$ and an associated learning target set $\mathbf{Y}=\gamma\left(\mathbf{X}\right)$ (e.g., labels), where $\gamma$ is a mapping defined by the learning task. A neural network parameterized by $\phi$ is expected to optimize a representation $\phi\left(\mathbf{X}\right)$ with $\operatorname{dim}\left(\phi\left(\mathbf{X}\right)\right)<\operatorname{dim}\left(\mathbf{X}\right)$ (here $\operatorname{dim}\left(\cdot\right)$ measures the dimensionality) such that an ideal mapping $\gamma_{\phi}:\phi\left(\mathbf{X}\right)\rightarrow\mathbf{Y}$ can be readily learned to solve the task. This objective requires an appropriate evaluation of the optimality of neural network representation $\phi\left(\mathbf{X}\right)$.

\begin{figure*}[!t]
\includegraphics[width=1\columnwidth]{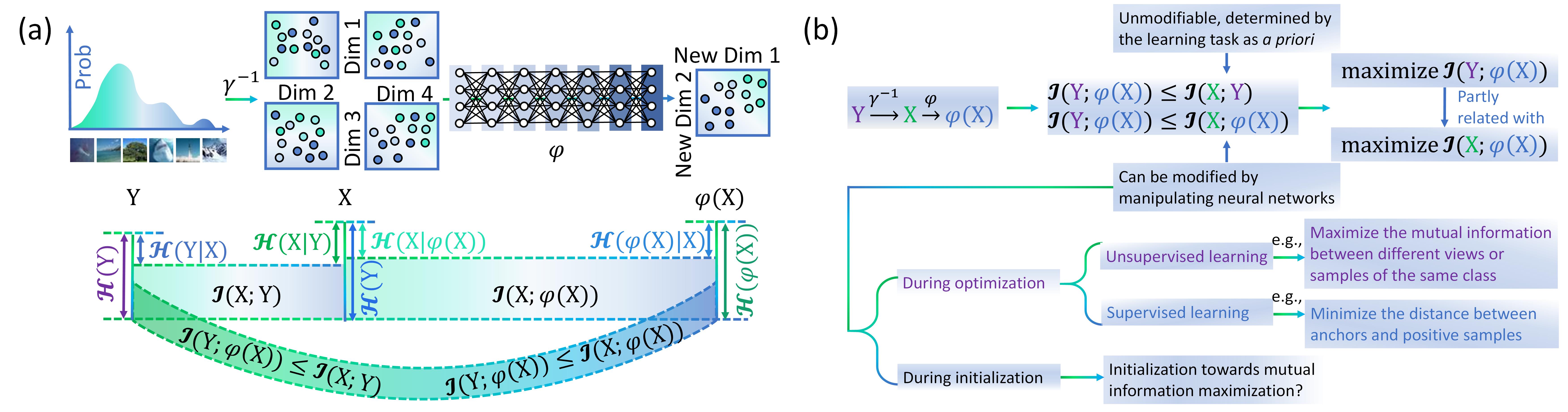}
\caption{Conceptual illustrations of research objectives. (a) The learning target $\mathbf{Y}$ (e.g., a distribution of different classes of objects) is represented by a sample set $\mathbf{X}$, where different classes of samples (denoted by nodes with different colors) are promiscuously distributed in the original sample space. A neural network is expected to learn an appropriate neural representation $\phi\left(\cdot\right)$ such that all classes are distributed following clear patterns in $\phi\left(\mathbf{X}\right)$ to optimally capture the information of $\mathbf{Y}$. During this process, a joint channel consisting of two sub-channels is defined between $\mathbf{Y}$ and $\phi\left(\mathbf{X}\right)$. The fist sub-channel, bridging between $\mathbf{Y}$ and $\mathbf{X}$, determines how two Shannon entropies, $\mathcal{H}\left(\mathbf{Y}\right)$ and $\mathcal{H}\left(\mathbf{X}\right)$, share a common part of information measured by $\mathcal{I}\left(\mathbf{X};\mathbf{Y}\right)$ given the information loss measured by conditional entropies, $\mathcal{H}\left(\mathbf{Y}\mid\mathbf{X}\right)$ and $\mathcal{H}\left(\mathbf{X}\mid\mathbf{Y}\right)$. The second sub-channel, bridging between $\mathbf{Y}$ and $\phi\left(\mathbf{X}\right)$, is optimizable for neural networks. These sub-channels jointly define the upper-bound of $\mathcal{I}\left(\phi\left(\mathbf{X}\right);\mathbf{Y}\right)$ shown in Eq. (\ref{EQ2}). (b) Because $\mathcal{I}\left(\phi\left(\mathbf{X}\right);\mathbf{Y}\right)$ is unmodifiable for neural networks, deep learning studies primarily explore maximizing $\mathcal{I}\left(\phi\left(\mathbf{X}\right);\mathbf{X}\right)$. This objective is partly related with, not directly equivalent to, maximizing $\mathcal{I}\left(\phi\left(\mathbf{X}\right);\mathbf{Y}\right)$. While previous works mainly focus on maximizing $\mathcal{I}\left(\phi\left(\mathbf{X}\right);\mathbf{X}\right)$ during optimization (e.g., training), we shall suggest a possibility to realize this objective during neural network initialization by proposing a statistical physics theory of neural networks.} 
\end{figure*}

Although the optimality of $\phi\left(\mathbf{X}\right)$ can be evaluated by diverse metrics according to concrete task demands (e.g., see instances in reinforcement \cite{lesort2018state}, graph \cite{hamilton2017representation}, and causal \cite{scholkopf2021toward} representation learning frameworks), a mainstream idea is to consider the neural network as a noisy channel between $\mathbf{X}$ and its representation $\phi\left(\mathbf{X}\right)$. This perspective naturally leads to the consideration of two cases:
\begin{itemize}
    \item[\textbf{(1)} ] In unsupervised learning, the information of learning target $\mathbf{Y}$ is not known to the neural network \cite{dike2018unsupervised}. To make the learning task resolvable, mapping $\gamma$ is assumed as a bijective function from sample $\mathbf{X}$ to target $\mathbf{Y}$ such that the neural network can learn the distribution of $\mathbf{Y}$ by representing the distribution of $\mathbf{X}$ (e.g., the target distribution is exactly an optimal representation of sample distribution in clustering tasks \cite{xu2005survey,xu2015comprehensive} and unsupervised representation learning \cite{radford2015unsupervised,arora2019theoretical}). If samples and targets have distinct or irrelevant distributions, the unsupervised learning of $\mathbf{Y}$ based on $\mathbf{X}$ is ill-posed. 
    \item[\textbf{(2)} ] In supervised learning, the information of learning target $\mathbf{Y}$ is known to the neural network as supervision \cite{kotsiantis2007supervised}. In an ideal situation where $\gamma$ is a well-defined bijective mapping from $\mathbf{X}$ to $\mathbf{Y}$ (i.e., samples and targets are perfectly paired), learning the distribution of $\mathbf{Y}$ is principally equivalent to learning the distribution of $\mathbf{X}$ (e.g., consider the separable case where $\mathbf{X}$ can be subdivided into disjoint convex sets in Euclidean space according to the label information in $\mathbf{Y}$ \cite{hastie2009elements,deisenroth2020mathematics}). In more realistic situations where samples and targets are not perfectly paired, learning the distribution of $\mathbf{Y}$ is non-trivial and not necessarily consistent with representing the distribution of $\mathbf{X}$ (e.g., consider the case where noisy labels exit \cite{song2022learning,frenay2013classification}).
\end{itemize}

\subsection{How can neural networks become optimal channels} 
How can neural networks become optimal channels favorable for deep learning tasks? This is the central question concerned in our research. 

Mathematically, learning tasks are unified by a Markov chain of data processing in information theory \cite{cover1999elements} (see \textbf{Fig. 1} for illustration)
\begin{align}
\mathbf{Y}\xrightarrow{\gamma^{-1}}\mathbf{X}\xrightarrow{\phi}\phi\left(\mathbf{X}\right).\label{EQ1}
\end{align}
Note that the above Markov chain is different from $\mathbf{X}\xrightarrow{\phi}\phi\left(\mathbf{X}\right)\xrightarrow{\gamma_{\phi}}\mathbf{Y}$, the Markov chain of hidden variable models \cite{shamir2010learning}. This is because the joint distribution between samples and targets is given as \emph{a priori} rather than something adjustable in deep learning (e.g., neural networks can not modify task designs). The Markov property in Eq. (\ref{EQ1}) implies zero conditional mutual information values $\mathcal{I}\left(\mathbf{Y};\phi\left(\mathbf{X}\right)\vert\mathbf{X}\right)=\mathcal{I}\left(\phi\left(\mathbf{X}\right);\mathbf{Y}\vert\mathbf{X}\right)=0$ \cite{cover1999elements}, leading to a generalized version of the data processing inequality \cite{cover1999elements,kang2010new,zhou2021strong}
\begin{align}
\mathcal{I}\left(\phi\left(\mathbf{X}\right);\mathbf{Y}\right)\leq\min\left\{\mathcal{I}\left(\mathbf{X};\mathbf{Y}\right),\mathcal{I}\left(\phi\left(\mathbf{X}\right);\mathbf{X}\right)\right\},\label{EQ2}
\end{align}
which can be readily derived from $\mathcal{I}\left(\phi\left(\mathbf{X}\right);\mathbf{Y}\right)+\mathcal{I}\left(\mathbf{Y};\mathbf{X}\vert\phi\left(\mathbf{X}\right)\right)=\mathcal{I}\left(\mathbf{X};\mathbf{Y}\right)$ and $\mathcal{I}\left(\phi\left(\mathbf{X}\right);\mathbf{Y}\right)+\mathcal{I}\left(\phi\left(\mathbf{X}\right);\mathbf{Y}\vert\mathbf{X}\right)=\mathcal{I}\left(\mathbf{X};\phi\left(\mathbf{X}\right)\right)$ because conditional mutual information is non-negative.

Eq. (\ref{EQ2}) suggests a clear direction to evaluate the optimality of neural network representation $\phi\left(\mathbf{X}\right)$. Because the mutual information between samples and targets, denoted by $\mathcal{I}\left(\mathbf{X};\mathbf{Y}\right)$, is unoptimizable for the neural network, the optimality of $\phi\left(\mathbf{X}\right)$ is, at least partly, determined by maximizing $\mathcal{I}\left(\phi\left(\mathbf{X}\right);\mathbf{X}\right)$, the mutual information between samples and their represented counterparts. 

\subsection{Previous studies on neural networks as optimal channels}

Given the possible direction for neural networks to become optimal channels, we review previous efforts that devote to realize this condition during optimization (e.g., training).

In unsupervised learning, mutual information maximization has been demonstrated as a promising approach in recent works \cite{oord2018representation,henaff2020data,tian2020contrastive,hjelm2018learning}. Although this approach arises from the idea of maximizing $\mathcal{I}\left(\phi\left(\mathbf{X}\right);\mathbf{X}\right)$ in Eq. (\ref{EQ2}), it differs from the direct maximization because estimating the mutual information between the entire sample set and its neural network representation is statistically deficient \cite{poole2019variational} (i.e., distribution-free estimators (e.g., statistical bounds) of entropy-related quantities have high sample complexity \cite{mcallester2020formal}). In general, what mutual information maximization approach follows is a kind of multi-view formulation \cite{tschannen2019mutual}. Given each canonical sample $X$ of $\mathbf{X}$ (e.g., an image) and its two different and potentially overlapping observations (e.g., two views of the image), $X^{i}$ and $X^{j}$, neural networks are optimized for $\max_{\phi}\mathcal{I}\left(\phi\left(X^{i}\right);\phi\left(X^{j}\right)\right)$. Such an idea can date back to Refs. \cite{becker1992self,linsker1988self} and is valid for optimizing neural networks because $\mathcal{I}\left(\phi\left(X^{i}\right);\phi\left(X^{j}\right)\right)\leq\mathcal{I}\left(X;\phi\left(X^{i}\right),\phi\left(X^{j}\right)\right)$ \cite{tschannen2019mutual}, which can be generally used as a lower bound of the mutual information maximization objective $\mathcal{I}\left(\phi\left(\mathbf{X}\right);\mathbf{X}\right)$ \cite{linsker1988self,tschannen2019mutual}.

In supervised learning, although mutual information is not explicitly used as an optimization objective, it has been discovered as intrinsically related to deep metric learning \cite{ge2018deep,tschannen2019mutual}. Let us consider $\left(X,Y,Z\right)$, a triplet where $X$ is an anchor, $Y$ is a positive sample (e.g., belongs to the same class as $X$), and $Z$ is a negative sample (e.g., belongs to a different class compared with $X$). In deep metric learning, neural networks are optimized to learn a representation $\phi$ such that $d\left(\phi\left(X\right),\phi\left(Y\right)\right)<d\left(\phi\left(X\right),\phi\left(Z\right)\right)$ for any $\left(X,Y,Z\right)$, where $d\left(\cdot,\cdot\right)$ denotes a distance measure \cite{ge2018deep,ge2018deep,yu2019deep}. Meanwhile, InfoNCE, an extensively applied lower bound of mutual information \cite{oord2018representation}, can be derived to support maximizing $\mathcal{I}\left(\phi\left(\mathbf{X}\right);\mathbf{X}\right)$ in Eq. (\ref{EQ2}) if all negative samples are drawn from the true marginal distribution \cite{poole2019variational}. As proven by Ref. \cite{tschannen2019mutual}, InfoNCE can be equivalently reformulated as an expectation of the multi-class $n$-pair loss \cite{sohn2016improved}, which is a standard triplet loss in deep metric learning. In the case where negative samples are not independently drawn, InfoNCE is not valid in estimating mutual information \cite{tschannen2019mutual}.

In sum, existing deep learning approaches, irrespective of being unsupervised or supervised, primarily focus on driving neural networks toward optimal channels during optimization. Maximizing $\mathcal{I}\left(\phi\left(\mathbf{X}\right);\mathbf{X}\right)$ is used as, or at least coincides with, a part of the optimization objectives of mainstream frameworks.

\begin{figure*}[!t]
\includegraphics[width=1\columnwidth]{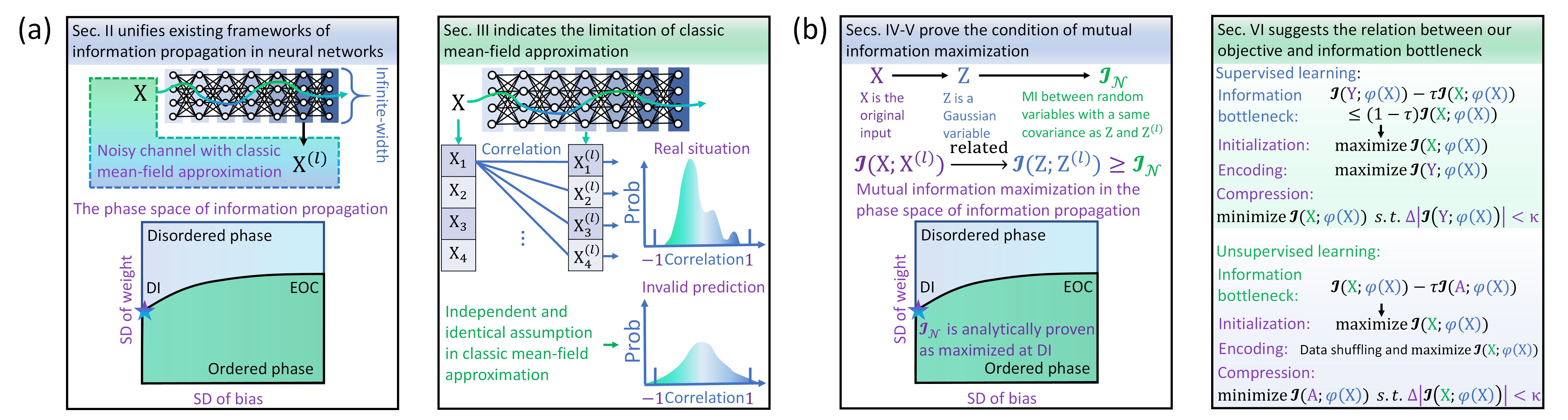}
\caption{Summary of our main framework and contributions. (a) In Sec. \ref{Sec2}, we review and unify classic mean-field theories of neural networks, measure signal moments and correlation, construct the phase space of information propagation, and define edge of chaos (EOC) as well as dynamic isometry (DI). Based on the unified framework, we indicate the limitation of class mean-field approximation in characterizing neural networks as information channels in Sec. \ref{Sec3}. This limitation arises from the independent and identical assumption held by class mean-field approximation, which enables the correlation between propagated signal and its original form to be larger than $1$ or smaller than $-1$. (b) In Secs. \ref{Sec4}-\ref{Sec5}, we overcome the limitation of existing framework by proposing a corrected mean-field approximation theory of neural networks. We relate our analysis with Gaussian information bottleneck, based on which, we can derive an important lower bound of mutual information that is analytically proven as maximized at dynamic isometry. Apart from analytic proofs, our theory is also computationally validated on real neural networks. In Sec. \ref{Sec6}, we suggest the relation between our objective of maximizing $\mathcal{I}\left(\phi\left(\mathbf{X}\right);\mathbf{X}\right)$ and driving neural networks toward optimal channels from the perspective of information bottleneck theory, revealing the insights of our theory on deep learning.} 
\end{figure*}

\subsection{Open questions on neural networks as optimal channels}
Given our review and unified formalization presented above, one may erroneously expect that the properties of neural networks as optimal channels have been completely confirmed. This is generally not true because the studies on neural networks as optimal channels, just like any other deep learning research, still remain at their very beginnings. Below, we summarize three critical open questions in this direction:
\begin{itemize}
    \item[\textbf{(I)} ] Is it possible to maximize $\mathcal{I}\left(\phi\left(\mathbf{X}\right);\mathbf{X}\right)$ during neural network such that the subsequent training process can be improved or at least be accelerated?
    \item[\textbf{(II)} ] If the properties of neural networks as optimal channels are determined by the dynamics of information propagation within neural networks? Is it possible to bridge between the static framework of Shannon theory \cite{cover1999elements} and the dynamic characterization of information propagation via existing statistical physics theories of neural networks (e.g., mean-field approximation \cite{li2021statistical,bahri2020statistical} and neural tangent kernel \cite{jacot2018neural,bahri2020statistical,golikov2022neural} theories)? If it is not possible, what are the main limitations of existing theories and how to resolve them?
    \item[\textbf{(III)} ] What is the precise relation between maximizing $\mathcal{I}\left(\phi\left(\mathbf{X}\right);\mathbf{X}\right)$ during initialization and driving neural networks toward optimal channels? Can we further relate this relation with other deep learning theories (e.g., information bottleneck \cite{tishby2000information,alemi2016deep,higgins2017betavae}) to explore a unified view?
\end{itemize}

 \subsection{Our framework and contribution}
Motivated by these open questions, we attempt to develop general theories to verify the possibility of initializing neural networks at optimal states for maximizing $\mathcal{I}\left(\phi\left(\mathbf{X}\right);\mathbf{X}\right)$, based on which, we aim at exploring the underlying connections between information theory and statistical physics in deep learning. Specifically, our framework and contributions are summarized as the following.

   In Sec. \ref{Sec2}, we review and unify existing works about characterizing neural networks as information channels and analyzing information propagation within them (see \textbf{Fig. 2a}). To make analytic derivations possible, our work primarily focuses on infinite-width neural networks, whose formal definitions are presented in Sec. \ref{Sec2-1}. Infinite-width neural networks, irrespective of being optimized in a Bayesian manner \cite{lee2017deep,matthews2018gaussian,garriga2018deep} or by gradient descent approaches \cite{jacot2018neural,lee2019wide,chizat2019lazy}, are favorable in analytic formulations because they become Gaussian processes defined with specific kernels (e.g., neural tangent kernel \cite{jacot2018neural,bahri2020statistical,golikov2022neural}) at the limits. This property holds across different network architectures (e.g., fully-connected layers \cite{lee2017deep}, convolutional layers \cite{garriga2018deep,novak2018bayesian}, residual connections \cite{garriga2018deep}, and recurrent networks \cite{yang2019wide}), enabling our theory to analyze mainstream deep learning models. Given an infinite-width neural network, we unify previous studies to formalize how information propagates within the network and analytically measure diverse properties of propagated signals (e.g., the second moment) on each layer in Secs. \ref{Sec2-1}-\ref{Sec2-2}. This formalization relates our analysis with the studies on edge of chaos and dynamic isometry in deep neural networks \cite{sirignano2022mean,schoenholz2016deep,pennington2017resurrecting,yang2017mean,pennington2018emergence,chen2018dynamical,xiao2018dynamical}, whose formal definitions are presented in Sec. \ref{Sec2-3}. In sum, our unified framework lays the foundation of our subsequent analysis. 

   Given the fundamental definitions in Sec. \ref{Sec2}, we do not limit ourselves to adapting classic theories completely. On the contrary, we show that the independent and identical assumption held by the classic mean-field approximation of infinite-width neural networks is invalid in analyzing neural networks as information channels in Sec. \ref{Sec3} (see \textbf{Fig. 2a}). Specifically, the independent and identical assumption is proven to imply an Gaussian distribution of the correlation between propagated signal and its original form in inputs in Sec. \ref{Sec3-1}. In other words, the classic mean-field approximation creates a non-zero probability for the correlation to be larger than $1$ or smaller than $-1$, which contradicts the definition of correlation (i.e., a correlation must belongs to the interval of $\left[-1,1\right]$). Although a correlation quantity can be empirically treated as a constant and may still be valid (e.g., when the constant is located within $\left[-1,1\right]$) if it follows a Gaussian distribution with an infinitesimal variance approaching to 0 or a strictly zero variance, we prove that the correlation measured under the classic mean-field framework dissatisfies these two conditions in Sec. \ref{Sec3-2}. Therefore, the classic mean-field approximation is invalid in correlation measurement even from an empirical perspective.

  To resolve the limitation of classic mean-field approximation suggested in Sec. \ref{Sec3}, we develop a new mean-field-like theory with corrected independent and identical assumption in Sec. \ref{Sec4-1} (see \textbf{Fig. 2b}). Different from the classic one, our corrected mean-field approximation does not implies a Gaussian distribution of correlation with non-zero constant variance because it excludes the independent and identical assumption on input $\mathbf{X}$. Certainly, the loose constraints held by the corrected mean-field approximation propose challenges to analytic characterization
of information channel properties. To overcome these challenges, we introduce Gaussian information
bottleneck into our analysis to relate optimizing $\mathcal{I}\left(\phi\left(\mathbf{X}\right);\mathbf{X}\right)$ with maximizing a certain lower bound of mutual information in Sec. \ref{Sec4-2}. In the phase space of information propagation, the lower bound is analytically proven as maximized at dynamic isometry point, a case where each layer serves as a random mapping with norm-preserving property during information transmission, in Secs. \ref{Sec4-2}-Sec. \ref{Sec4-3}. In other situations where dynamic isometry is absent, neural networks become highly noisy channels with high information dissipation rates. In Sec. \ref{Sec5}, our theory is computationally validated on real neural networks. Although our theory is initially developed on infinite-width neural networks, we demonstrate its general applicability on real deep learning models by showing consistency between our theoretical predictions and empirical observations on finite-width neural networks with diverse settings (e.g., different layer widths, layer quantities, and activation functions).

   In Sec. \ref{Sec5}, we relate our theories with information bottleneck theory \cite{tishby2000information,alemi2016deep,higgins2017betavae}, a special case of rate distortion theory \cite{berger2003rate} and sufficient statistics theory \cite{kleven2021sufficient}, to present a unified discussion on the role of maximizing $\mathcal{I}\left(\phi\left(\mathbf{X}\right);\mathbf{X}\right)$ in driving neural networks toward optimal channels in deep learning (see \textbf{Fig. 2b}). We show that supervised learning and unsupervised learning share similar optimization objectives in terms of information bottleneck. When neural networks are trained with random data shuffling tricks \cite{nguyen2022globally,summers2021nondeterminism}, we theoretically suggests the possibility that maximizing $\mathcal{I}\left(\phi\left(\mathbf{X}\right);\mathbf{X}\right)$ in Eq. (\ref{EQ2}) serves as a conditional mechanism for neural networks to become optimal in both supervised learning and unsupervised learning. Our discussions suggest the potential insights of our work on deep learning and statistical physics.

 \section{Infinite-width neural networks as information channels}\label{Sec2}
In this section, we introduce a framework to characterize infinite-width neural networks as information channels by summarizing and reformulating existing studies on mean-field behaviours and dynamic isometry of neural networks \cite{mei2019mean,nguyen2019mean,poole2016exponential,pennington2017resurrecting,schoenholz2016deep,pennington2018emergence,chen2018dynamical,xiao2018dynamical}.
 
 \subsection{Mean-field behaviours of infinite-width neural networks}\label{Sec2-1}
 Let us consider an arbitrary deep neural network with multiple layers. The dynamics of cross-layer information propagation (i.e., a signal propagates from the $l$-th layer to the $\left(l+1\right)$-th layer) is characterized as
 \begin{align}
\mathbf{X}^{\left(l+1\right)}=\mathbf{W}^{\left(l+1\right)}\psi\left(\mathbf{X}^{\left(l\right)}\right)+\varepsilon^{\left(l+1\right)}, \label{EQ3}
 \end{align}
 where $\mathbf{X}^{\left(l\right)}=\left(\mathbf{X}_{1}^{\left(l\right)},\ldots,\mathbf{X}_{N_{l}}^{\left(l\right)}\right)\in\mathbb{R}^{N_{l}}$ denotes the vector of pre-activation signals in the $l$-th layer, parameter $N_{l}\in\mathbb{N}^{+}$ denotes the width of the $l$-th layer, mapping $\psi\left(\cdot\right)$ denotes a non-linear activation function, matrix $\mathbf{W}^{\left(l+1\right)}\in\mathbb{R}^{N_{l+1}}\times\mathbb{R}^{N_{l}}$ defines the weights of all connections between the $l$-th layer and the $\left(l+1\right)$-th layer, and $\varepsilon^{\left(l+1\right)}=\left(\varepsilon^{\left(l+1\right)}_{1},\ldots,\varepsilon_{N_{l+1}}^{\left(l+1\right)}\right)\in\mathbb{R}^{N_{l+1}}$ denotes the associated residuals. In common cases, each residual $\varepsilon^{\left(l+1\right)}_{i}$ is frequently assumed as a Gaussian variable \cite{mei2019mean,nguyen2019mean}. Please see \textbf{Fig. 3a} for illustrations.

 To offer a clear vision, we begin with formalizing the classic mean-field approximation of the above neural network before we analyze its limitations in Sec. \ref{Sec3-1}. Under the independent and identical assumption of $\mathbf{W}^{\left(l+1\right)}$ and $\mathbf{X}^{\left(l\right)}$ (e.g., each element $\mathbf{W}^{\left(l+1\right)}_{i,j}$ in $\mathbf{W}^{\left(l+1\right)}$ is independently and identically distributed), we can derive 
 \begin{align}
\mathbf{X}^{\left(l+1\right)}_{i}=\Big\langle \mathbf{W}^{\left(l+1\right)}_{i},\psi\left(\mathbf{X}^{\left(l\right)}\right)\Big\rangle+\varepsilon^{\left(l+1\right)}_{i}\xrightarrow{d}\mathcal{N}\left(\mu_{i},\sigma_{i}^{2}\right) \label{EQ4}
 \end{align}
 as $N_{l+1}\rightarrow\infty$, where $\mathbf{X}^{\left(l+1\right)}_{i}$ and $\mathbf{W}^{\left(l+1\right)}_{i}$ respectively denote the $i$-th rows of $\mathbf{X}^{\left(l+1\right)}$ and $\mathbf{W}^{\left(l+1\right)}$ for any $i \in \{1,\ldots,N_{l+1}\}$, notion $\langle\cdot,\cdot\rangle$ defines the inner product, and $\mathcal{N}\left(\mu_{i},\sigma_{i}\right)$ defines a specific Gaussian variable.
 
 In general, Eq. (\ref{EQ4}) approximates each signal propagating in the infinite-width neural network as a certain Gaussian variable under the central limit theorem, enabling us to study an ensemble of random neural networks associated with the original neural network \cite{poole2016exponential} (see \textbf{Fig. 3a}). This idea has been demonstrated as effective in characterizing the mean-field behaviours of two-layer \cite{mei2019mean} and multi-layer \cite{nguyen2019mean} neural networks.
 
 \subsection{Information propagation in infinite-width neural networks}\label{Sec2-2}
 As we have explained, our motivation to consider infinite-width neural networks is to analytically study information propagation dynamics within them. Because the signal propagating across layers in an infinite-width neural network has become a Gaussian variable, we can capture most of its dynamics by studying the first two moments of it. In our research, we primarily focus on the second moment since it has been demonstrated as relevant with the expressivity of neural networks (i.e., the second moment of a signal coincides with the length of its internal Riemannian manifold in downstream layers) \cite{poole2016exponential}, enabling us to analyze the order-to-chaos expressivity phase transition \cite{poole2016exponential}.

\begin{figure*}[!t]
\includegraphics[width=1\columnwidth]{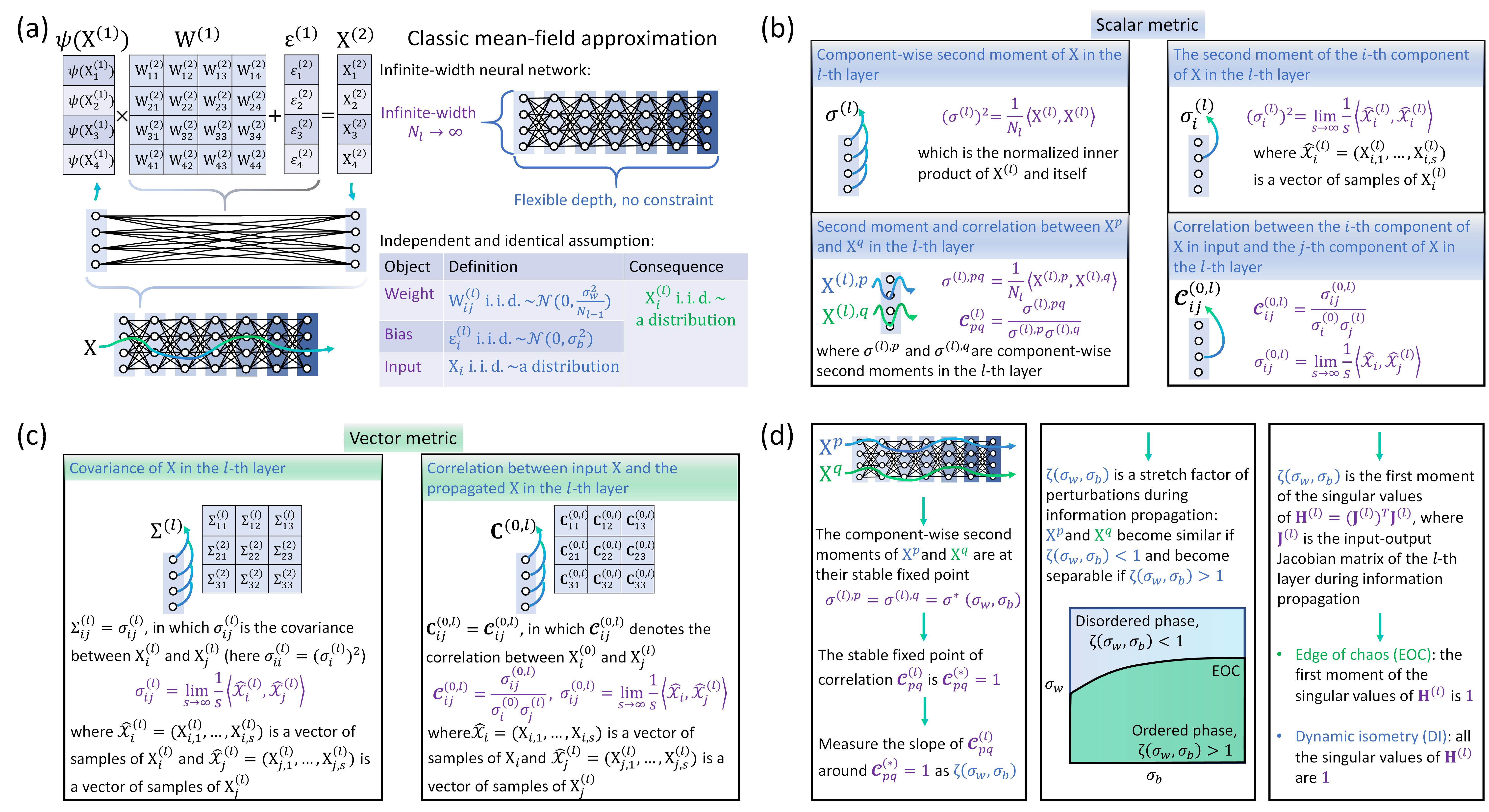}
\caption{Conceptual illustrations of the classic mean-field approximation of neural networks. (a) Key settings of classic mean-field approximation, such as infinite-width condition as well as independent and identical assumption, are summarized. (b-c) Illustrations of the scalar and vector metrics used in characterizing information propagation processes are presented. (d) Main steps of the derivation processes of edge of chaos (EOC) and dynamic isometry (ID) in the phase space of information propagation are shown.} 
\end{figure*}

For convenience, we consider the case studied by Ref. \cite{poole2016exponential}, where each $\mathbf{W}^{\left(l\right)}_{ij}\;\text{i.i.d.}\sim\mathcal{N}\left(0,\frac{\sigma_{w}^{2}}{N_{l-1}}\right)$ and each $\varepsilon^{\left(l\right)}_{i}\;\text{i.i.d.}\sim\mathcal{N}\left(0,\sigma_{b}^{2}\right)$ (note that ``i.i.d." stands for being independently and identically distributed). In the infinite-width limit (i.e., $N_{l}\rightarrow\infty$), the second moment (i.e., variance) of signals in the $l$-layer, denoted by $\left(\sigma^{\left(l\right)}\right)^{2}$, can be calculated as
 \begin{align}
    \left(\sigma^{\left(l\right)}\right)^{2}&=\frac{1}{N_{l}}\sum_{i=1}^{N_{l}}\left(\mathbf{X}^{\left(l\right)}_{i}\right)^{2},\label{EQ5}\\&=\mathbb{E}\left(\Big\langle \mathbf{W}^{\left(l\right)}_{i},\psi\left(\mathbf{X}^{\left(l-1\right)}\right)\Big\rangle+\varepsilon^{\left(l\right)}_{i}\right)^{2},\label{EQ6}
\end{align}
where $\mathbb{E}\left(\cdot\right)$ denotes the expectation (see \textbf{Fig. 3b}). Note that the second moment in Eq. (\ref{EQ5}) reduces to the second origin moment because $\mathbb{E}\left(\mathbf{X}^{\left(l\right)}_{i}\right)=0$ holds for each $\mathbf{X}^{\left(l\right)}_{i}$. Eq. (\ref{EQ6}) can be further reformulated as 
 \begin{align}
    \left(\sigma^{\left(l\right)}\right)^{2}&=\mathbb{E}\left[\sum_{j=1}^{N_{l-1}}\mathbf{W}^{\left(l\right)}_{ij}\psi\left(\mathbf{X}^{\left(l-1\right)}_{j}\right)+\varepsilon^{\left(l\right)}_{i}\right]^{2},\label{EQ7}\\&=\sum_{j=1}^{N_{l-1}}\mathbb{E}\left[\left(\mathbf{W}^{\left(l\right)}_{ij}\right)^2\right]\mathbb{E}\left[\psi\left(\mathbf{X}^{\left(l-1\right)}_{j}\right)^{2}\right]+\mathbb{E}\left[\left(\varepsilon^{\left(l\right)}_{i}\right)^{2}\right],\label{EQ8}\\
    &=\sigma_{w}^{2}\int_{\mathbb{R}}\psi\left(\sqrt{\sigma^{\left(l-1\right)}}x\right)^{2}\mathsf{D}x+\sigma_{b}^{2},\label{EQ9}
\end{align}
where Eq. (\ref{EQ8}) is derived from the fact that $\mathbb{E}\left(\mathbf{W}^{\left(l\right)}_{ij}\mathbf{W}^{\left(l\right)}_{ik}\right)=\mathbb{E}\left(\mathbf{W}^{\left(l\right)}_{ij}\right)\mathbb{E}\left(\mathbf{W}^{\left(l\right)}_{ik}\right)$ for any $j\neq k$ under the independent and identical assumption and Eq. (\ref{EQ9}) is derived using $\mathbb{E}\left[\sum_{j=1}^{N_{l-1}}\left(\mathbf{W}^{\left(l\right)}_{ij}\right)^2\right]=\sigma_{w}^{2}$ (i.e., weights are independently and identically distributed). In Eq. (\ref{EQ9}), notion $\mathsf{D}x=\frac{\mathsf{d}x}{\sqrt{2\pi}}\exp\left(-\frac{x^{2}}{2}\right)$ is a standard Gaussian measure, where $x\in\mathbb{R}$. In general, Eq. (\ref{EQ9}) defines an iterative dynamic process of the second moment of signals in each layer \cite{poole2016exponential}, whose initial condition is 
\begin{align}
    \left(\sigma^{\left(1\right)}\right)^{2}=\frac{\sigma_{w}^{2}}{N_{1}}\langle\mathbf{X},\mathbf{X}\rangle+\sigma_{b}^{2},\label{EQ10}
\end{align}
where $\mathbf{X}$ is the input vector to the neural network, which is assumed to obey the independent and identical assumption, i.e., each $\mathbf{X}_{i}\;\text{i.i.d.}\sim\text{a certain distribution}$, to ensure that $\mathbf{X}_{i}^{\left(l\right)}\;\text{i.i.d.}\sim\text{a certain distribution}$ in Eq. (\ref{EQ4}). One can see Ref. \cite{poole2016exponential} for more analyses of Eq. (\ref{EQ9}).

Apart from the second moment of a single signal, we can also consider the relation between two propagating signals. For two input signals, $\mathbf{X}^{p}$ and $\mathbf{X}^{q}$, correlated with each other because they propagate within the same neural network, we can mark their representations in the $l$-layer as $\mathbf{X}^{\left(l\right),p}$ and $\mathbf{X}^{\left(l\right),q}$ to measure their second moment as (see \textbf{Fig. 3b} for illustration)
\begin{widetext}
 \begin{align}
    \sigma^{\left(l\right),pq}=&\frac{1}{N_{l}}\sum_{i=1}^{N_{l}}\mathbf{X}^{\left(l\right),p}_{i}\mathbf{X}^{\left(l\right),q}_{i},\label{EQ11}\\=&\mathbb{E}\Bigg[\left(\sum_{j=1}^{N_{l-1}}\mathbf{W}^{\left(l\right)}_{ij}\psi\left(\mathbf{X}^{\left(l-1\right),p}_{j}\right)+\varepsilon^{\left(l\right)}_{i}\right)\left(\sum_{j=1}^{N_{l-1}}\mathbf{W}^{\left(l\right)}_{ij}\psi\left(\mathbf{X}^{\left(l-1\right),q}_{j}\right)+\varepsilon^{\left(l\right)}_{i}\right)\Bigg],\label{EQ12}\\=&\sum_{j=1}^{N_{l-1}}\mathbb{E}\left[\left(\mathbf{W}^{\left(l\right)}_{ij}\right)^2\right]\mathbb{E}\left[\psi\left(\mathbf{X}^{\left(l-1\right),p}_{j}\right)\psi\left(\mathbf{X}^{\left(l-1\right),q}_{j}\right)\right]+\mathbb{E}\left[\left(\varepsilon^{\left(l\right)}_{i}\right)^{2}\right],\label{EQ13}\\
    =&\sigma_{w}^{2}\int_{\mathbb{R}}\int_{\mathbb{R}}\psi\left(\sigma^{\left(l-1\right),p}x\right)\psi\left[\sigma^{\left(l-1\right),q}\left(\mathcal{C}_{pq}^{\left(l-1\right)}x+\sqrt{1-\left(\mathcal{C}_{pq}^{\left(l-1\right)}\right)^{2}}y\right)\right]\mathsf{D}x\mathsf{D}y+\sigma_{b}^{2},\label{EQ14}
\end{align}
\end{widetext}
where $\sigma^{\left(l-1\right),p}$ and $\sigma^{\left(l-1\right),q}$ denotes the standard deviations of $\mathbf{X}^{p}$ and $\mathbf{X}^{q}$ in the $\left(l-1\right)$-layer defined following Eq. (\ref{EQ9}). Variables $x$ and $y$ are independent standard Gaussian variables. Notion $\mathcal{C}_{pq}^{\left(l-1\right)}$ denotes the correlation between $\mathbf{X}^{p}$ and $\mathbf{X}^{q}$ in the $\left(l-1\right)$-layer. Please note that subscript $i$ in Eqs. (\ref{EQ11}-\ref{EQ13}) can be eventually dropped in Eq. (\ref{EQ14}) because there is independent and identical assumption on the components of $\mathbf{X}^{p}$ and $\mathbf{X}^{q}$. 

 \subsection{Phase space of information propagation}\label{Sec2-3}
 Let us contextualize the above mathematical definitions with physics backgrounds. While studying information propagation, we expect to understand how the global extrinsic curvature of latent Riemannian geometry in inputs (i.e., the relation between two input signals $\mathbf{X}^{p}$ and $\mathbf{X}^{q}$), a key factor underlying the expressivity of neural networks \cite{poole2016exponential}, evolves across layers. 

 We wonder if the difference between $\mathbf{X}^{p}$ and $\mathbf{X}^{q}$ will be principally maintained, enlarged, or reduced during information propagation. To answer this question, we first explore the stable fixed point of $\sigma^{\left(l\right)}$ in Eq. (\ref{EQ9}) as a function of $\left(\sigma_{w},\sigma_{b}\right)\in\mathbb{R}\times\left(0,\infty\right)$ because the length of each propagating signal in the downstream layer will rapidly converges to this stable fixed point. After confirming the stable fixed point, denoted by $\sigma^{*}\left(\sigma_{w},\sigma_{b}\right)$, we set $\sigma^{\left(l\right),p}=\sigma^{\left(l\right),q}=\sigma^{*}\left(\sigma_{w},\sigma_{b}\right)$ in Eq. (\ref{EQ14}) and divide Eq. (\ref{EQ14}) by $\sigma^{*}\left(\sigma_{w},\sigma_{b}\right)$ to obtain the iterative dynamics of the correlation between $\mathbf{X}^{p}$ and $\mathbf{X}^{q}$ (see \textbf{Fig. 3b})
 \begin{widetext}
 \begin{align}
   \mathcal{C}_{pq}^{\left(l\right)}=&\frac{1}{\sigma^{*}\left(\sigma_{w},\sigma_{b}\right)^{2}}\left\{\sigma_{w}^{2}\int_{\mathbb{R}} \int_{\mathbb{R}}\psi\left(\sigma^{*}\left(\sigma_{w},\sigma_{b}\right)x\right)\psi\left[\sigma^{*}\left(\sigma_{w},\sigma_{b}\right)\left(\mathcal{C}_{pq}^{\left(l-1\right)}x+\sqrt{1-\left(\mathcal{C}_{pq}^{\left(l-1\right)}\right)^{2}}y\right)\right]\mathsf{D}x\mathsf{D}y+\sigma_{b}^{2}\right\},\label{EQ15}
\end{align}
\end{widetext}
whose fixed point can be readily found as $\mathcal{C}_{pq}^{*}=1$ after direct calculation. The stability of this fixed point, however, can not be directly confirmed. Therefore, we need to analyze $\zeta\left(\sigma_{w},\sigma_{b}\right)$, the slope of $\mathcal{C}_{pq}^{\left(l\right)}$ around $\mathcal{C}_{pq}^{*}=1$ given a setting $\left(\sigma_{w},\sigma_{b}\right)$ (see \textbf{Fig. 3d})
 \begin{align}
   \zeta\left(\sigma_{w},\sigma_{b}\right)=&\frac{\partial \mathcal{C}_{pq}^{\left(l\right)}}{\partial \mathcal{C}_{pq}^{\left(l-1\right)}}\Bigg\vert_{\mathcal{C}_{pq}^{\left(l\right)}=1},\label{EQ16}\\
   =&\sigma_{w}^{2}\int_{\mathbb{R}}\psi^{\prime}\left(\sigma^{*}\left(\sigma_{w},\sigma_{b}\right)x\right)^{2}\mathsf{D}x,\label{EQ17}\\
   =&\frac{1}{N_{l}}\mathbb{E}\left\{\operatorname{tr}\left[\left(\mathbf{D}^{\left(l\right)}\mathbf{W}^{\left(l\right)}\right)^{T}\mathbf{D}^{\left(l\right)}\mathbf{W}^{\left(l\right)}\right]\right\}.\label{EQ18}
\end{align}
In Eq. (\ref{EQ17}), notion $\psi^{\prime}\left(\cdot\right)$ denotes the derivative function of $\psi\left(\cdot\right)$. In Eq. (\ref{EQ18}), notion $\mathbf{D}^{\left(l\right)}$ is a diagonal matrix $\mathbf{D}^{\left(l\right)}=\operatorname{diag}\left(\left[\psi^{\prime}\left(\mathbf{X}^{\left(l\right)}_{1}\right),\ldots,\psi^{\prime}\left(\mathbf{X}^{\left(l\right)}_{N_{l}}\right)\right]\right)$ such that $\mathbf{J}^{\left(l\right)}=\mathbf{D}^{\left(l\right)}\mathbf{W}^{\left(l\right)}$ can be understood as the input-output Jacobian matrix of the $l$-th layer \cite{pennington2017resurrecting,schoenholz2016deep}. Notion $\operatorname{tr}\left(\cdot\right)$ denotes matrix trace. The expectation is calculated by averaging across all possible configurations of $\mathbf{D}^{\left(l\right)}\mathbf{W}^{\left(l\right)}$ in Eq. (\ref{EQ18}). As suggested by Eq. (\ref{EQ18}), parameter $\zeta\left(\sigma_{w},\sigma_{b}\right)$ can be understood as a stretch factor because any random perturbation $\eta$ in the $\left(l-1\right)$-th layer, $\mathbf{X}^{\left(l\right)}+\nu$, implies a subsequent perturbation in the $l$-th layer, $\mathbf{X}^{\left(l+1\right)}+\mathbf{J}^{\left(l-1\right)}\nu$, with a stretch effect measured by $\zeta\left(\sigma_{w},\sigma_{b}\right)=\mathbb{E}\left(\Vert\mathbf{J}^{\left(l-1\right)}\nu\Vert_{2}^{2}/\Vert\nu\Vert_{2}^{2}\right)$ \cite{poole2016exponential}. The effect corresponds to growth if $\zeta\left(\sigma_{w},\sigma_{b}\right)>1$ and corresponds to shrinkage if $\zeta\left(\sigma_{w},\sigma_{b}\right)<1$ (see \textbf{Fig. 3d}).

The fixed point $\mathcal{C}_{pq}^{*}=1$ is stable when $\zeta\left(\sigma_{w},\sigma_{b}\right)<1$ while it is unstable when $\zeta\left(\sigma_{w},\sigma_{b}\right)>1$ \cite{poole2016exponential,pennington2017resurrecting,schoenholz2016deep}. In the case where $\mathcal{C}_{pq}^{*}=1$ is stable, all possible relations between $\mathbf{X}^{p}$ and $\mathbf{X}^{q}$ eventually converge to a strong correlation (i.e., $\mathbf{X}^{\left(l\right),p}$ and $\mathbf{X}^{\left(l\right),q}$ become increasingly similar as $l$ increases). In the case where $\mathcal{C}_{pq}^{*}=1$ is not stable, $\mathbf{X}^{p}$ and $\mathbf{X}^{q}$ become increasingly separable as they propagate. Consequently, the condition with $\zeta\left(\sigma_{w},\sigma_{b}\right)=1$ naturally defines a boundary separating between two phases on the plane of $\left(\sigma_{w},\sigma_{b}\right)$. The first phase, corresponding to the case where signals become separable during information propagation (i.e., $\zeta\left(\sigma_{w},\sigma_{b}\right)>1$), is referred to as the disordered phase. The second phase, corresponding to the case where signals converge to correlated states (i.e., $\zeta\left(\sigma_{w},\sigma_{b}\right)<1$), is the ordered phase. Please see \textbf{Fig. 3d} for illustrations of the phase space. As suggested by Refs. \cite{schoenholz2016deep,pennington2018emergence}, the ordered and disordered phases correspond to vanishing and exploding gradient problems, respectively. 

Apart from defining a phase transition boundary between ordered and disordered phases, we can further consider the dynamic isometry condition, a special point on this boundary (see \textbf{Fig. 3d}). Specifically, we can understand $\zeta\left(\sigma_{w},\sigma_{b}\right)$ in Eq. (\ref{EQ18}) as the second moment of the singular values of $\mathbf{J}^{\left(l\right)}=\mathbf{D}^{\left(l\right)}\mathbf{W}^{\left(l\right)}$ or, equivalently, the first moment of the singular values of $\mathbf{H}^{\left(l\right)}=\left(\mathbf{J}^{\left(l\right)}\right)^{T}\mathbf{J}^{\left(l\right)}$
\begin{align}
    \zeta\left(\sigma_{w},\sigma_{b}\right)=\frac{1}{N_{l}}\sum_{i=1}^{N_{l}}\theta_{i}^{2}=\frac{1}{N_{l}}\sum_{i=1}^{N_{l}}\lambda_{i},\label{EQ19}
\end{align}
where $\left[\theta_{1},\ldots,\theta_{N_{l}}\right]$ and $\left[\lambda_{1},\ldots,\lambda_{N_{l}}\right]$ are the singular values of $\mathbf{J}^{\left(l\right)}$ and $\mathbf{H}^{\left(l\right)}$, respectively. Dynamic isometry is defined as a case where the singular values of $\mathbf{H}^{\left(l\right)}$ not only have a first moment of $1$ but also satisfy $\lambda_{i}=1$ for each $i\in\{1,\ldots,N_{l}\}$. To reach the dynamic isometry condition, we can consider a situation where the second moment of the singular values of $\mathbf{H}^{\left(l\right)}$ approaches to $0$ while the first moment equals $1$. As suggested by Ref. \cite{pennington2018emergence}, free probability theory \cite{mingo2017free} can be applied to derive the probability distribution of the singular values of $\mathbf{H}^{\left(l\right)}$ to realize this objective. Detailed calculations of dynamic isometry point $\left(\sigma_{w}^{\diamond},\sigma_{b}^{\diamond}\right)$ across different activation functions $\psi\left(\cdot\right)$ or network architectures have been provided by Refs. \cite{pennington2018emergence,chen2018dynamical,xiao2018dynamical,pennington2017resurrecting} and we summarize the general method in Appendix \ref{ASec1}. Based on the method, it has been demonstrated that orthogonal weights, i.e., $\left(\mathbf{W}^{\left(l\right)}\right)^{T}\mathbf{W}^{\left(l\right)}=\mathbf{I}$ for each $l$ (notion $\mathbf{I}$ denotes the identity matrix), and a non-ReLU-type activation function $\psi\left(\cdot\right)$, i.e., $\psi\left(r\right)\neq\max\left(0,r\right)$ for each $r\in\mathbb{R}$, can achieve dynamical isometry in neural networks \cite{pennington2017resurrecting,pennington2018emergence}.

In sum, we have reviewed and unified existing studies on mean-field behaviours and dynamic isometry of neural networks \cite{mei2019mean,nguyen2019mean,poole2016exponential,pennington2017resurrecting,schoenholz2016deep,pennington2018emergence,chen2018dynamical,xiao2018dynamical} to present a general framework for analyzing information propagation in infinite-width neural networks. This framework supports us to rethink the limitation of existing theories and explore undiscovered laws governing neural networks.

 \section{On the limitation of classic mean-field approximation}\label{Sec3}
 In this section, we suggest the limitation of classic mean-field approximation in characterizing neural networks as information channels.
 
   \subsection{Rethinking the classic mean-field approximation: a strict perspective}\label{Sec3-1}
   Let us rethink the validity of the classic mean-field approximation summarized in Sec. \ref{Sec2} in defining the channel capacity of neural networks. At the first glance, the rethinking seems to be unnecessary because Sec. \ref{Sec2} has suggested how neural networks serve as channels where information propagates across layers in a mean-field manner. However, as we suggested below, the independent and identical assumption in classic mean-field theory may imply unexpected errors in correlation measurements.

   \begin{figure*}[!t]
\includegraphics[width=1\columnwidth]{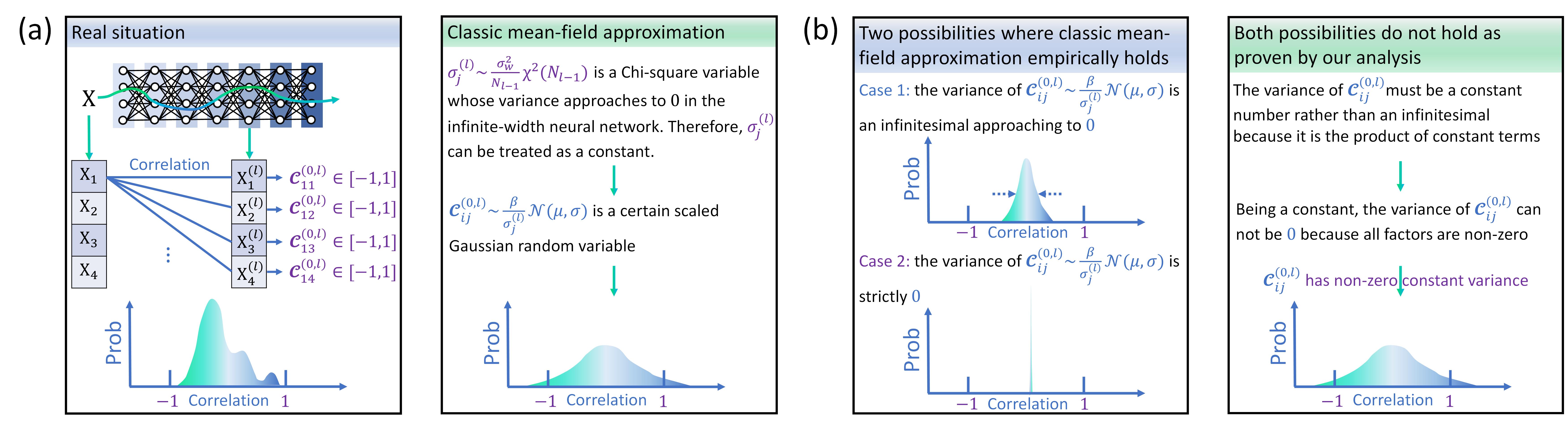}
\caption{Conceptual illustrations of the limitation of classic mean-field approximation in correlation measurement. (a) In real cases, the $i$-th component of an input signal, $\mathbf{X}_{i}$, and the $j$-th component of propagated signal in the $l$-th layer, $\mathbf{X}^{\left(l\right)}_{j}$ should always be located within $\left[-1,1\right]$. However, the independent and identical assumption held by classic mean-field assumption implies a non-zero probability for the measured correlation, $\mathcal{C}_{ij}^{\left(0,l\right)}$, to be larger than $1$ or small than $-1$ because $\mathcal{C}_{ij}^{\left(0,l\right)}$ follows a specific Gaussian distribution. (b) Although a Gaussian variable with an infinitesimal variance approaching to $0$ or a strictly zero variance can be empirically treated as a constant and may be valid to serve as a correlation (e.g., when the constant is located within $\left[-1,1\right]$), the measured correlation $\mathcal{C}_{ij}^{\left(0,l\right)}$ is proven to dissatisfy both cases and has a constant non-zero variance. Consequently, it is invalid even from an empirical perspective.} 
\end{figure*}

  Correlation $\mathcal{C}$, irrespective of being measured between any pair of variables, should be located within the interval of $\left[-1,1\right]$ (see \textbf{Fig. 4a}). An approximation framework is invalid if it enables a correlation to be larger than $1$ or smaller than $-1$ with a non-zero probability (see \textbf{Fig. 4a}). Different from the correlation between two inputs, $\mathbf{X}^{p}$ and $\mathbf{X}^{q}$, in the $l$-th layer defined by Eq. (\ref{EQ15}), here we consider $\mathcal{C}_{ij}^{\left(0,l\right)}$, the correlation between the $i$-th component of an input signal, $\mathbf{X}_{i}$, and the $j$-th component of propagated signal in the $l$-th layer, $\mathbf{X}^{\left(l\right)}_{j}$, in Eq. (\ref{EQ20}). Please see \textbf{Fig. 3b} for illustration. This correlation reflects how an input evolves during its propagation from the $1$-st layer to the $l$-th layer. Specifically, we have

 \begin{widetext}
  \begin{align}
\mathcal{C}_{ij}^{\left(0,l\right)}&=\frac{\mathbb{E}\left[\left(\mathbf{X}^{\left(l\right)}_{j}-\mathbb{E}\left(\mathbf{X}^{\left(l\right)}_{j}\right)\right)\Big(\mathbf{X}_{i}-\mathbb{E}\left(\mathbf{X}_{i}\right)\Big)\right]}{\sigma^{\left(l\right)}_{j}\sigma^{\left(0\right)}_{i}}, \label{EQ20}\\
&=\frac{1}{\sigma^{\left(l\right)}_{j}\sigma^{\left(0\right)}_{i}}\sum_{k=1}^{N_{l-1}}\left\{\mathbf{W}^{\left(l\right)}_{jk}\int_{\mathbb{R}}\int_{\mathbb{R}}\psi\left(\sigma^{\left(l-1\right)}_{k}x_{k}\right)\sigma^{\left(0\right)}_{i}\left[\mathcal{C}_{ik}^{\left(0,l-1\right)}x_{k}+\sqrt{1-\left(\mathcal{C}_{ik}^{\left(0,l-1\right)}\right)^{2}}y_{k}\right]\mathsf{D}x_{k}\mathsf{D}y_{k}\right\}, \label{EQ21}\\
&=\frac{1}{\sigma^{\left(l\right)}_{j}}\sum_{k=1}^{N_{l-1}}\mathbf{W}^{\left(l\right)}_{jk}\mathcal{C}_{ik}^{\left(0,l-1\right)}\left(\int_{\mathbb{R}}\psi\left(\sigma^{\left(l-1\right)}_{k}z_{k}\right)z_{k}\mathsf{D}z_{k}\right), \label{EQ22}
\end{align}
\end{widetext}
where $\sigma^{\left(0\right)}_{i}$ denotes the second moment of $\mathbf{X}_{i}$, the $i$-th component of input signal $\mathbf{X}$, and $\sigma^{\left(l\right)}_{j}$ denotes the second moment of $\mathbf{X}_{j}^{\left(l\right)}$, the $j$-th component of propagated signal $\mathbf{X}^{\left(l\right)}$ (see \textbf{Fig. 3b} for illustration). In Eq. (\ref{EQ21}), every $\mathsf{D}x_{k}$ and $\mathsf{D}y_{k}$ are standard Gaussian measures. Because these measures are identically and independently distributed, their integration over $\mathbb{R}$ can be uniformly represented by the integration of a standard Gaussian measure, $\mathsf{D}z_{k}$, over $\mathbb{R}$ in Eq. (\ref{EQ22}). 

To verify the possibility for $\mathcal{C}_{ij}^{\left(0,l\right)}$ to be larger than $1$ or smaller than $-1$, we can analyze its support set (i.e., if $\left[-1,1\right]$ is a proper subset of the support of $\mathcal{C}_{ij}^{\left(0,l\right)}$, then $\mathcal{C}_{ij}^{\left(0,l\right)}$ has a non-zero probability to reach an invalid value). As shown below, our analysis is implemented based on two main steps.

First, the second moment term in Eq. (\ref{EQ22}) is formally measured as (see \textbf{Fig. 3b} for illustration)
\begin{align}
\left(\sigma^{\left(l\right)}_{j}\right)^{2}=&\mathbb{E}\left[\left(\sum_{k=1}^{N_{l-1}}\mathbf{W}^{\left(l\right)}_{jk}\psi\left(\mathbf{X}_{k}^{\left(l-1\right)}\right)+\varepsilon_{j}^{\left(l\right)}\right)^{2}\right]\notag\\&-\mathbb{E}\left[\sum_{k=1}^{N_{l-1}}\mathbf{W}^{\left(l\right)}_{jk}\psi\left(\mathbf{X}_{k}^{\left(l-1\right)}\right)+\varepsilon_{j}^{\left(l\right)}\right]^{2}, \label{EQ23}
\end{align}
which can be reformulated as
\begin{align}
\left(\sigma^{\left(l\right)}_{j}\right)^{2}=&\left(\varepsilon_{j}^{\left(l\right)}\right)^{2}+\sum_{k=1}^{N_{l-1}}\left(\mathbf{W}^{\left(l\right)}_{jk}\right)^{2}\mathbb{E}\left[\psi\left(\mathbf{X}_{k}^{\left(l-1\right)}\right)^{2}\right]-\left(\varepsilon_{j}^{\left(l\right)}\right)^{2}, \label{EQ24}\\
=&\sum_{k=1}^{N_{l-1}}\left(\mathbf{W}^{\left(l\right)}_{jk}\right)^{2}\mathbb{E}\left[\psi\left(\mathbf{X}_{k}^{\left(l-1\right)}\right)^{2}\right]. \label{EQ25}
\end{align}
The reformulation holds because of $\mathbb{E}\left[\psi\left(\mathbf{X}_{k}^{\left(l-1\right)}\right)\right]=0$ (hint: all common non-ReLU-type activation functions in deep learning are odd functions while the probability density of $\mathbf{X}_{k}^{\left(l-1\right)}$ is an even function) and $\mathbb{E}\left[\psi\left(\mathbf{X}_{a}^{\left(l-1\right)}\right)\psi\left(\mathbf{X}_{b}^{\left(l-1\right)}\right)\right]=\delta_{a,b}$ where $\delta_{\cdot,\cdot}$ denotes the Kronecker delta function (hint: the independent and identical assumption). It is trivial that Eq. (\ref{EQ25}) is equivalent to 
\begin{align}
\left(\sigma^{\left(l\right)}_{j}\right)^{2}=&\alpha\sum_{k=1}^{N_{l-1}}\left(\mathbf{W}^{\left(l\right)}_{jk}\right)^{2}
\sim \frac{\sigma_{w}^{2}}{N_{l-1}}\chi^{2}\left(N_{l-1}\right),\label{EQ26}
\end{align}
where we define $\alpha=\mathbb{E}\left[\psi\left(\mathbf{X}_{k}^{\left(l-1\right)}\right)^{2}\right]$ for each $k$ under the independent and identical assumption (i.e., $\alpha$ is same for every $k$ in the $\left(l-1\right)$-th layer), notion $\chi^{2}\left(\cdot\right)$ denotes the Chi-square random variable (see \textbf{Fig. 4a}). Eq. (\ref{EQ26}) is derived from the independent and identical assumption that $\mathbf{W}^{\left(l\right)}_{jk}\;\text{i.i.d.}\sim\mathcal{N}\left(0,\frac{\sigma_{w}^{2}}{N_{l-1}}\right)$. In the infinite-width limit, we discover that the variance of such a Chi-square variable vanishes
\begin{align}
\lim_{N_{l-1}\rightarrow\infty} 2 N_{l-1}\frac{\sigma_{w}^{4}}{N_{l-1}}=0.\label{EQ27}
\end{align}
In other words, variable $\sigma^{\left(l\right)}_{j}$ can be reasonably treated as a constant in an infinite-width neural network (\textbf{Fig. 4a}). 

Second, we can define $\beta=\int_{\mathbb{R}}\psi\left(\sigma^{\left(l-1\right)}_{k}z_{k}\right)z_{k}\mathsf{D}z_{k}$ for each $k$ under the independent and identical assumption (i.e., $\beta$ is same for every $k$ in the $\left(l-1\right)$-th layer), which supports us to rewrite Eq. (\ref{EQ22}) as
\begin{align}
\mathcal{C}_{ij}^{\left(0,l\right)}
&=\frac{\beta}{\sigma^{\left(l\right)}_{j}}\sum_{k=1}^{N_{l-1}}\mathbf{W}^{\left(l\right)}_{jk}\mathcal{C}_{ik}^{\left(0,l-1\right)}\xrightarrow{d}\frac{\beta}{\sigma^{\left(l\right)}_{j}}\mathcal{N}\left(\mu,\sigma\right), \label{EQ28}
\end{align}
where the right part is derived based on the central limit theorem (i.e., $\sum_{k=1}^{N_{l-1}}\mathbf{W}^{\left(l\right)}_{jk}\mathcal{C}_{ik}^{\left(0,l-1\right)}\xrightarrow{d}\mathcal{N}\left(\mu,\sigma\right)$). As suggested by Eq. (\ref{EQ28}), $\mathcal{C}_{ij}^{\left(0,l\right)}$ is a Gaussian random variable. Therefore, interval $\left[-1,1\right]$ is always a proper subset of the support of $\mathcal{C}_{ij}^{\left(0,l\right)}$, suggesting the limitation of the independent and identical assumption held by classic mean-field approximation (see \textbf{Fig. 4a} for illustration).

\subsection{Rethinking the classic mean-field approximation: an empirical perspective}\label{Sec3-2}

Certainly, one can still treat the independent and identical assumption as reasonably valid if the variance of $\mathcal{C}_{ij}^{\left(0,l\right)}$ equals $0$ or becomes an infinitesimal approaching to $0$. In these cases, correlation $\mathcal{C}_{ij}^{\left(0,l\right)}$ can be generally analyzed as a constant from an empirical perspective (see \textbf{Fig. 4b}). 

However, as we suggest below, correlation $\mathcal{C}_{ij}^{\left(0,l\right)}$ intrinsically features a non-zero and finite (i.e., not being infinitesimal) variance in deep neural networks. Before our formal explanations, we first note that any input $\mathbf{X}$ (e.g., data) of deep neural networks should have a finite dimension even though we apply classic mean-field approximation to consider infinite-width neural networks. This is because the dimensionality of $\mathbf{X}$ is determined by the learning task itself as \emph{a priori} and should not be tampered. Given this property, let us consider a case where each component of $\mathbf{X}$ is independently and identically distributed (i.e., uncorrelated)
  \begin{align}
\mathcal{C}\left(\mathbf{X}_{i},\mathbf{X}_{j}\right):=\mathcal{C}_{ij}^{\left(0,0\right)}
=\delta_{i,j}. \label{EQ29}
\end{align}
By simple calculation based on Eqs. (\ref{EQ20}-\ref{EQ22}) and Eq. (\ref{EQ29}), we can derive the correlation between the $i$-th component of an input signal, $\mathbf{X}_{i}$, and the $j$-th component of propagated signal in the $1$-th layer
  \begin{align}
\mathcal{C}_{ij}^{\left(0,1\right)}&=\frac{1}{\sigma^{\left(1\right)}_{j}}\mathbf{W}^{\left(1\right)}_{ij}\left(\int_{\mathbb{R}}\psi\left(\sigma_{i}z_{i}\right)z_{i}\mathsf{D}z_{i}\right),\label{EQ30}
\end{align}
where $\sigma^{\left(1\right)}_{j}$ can be further calculated following Eq. (\ref{EQ25})
  \begin{align}
\mathcal{C}_{ij}^{\left(0,1\right)}&=\frac{\mathbf{W}^{\left(1\right)}_{ij}\left(\int_{\mathbb{R}}\psi\left(\sigma_{i}z_{i}\right)z_{i}\mathsf{D}z_{i}\right)}{\sum_{k=1}^{N_{0}}\left(\mathbf{W}^{\left(1\right)}_{jk}\right)^{2}\mathbb{E}\left[\psi\left(\mathbf{X}_{k}\right)^{2}\right]}.\label{EQ31}
\end{align}
Notion $N_{0}$ measures the dimension of input $\mathbf{X}$. Based on Eq. (\ref{EQ31}) and Eqs. (\ref{EQ20}-\ref{EQ22}), it is trivial to derive the following variance items
  \begin{align}
\operatorname{\mathbb{V}}\left(\mathcal{C}_{ij}^{\left(0,1\right)}\right)&=\frac{1}{\mathbb{E}\left[\psi\left(\mathbf{X}_{k}\right)^{2}\right]^{2}}\operatorname{\mathbb{V}}\left(\frac{\mathbf{W}^{\left(1\right)}_{ij}}{\sum_{k=1}^{N_{0}}\left(\mathbf{W}^{\left(1\right)}_{jk}\right)^{2}}\right)\notag\\&\times\left(\int_{\mathbb{R}}\psi\left(\sigma_{i}z_{i}\right)z_{i}\mathsf{D}z_{i}\right)^{2}\label{EQ32}
\end{align}
and 
  \begin{align}
\operatorname{\mathbb{V}}\left(\mathcal{C}_{ij}^{\left(0,l\right)}\right)&=\frac{\sigma_{w}^{2}\left(\int_{\mathbb{R}}\psi\left(\sigma_{i}z_{i}\right)z_{i}\mathsf{D}z_{i}\right)^{2}}{\left(\sigma^{\left(l\right)}_{j}\right)^{2}}\operatorname{\mathbb{V}}\left(\mathcal{C}_{ij}^{\left(0,l-1\right)}\right),\label{EQ33}
\end{align}
where we momentarily use $\operatorname{\mathbb{V}}\left(\cdot\right)$ to denote the variance to avoid confusions on mathematical symbols. 

Let us primarily prove that $\operatorname{\mathbb{V}}\left(\mathcal{C}_{ij}^{\left(0,l\right)}\right)$ in Eq. (\ref{EQ33}) can not be an infinitesimal approaching to $0$ (i.e., $\operatorname{\mathbb{V}}\left(\mathcal{C}_{ij}^{\left(0,l\right)}\right)$ is a finite real number). Because $N_{0}\in\mathbb{N}^{+}$ is finite, we know that each $\mathbf{W}^{\left(1\right)}_{ij}\;\text{i.i.d.}\sim\mathcal{N}\left(0,\frac{\sigma_{w}^{2}}{N_{0}}\right)$ has a finite variance, which further implies that $\operatorname{\mathbb{V}}\left(\frac{\mathbf{W}^{\left(1\right)}_{ij}}{\sum_{k=1}^{N_{0}}\left(\mathbf{W}^{\left(1\right)}_{jk}\right)^{2}}\right)=\mathbb{E}\left(\frac{\left(\mathbf{W}^{\left(1\right)}_{ij}\right)^{2}}{\left(\sum_{k=1}^{N_{0}}\left(\mathbf{W}^{\left(1\right)}_{jk}\right)^{2}\right)^{2}}\right)$ is finite. Because the rest part of terms in Eq. (\ref{EQ32}) equal certain finite real numbers, a finite value of $\operatorname{\mathbb{V}}\left(\frac{\mathbf{W}^{\left(1\right)}_{ij}}{\sum_{k=1}^{N_{0}}\left(\mathbf{W}^{\left(1\right)}_{jk}\right)^{2}}\right)$ makes $\mathcal{C}_{ij}^{\left(0,1\right)}$ intrinsically have a finite variance. According to Eq. (\ref{EQ33}), this property further makes $\operatorname{\mathbb{V}}\left(\mathcal{C}_{ij}^{\left(0,l\right)}\right)$ finite for any $l\in\mathbb{N}^{+}$. Please see \textbf{Fig. 4b} for a summary.

Given that $\operatorname{\mathbb{V}}\left(\mathcal{C}_{ij}^{\left(0,l\right)}\right)$ in Eq. (\ref{EQ33}) is a finite number, we turn to proving $\operatorname{\mathbb{V}}\left(\mathcal{C}_{ij}^{\left(0,l\right)}\right)\neq 0$. The proof can be readily derived from the following facts. First, term $\frac{1}{\mathbb{E}\left[\psi\left(\mathbf{X}_{k}\right)^{2}\right]^{2}}$ in Eq. (\ref{EQ32}) can not be $0$ because the squared output of an appropriate non-linear activation function $\psi\left(\cdot\right)$ can not be infinite (otherwise deep neural networks inevitably involve with numerical issues). Second, term $\int_{\mathbb{R}}\psi\left(\sigma_{i}z_{i}\right)z_{i}\mathsf{D}z_{i}$ can not be $0$ because the integral of the product of $\psi\left(\cdot\right)$ and $z_{i}$, two odd functions, over $\mathbb{R}$ must be non-zero. Third, term $\frac{\mathbf{W}^{\left(1\right)}_{ij}}{\sum_{k=1}^{N_{0}}\left(\mathbf{W}^{\left(1\right)}_{jk}\right)^{2}}$ can not have a zero variance otherwise all weights in the $1$-st layer become the same and immutable. Taken together, we know that $\operatorname{\mathbb{V}}\left(\mathcal{C}_{ij}^{\left(0,1\right)}\right)\neq 0$ always holds in Eq. (\ref{EQ32}). Based on the iterative dynamics defined in Eq. (\ref{EQ33}), it is not difficult to see that $\operatorname{\mathbb{V}}\left(\mathcal{C}_{ij}^{\left(0,l\right)}\right)\neq 0$ holds for any $l\in\mathbb{N}^{+}$ because the coefficient term of $\operatorname{\mathbb{V}}\left(\mathcal{C}_{ij}^{\left(0,l-1\right)}\right)$ can be trivially proven as non-zero following a similar idea. Here we no-longer repeatedly elaborate these details. One can see \textbf{Fig. 4b} for a summary.

In sum, even from an empirical perspective, the independent and identical assumption is invalid in correlation measurement because $\mathcal{C}_{ij}^{\left(0,l\right)}$ is a Gaussian variable with non-zero and finite variance, whose support may include illogical values (i.e., smaller than $-1$ or larger than $1$)

\section{Mutual information maximization at dynamic isometry}\label{Sec4}
In this section, we present our theory on the possibility for neural networks to be initialized toward optimal information channels. To overcome the limitation of classic mean-field approximation, we propose a corrected mean-field approximation framework that does not imply a Gaussian distribution of correlation with non-zero and finite variance. Then, we introduce Gaussian information bottleneck \cite{painsky2017gaussian,chechik2003information} into our analysis, which relates our objective of optimizing $\mathcal{I}\left(\phi\left(\mathbf{X}\right);\mathbf{X}\right)$ with maximizing a lower bound of mutual information and supports analytic derivations. Based on it, we analytically prove that mutual information between input and propagated signals is maximized at dynamic isometry.

\subsection{Corrected mean-field approximation and Gaussian information bottleneck}\label{Sec4-1}
As suggested in Sec. \ref{Sec3-1}, the key limitation of classic mean-field approximation arises from the independent and identical assumption applied on all variables without constraints. Although some variables, such as weight and bias, can be assumed as independently and identically distributed (i.e., each $\mathbf{W}^{\left(l\right)}_{ij}\;\text{i.i.d.}\sim\mathcal{N}\left(0,\frac{\sigma_{w}^{2}}{N_{l-1}}\right)$ and each $\varepsilon^{\left(l\right)}_{i}\;\text{i.i.d.}\sim\mathcal{N}\left(0,\sigma_{b}^{2}\right)$), it is less reasonable to apply independent and identical assumption on the components of input $\mathbf{X}$ (i.e., each $\mathbf{X}_{i}\;\text{i.i.d.}\sim\text{a certain distribution}$). This is because the joint distribution of the components of $\mathbf{X}$ has been defined by the learning task as \emph{a priori} and should not be modified. Meanwhile, this unreasonable independent and identical assumption also makes $\mathbf{X}_{i}^{\left(l\right)}\;\text{i.i.d.}\sim\text{a certain distribution}$ for each $l\in\mathbb{N}^{+}$, which eventually leads to the central limit theorem in Eq. (\ref{EQ28}) and implies a Gaussian distribution of correlation $\mathcal{C}_{ij}^{\left(0,l\right)}$.

Given the above analysis, a natural idea for developing a corrected mean-field approximation is to exclude the independent and identical assumption on the components of $\mathbf{X}$ (see \textbf{Fig. 5a}). In the corrected approximation, there is no constraint on the joint distribution of $\mathbf{X}_{i}$. Therefore, $\mathbf{X}_{i}^{\left(l\right)}$ in the propagated signal may not be independently and identically distributed as well. A direct consequence of this correction lies in that the central limit theorem in Eq. (\ref{EQ28}) no longer holds and the empirical distribution of $\mathcal{C}_{ij}^{\left(0,l\right)}$ does not converges to a Gaussian distribution (see \textbf{Fig. 5a} for illustration). Consequently, the corrected mean-field approximation does not suffer from the limitation of the classic one while characterizing neural networks as information channels. Certainly, this correction also proposes critical challenges to analytic derivations because the distributions of $\mathbf{X}_{i}^{\left(l\right)}$ and $\mathcal{C}_{ij}^{\left(0,l\right)}=\frac{\beta}{\sigma^{\left(l\right)}_{j}}\sum_{k=1}^{N_{l-1}}\mathbf{W}^{\left(l\right)}_{jk}\mathcal{C}_{ik}^{\left(0,l-1\right)}$ lack close-form expressions in most general cases. 

To create a possibility for analytic derivations, we suggest to include Gaussian information bottleneck \cite{painsky2017gaussian,chechik2003information} into our analysis. In general, we can consider a transform that maps $\mathbf{X}$ to an arbitrary Gaussian variable $\mathbf{Z}$ (there is no constraint on the first and second moments of $\mathbf{Z}$). With an appropriate transform, we can principally treat $\mathbf{Z}$ as the ``Gaussian part" of $\mathbf{X}$. As suggested by Ref. \cite{painsky2017gaussian}, optimizing the information bottleneck or mutual information associated with $\mathbf{Z}$ will also reflect the corresponding optimization associated with $\mathbf{X}$ (see \textbf{Fig. 5b}). The benefit of such a connection lies in that Gaussian information bottleneck and Gaussian mutual information have analytic expressions and clear mathematical properties to support our analysis \cite{painsky2017gaussian,chechik2003information}. Following the idea of Gaussian information bottleneck \cite{painsky2017gaussian,chechik2003information}, we suggest to empirically consider a Gaussian counterpart $\mathbf{Z}$ of a given $\mathbf{X}$ where $\mathbf{Z}$ is derived following the approach introduced in Ref. \cite{painsky2017gaussian}. The algorithm proposed by Ref. \cite{painsky2017gaussian} ensures that the derived $\mathbf{Z}$ is a ``Gaussian part" of $\mathbf{X}$ with $\mathcal{I}\left(\phi\left(\mathbf{Z}\right);\mathbf{Z}\right)\leq\mathcal{I}\left(\phi\left(\mathbf{X}\right);\mathbf{X}\right)$. There is no other constraint on $\mathbf{Z}$. Meanwhile, signal $\mathbf{Z}^{\left(l\right)}$ is not required to maintain its Gaussian properties during information propagation in a neural network (i.e., $\phi\left(\mathbf{Z}\right)$, the output of a neural network, can be an arbitrary variable). 

In our work, we do not need to explicitly consider the actual form of $\mathbf{Z}$ derived following Ref. \cite{painsky2017gaussian}. This is because the following inequality holds irrespective of what detailed properties that $\mathbf{Z}$ features as long as $\mathbf{Z}$ is a Gaussian variable (see \textbf{Fig. 5b})
\begin{align}
\mathcal{I}\left( \mathbf{Z};\mathbf{Z}^{\left(l\right)} \right) &\geq
\mathcal{I}_{\mathcal{N}}\left(\mathbf{Z};\mathbf{Z}^{\left(l\right)}\right),\label{EQ34}\\
&=\frac{1}{2}\log\left(\frac{\operatorname{det}\left(\mathbf{\Sigma}^{\left(0\right)}\right)\operatorname{det}\left(\mathbf{\Sigma}^{\left(l\right)}\right)}{\operatorname{det}\left(\mathbf{\Sigma}^{\left(0,l\right)}\right)}\right),\label{EQ35}
\end{align}
where $\Sigma^{\left(0\right)}$ and $\Sigma^{\left(l\right)}$ denote the covariance matrix of $\mathbf{Z}$ and $\mathbf{Z}^{\left(l\right)}$, respectively (i.e., each element in the matrix denotes the covariance between a pair of components of signals). Matrix $\Sigma^{\left(0,l\right)}$ is the covariance matrix measured between the components of $\mathbf{Z}$ and $\mathbf{Z}^{\left(l\right)}$, which is a direct generalization of $\Sigma^{\left(0\right)}$ and $\Sigma^{\left(l\right)}$. Please see \textbf{Fig. 3(c)} for illustrations. Notion $\mathcal{I}_{\mathcal{N}}\left(\mathbf{Z};\mathbf{Z}^{\left(l\right)}\right)$ denotes a special case of mutual information where $\mathbf{Z}$ and $\mathbf{Z}^{\left(l\right)}$ are jointly Gaussian while they maintain the original moment properties (i.e., expectation and covariance remain the same). Please see detailed proofs of Eqs. (\ref{EQ34}-\ref{EQ35}) in Appendix \ref{ASec1}. Meanwhile, one can find an equivalent version of Eqs. (\ref{EQ34}-\ref{EQ35}) in Ref. \cite{ughi2022studies}. 

Let us reversely think about the above process. We can analyze the case where $\mathbf{Z}$ is an arbitrary Gaussian variable to measure mutual information $\mathcal{I}\left(\phi\left(\mathbf{Z}\right);\mathbf{Z}\right)$. There always exists a certain input $\mathbf{X}$ that satisfies $\mathcal{I}\left(\phi\left(\mathbf{Z}\right);\mathbf{Z}\right)\leq\mathcal{I}\left(\phi\left(\mathbf{X}\right);\mathbf{X}\right)$ if we reversely solve the transform problem in Ref. \cite{painsky2017gaussian}. As suggested by Eqs. (\ref{EQ34}-\ref{EQ35}), we can primarily focus on $\mathcal{I}_{\mathcal{N}}\left(\mathbf{Z};\mathbf{Z}^{\left(l\right)}\right)$, a lower bound of $\mathcal{I}\left(\phi\left(\mathbf{Z}\right);\mathbf{Z}\right)$ in analysis because it has a closed-form expression. Once $\mathcal{I}\left(\phi\left(\mathbf{Z}\right);\mathbf{Z}\right)$ is maximized under specific condition, mutual information terms $\mathcal{I}\left(\phi\left(\mathbf{Z}\right);\mathbf{Z}\right)$ and $\mathcal{I}\left(\phi\left(\mathbf{X}\right);\mathbf{X}\right)$ are both maximized because of their lower bound relations (see \textbf{Fig. 5b} for illustration).

\subsection{Simultaneous maximization of mutual information and $\beta$}\label{Sec4-2}

\begin{figure*}[!t]
\includegraphics[width=1\columnwidth]{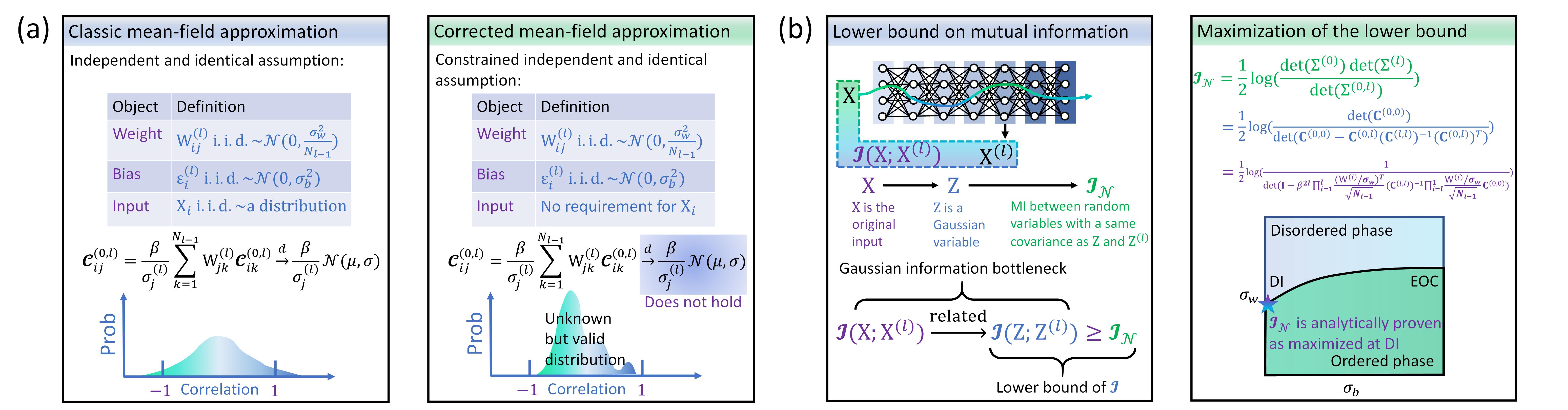}
\caption{Conceptual illustrations of the corrected mean-field approximation and the maximization of a lower bound of mutual information in the phase space of information propagation. (a) The corrected mean-field approximation differs from the classic one by excluding the independent and identical assumption on the components of input signal $\mathbf{X}$. This difference makes central limit theorem do not hold and imply a non-Gaussian distribution of correlation $\mathcal{C}_{ij}^{\left(0,l\right)}$. (b) Although the corrected mean-field approximation overcomes the limitation of the classic one, its loose constraints make the analytic characterization of information channel properties nearly impossible. To create the possibility of analytic derivations, Gaussian information bottleneck is introduced into our analysis to relate the arbitrarily distributed input signal $\mathbf{X}$ with a multivariate Gaussian variable $\mathbf{Z}$. Maximizing $\mathcal{I}_{\mathcal{N}}$, the lower bound of $\mathcal{I}\left(\mathbf{Z};\mathbf{Z}^{\left(l\right)}\right)$, is closely related to optimizing $\mathcal{I}\left(\mathbf{X};\mathbf{X}^{\left(l\right)}\right)$ according to Gaussian information bottleneck. Our theory analytically proves that $\mathcal{I}_{\mathcal{N}}$ is maximized at dynamic isometry, which is validated by computational experiments on real neural networks as well.} 
\end{figure*}

Given the importance of the lower bound of mutual information defined in Eqs. (\ref{EQ34}-\ref{EQ35}), let us begin our formal analysis on it. To relate our calculations of Eq. (\ref{EQ35}) with the phase space of information propagation, we can first consider the mathematical connection between covariance matrix and correlation matrix. In the limit of infinite-width, we can treat the variance of each component $\mathbf{Z}^{\left(l\right)}_{i}$ of the propagated signal $\mathbf{Z}^{\left(l\right)}$ as generally similar (i.e., the fluctuation of variance is sufficiently small compared with network width such that $\sigma^{\left(l\right)}_{k}$ principally maintains the same across different $k$ in the $l$-th layer). This simplification supports us to follow the idea underlying Eq. (\ref{EQ22}) to derive a correlation matrix
\begin{align}
    \mathbf{C}^{\left( 0,l \right)} = \frac{\beta}{\sqrt{N_{l-1}}} \mathbf{C}^{\left(0,l-1\right)}\left(\mathbf{W}^{\left(l\right)}/\sigma_{w}\right)^{2},\label{EQ36}
\end{align}
where the $\left(i,j\right)$-th element in the matrix measures the correlation between $\mathbf{Z}_{i}$ and $\mathbf{Z}^{\left(l\right)}_{j}$ (see \textbf{Fig. 3c}). Term $\beta$ remains same as its definition in Sec \ref{Sec3-1}
\begin{align}
    \beta&= \sigma_{w} \int_{\mathbb{R}} \frac{\psi\left(\sigma^{\left(l-1\right)}_{k} z_{k}\right)}{\sigma^{\left(l\right)}_j} z_{k}\mathsf{D}z_{k}, \label{EQ37}\\
    &= \sigma_{w} \int_{\mathbb{R}} \frac{\psi\left(\sigma^{*}\left(\sigma_{w},\sigma_{b}\right) z\right)}{\sigma^{*}\left(\sigma_{w},\sigma_{b}\right)} z\mathsf{D}z,\label{EQ38}
\end{align}
where Eq. (\ref{EQ38}) is derived by replacing $\sigma_j^{\left(l\right)}$ and $\sigma_k^{\left(l-1\right)}$ with the stable fixed point $\sigma^{*}\left(\sigma_{w},\sigma_{b}\right)$ mentioned in Sec. \ref{Sec2-3}. As suggested later in Sec. \ref{Sec5}, this replacement is reasonable because the convergence rates of $\sigma_j^{\left(l\right)}$ and $\sigma_k^{\left(l-1\right)}$ to $\sigma^{*}\left(\sigma_{w},\sigma_{b}\right)$ are high. Please note that that subscript $k$ in Eq. (\ref{EQ37}) can be eventually dropped in Eq. (\ref{EQ38}) because $\beta$ is same across different $k$ in the $\left(l-1\right)$-th layer. Based on Eq. (\ref{EQ36}) and Eq. (\ref{EQ38}), we can derive the recursion equation of $\mathbf{C}^{\left( 0,l \right)}$
\begin{align}
    \mathbf{C}^{\left( 0,l \right)} = \beta^{l}\mathbf{C}^{\left(0,0\right)} \prod_{i=1}^{l} \frac{\left(\mathbf{W}^{\left(i\right)}/\sigma_{w}\right)^{T}}{\sqrt{N_{i-1}}}.\label{EQ39}
\end{align}

Our next step is to relate Eq. (\ref{EQ39}) with the lower bound of mutual information in Eq. (\ref{EQ35}). Applying the property of the determinant of block matrix, we can reformulate term $\operatorname{det}\left(\mathbf{\Sigma}^{\left(0,l\right)}\right)$ in Eq. (\ref{EQ35}) as
\begin{align}
&\operatorname{det}\left(\mathbf{\Sigma}^{\left(0,l\right)}\right) =  \operatorname{det}\left(\mathbf{\Sigma}^{\left(l\right)}\right)\notag\\ &\times\operatorname{det}\left(\mathbf{\Sigma}^{\left(0\right)}-\mathbf{D}^{\left(0\right)}\mathbf{C}^{\left( 0,l \right)}\mathbf{D}^{\left(l\right)}\left(\mathbf{\Sigma}^{\left(l\right)}\right)^{-1}\mathbf{D}^{\left(l\right)}\left(\mathbf{C}^{\left( 0,l \right)} \right)^{T}\mathbf{D}^{\left(0\right)} \right),\label{EQ40}
\end{align}
where $D^{\left(l\right)}=\operatorname{diag}\left(\mathbf{\Sigma}^{\left(l\right)}\right)^{\frac{1}{2}}$ and $D^{\left(0\right)}=\operatorname{diag}\left(\mathbf{\Sigma}^{\left(0\right)}\right)^{\frac{1}{2}}$, i.e., their diagonal elements are the standard deviations of signals. Please note that $\mathbf{Z}$ and $\mathbf{Z}^{\left(l\right)}$ are jointly Gaussian while calculating $\mathcal{I}_{\mathcal{N}}\left(\mathbf{Z};\mathbf{Z}^{\left(l\right)}\right)$. Therefore, their covariance matrices, $\mathbf{\Sigma}^{\left(0\right)}$ and $\mathbf{\Sigma}^{\left(l\right)}$, are invertible in Eq. (\ref{EQ40}). Then, we can notice that $\mathbf{\Sigma}^{\left(0\right)}=\mathbf{D}^{(0)} \mathbf{C}^{\left( 0,0 \right)}\mathbf{D}^{(0)}$ and $\mathbf{\Sigma}^{\left(l\right)}=\mathbf{D}^{(l)} \mathbf{C}^{\left( l,l \right)}\mathbf{D}^{(l)}$. Therefore, we can reformulate the lower bound of mutual information as 
\begin{align}
    &\mathcal{I}_{\mathcal{N}}\left(\mathbf{Z};\mathbf{Z}^{\left(l\right)}\right)\notag\\=&\frac{1}{2}\log\left(\frac{\operatorname{det}\left(\mathbf{\Sigma}^{\left(0\right)}\right)\operatorname{det}\left(\mathbf{\Sigma}^{\left(l\right)}\right)}{\operatorname{det}\left(\mathbf{\Sigma}^{\left(0,l\right)}\right)}\right),\label{EQ41}\\
    =&\frac{1}{2}\log \left(\frac{\operatorname{det}\left(\mathbf{\Sigma}^{\left(0\right)} \right)}{\operatorname{det}\left(\mathbf{\Sigma}^{\left(0\right)}-\mathbf{D}^{\left(0\right)}\mathbf{C}^{\left( 0,l \right)}\left(\mathbf{C}^{\left(l,l\right)}\right)^{-1}\left(\mathbf{C}^{\left( 0,l \right)}\right)^{T} \mathbf{D}^{\left(0\right)} \right)}\right),\label{EQ42}\\
    =&\frac{1}{2}\log\left(\frac{\operatorname{det}\left(  \mathbf{C}^{\left( 0,0 \right)}\right)} {\operatorname{det}\left(\mathbf{C}^{\left(0,0\right)}-\mathbf{C}^{\left( 0,l \right)}\left(\mathbf{C}^{\left(l,l\right)}\right)^{-1}\left(\mathbf{C}^{\left( 0,l \right)}\right)^{T} \right)}\right).\label{EQ43}
\end{align}
In Eq. (\ref{EQ42}), we have replaced $\mathbf{D}^{\left(l\right)}\left(\mathbf{\Sigma}^{\left(l\right)}\right)^{-1}\mathbf{D}^{\left(l\right)}$ with $\left(\mathbf{C}^{\left(l,l\right)}\right)^{-1}$ for simplification. Eq. (\ref{EQ43}) is obtained by dividing the numerator and denominator of Eq. (\ref{EQ42}) by $\left(\operatorname{det}\left(\mathbf{D}^{\left(0\right)} \right)\right)^{2}$ simultaneously. In Eq. (\ref{EQ43}), matrix $\mathbf{C}^{\left(l,l\right)}$ is a constant matrix across different layers when signals arrive at their stable states shown in Eq. (\ref{EQ38}). Matrix $\mathbf{C}^{\left(0,0\right)}$ is fully determined by input $\mathbf{Z}$ (see \textbf{Fig. 6a}).

After substituting Eq. (\ref{EQ39}) into Eq. (\ref{EQ43}) and dividing the numerator and denominator of the derived result by $\mathbf{C}^{\left(0,0\right)}$, we can obtain the following equation
\begin{widetext}
\begin{align}
    \mathcal{I}_{\mathcal{N}}\left(\mathbf{X};\mathbf{X}^{\left(l\right)}\right)=\frac{1}{2}\log\left(\frac{1} {\operatorname{det}\left(\mathbf{I}-\beta^{2l}\prod_{i=1}^{l} \frac{\left(\mathbf{W}^{\left(i\right)}/\sigma_{w}\right)^{T}}{\sqrt{N_{i-1}}}\left(\mathbf{C}^{\left(l,l\right)}\right)^{-1}\prod_{i=l}^{1} \frac{\mathbf{W}^{\left(i\right)}/\sigma_{w}}{\sqrt{N_{i-1}}}  \mathbf{C}^{\left(0,0\right)} \right)}\right).\label{EQ44}
\end{align}
\end{widetext}
Please note that we have replaced $\left(\mathbf{C}^{(0,0)}\right)^{T}$ with $\mathbf{C}^{(0,0)}$ to improve the readability of Eq. (\ref{EQ44}) because $\mathbf{C}^{(0,0)}$ is a symmetric matrix. In Eq. (\ref{EQ44}), term $\prod_{i=1}^{l} \frac{\mathbf{W}^{\left(i\right)}/\sigma_{w}}{\sqrt{N_{i-1}}}$ is completely determined by the type of weight distribution used in neural network initialization. Meanwhile, matrices $\mathbf{C}^{\left(0,0\right)}$ and $\mathbf{C}^{\left(l,l\right)}$ have been suggested as fully deterministic. Therefore, term $\beta=\beta\left(\sigma_{w},\sigma_{b}\right)$ is the only one non-trivial variable in Eq. (\ref{EQ44}) remaining for analysis.

Our analysis guides us to focus on the possibility that $\mathcal{I}_{\mathcal{N}}\left(\mathbf{Z};\mathbf{Z}^{\left(l\right)}\right)$ can monotonically increase with $\beta=\beta\left(\sigma_{w},\sigma_{b}\right)$ in Eq. (\ref{EQ44}) (see \textbf{Fig. 6a} for illustration). Below, we present our detailed derivations. For convenience, we define
\begin{align}
    \widetilde{ \mathbf{M}}=\prod_{i=1}^{l} \frac{\left(\mathbf{W}^{\left(i\right)}/\sigma_{w}\right)^{T}}{\sqrt{N_{i-1}}}\left(\mathbf{C}^{\left(l,l\right)}\right)^{-1}\prod_{i=l}^{1} \frac{\mathbf{W}^{\left(i\right)}/\sigma_{w}}{\sqrt{N_{i-1}}}.\label{EQ45}
\end{align}
as a shorthand. We notice that we can equivalently verify whether $\operatorname{det}\left(\mathbf{I}-\beta^{2l}\widetilde{ \mathbf{M}} \mathbf{C}^{\left(0,0\right)} \right)$ monotonically decreases with $\beta$ since $\mathcal{I}_{\mathcal{N}}\left(\mathbf{X};\mathbf{X}^{\left(l\right)}\right)$ monotonically decreases with $\operatorname{det}\left(\mathbf{I}-\beta^{2l}\widetilde{ \mathbf{M}} \mathbf{C}^{\left(0,0\right)} \right)$. Because $\mathbf{C}^{\left(0,0\right)}$ is a positive definite matrix, we can apply Cholesky factorization on it, i.e., $\mathbf{C}^{\left(0,0\right)}=LL^{T}$ where $L$ is a lower triangular matrix whose diagonal elements are positive. Then, we have
\begin{align}
&\operatorname{det}\left(\mathbf{I}-\beta^{2l} \widetilde{ \mathbf{M}} \mathbf{C}^{\left(0,0\right)} \right)\notag\\
= &\operatorname{det}\left(\mathbf{I}-\beta^{2l}\widetilde{ \mathbf{M}}LL^T \right),\label{EQ46}\\
=& \operatorname{det}\left(L^{T}\left(L^{T}\right)^{-1}-\beta^{2l}L^{T}\widetilde{ \mathbf{M}} LL^{T}\left(L^{T}\right)^{-1} \right),\label{EQ47}\\
= &\operatorname{det}\left(\mathbf{I}-\beta^{2l} L^{T}\widetilde{ \mathbf{M}} L \right).\label{EQ48}
\end{align}
Given that $\widetilde{ \mathbf{M}}$ is a positive semi-definite matrix, we know that $L^{T}\widetilde{\mathbf{M}} L$ is also positive semi-definite. Let $\{\omega_{i}\vert i=1,\ldots,\operatorname{dim}\left(\widetilde{ \mathbf{M}}\right),\;\omega_{i}>0\}$ be a set of eigenvalues of matrix $L^{T}\widetilde{ \mathbf{M}} L$, we have
\begin{align}
    \operatorname{det}\left(\mathbf{I}-\beta^{2l} \widetilde{ \mathbf{M}} \mathbf{C}^{\left(0,0\right)} \right)=\prod_{i=1}^{\operatorname{dim}\left(\widetilde{ \mathbf{M}}\right)} (1-\beta^{2l}\omega_{i}).\label{EQ49}
\end{align}
Let us assume that the range of $\beta^{2}$ is $\left[0,\widehat{\beta}^{2}\right)$, where $0$ stands for zero correlation. To ensure the non-negativity of $\mathcal{I}_{\mathcal{N}}\left(\mathbf{X};\mathbf{X}^{\left(l\right)}\right)$ (i.e., mutual information can not be negative), we know that $\operatorname{det}\left(\mathbf{I}-\beta^{2l} \widetilde{ \mathbf{M}} \mathbf{C}^{\left(0,0\right)} \right)$, a continuous function with respect to $\beta^{2l}$, should be in an interval of $\left(0,1\right]$. According to Eq. (\ref{EQ49}), this non-negativity requires that $\omega_{i}\in \left(0,\widehat{\beta}^{-2l}\right)$. Meanwhile, to ensure that $\operatorname{det}\left(\mathbf{I}-\beta^{2l} \widetilde{ \mathbf{M}} \mathbf{C}^{\left(0,0\right)} \right)$ can monotonically decreases with $\beta=\beta\left(\sigma_{w},\sigma_{b}\right)$, we can derive $\omega_{i}\in \left(0,\widehat{\beta}^{-2l}\right)$ based on the continuity and non-negativity. We can immediately find that these two requirements of $\omega_{i}$ are consistent with each other. Therefore, we can prove that $\mathcal{I}_{\mathcal{N}}\left(\mathbf{Z};\mathbf{Z}^{\left(l\right)}\right)$ monotonically increases with $\beta\left(\sigma_{w},\sigma_{b}\right)$ (see \textbf{Fig. 6a} for a summary).

Given a specific weight distribution and a certain network depth defined for neural network initialization, terms $\mathbf{C}^{\left(0,0\right)}$, $\prod_{i=1}^{l} \frac{\mathbf{W}^{\left(i\right)}/\sigma_{w}}{\sqrt{N_{i-1}}}$, and $\mathbf{C}^{\left(l,l\right)}$ are principally fixed. Therefore, initializing neural networks for mutual information maximization essentially requires us to maximize $\beta\left(\sigma_{w},\sigma_{b}\right)$, whose condition is analyzed in Sec. \ref{Sec4-3}.

\begin{figure*}[!t]
\includegraphics[width=1\columnwidth]{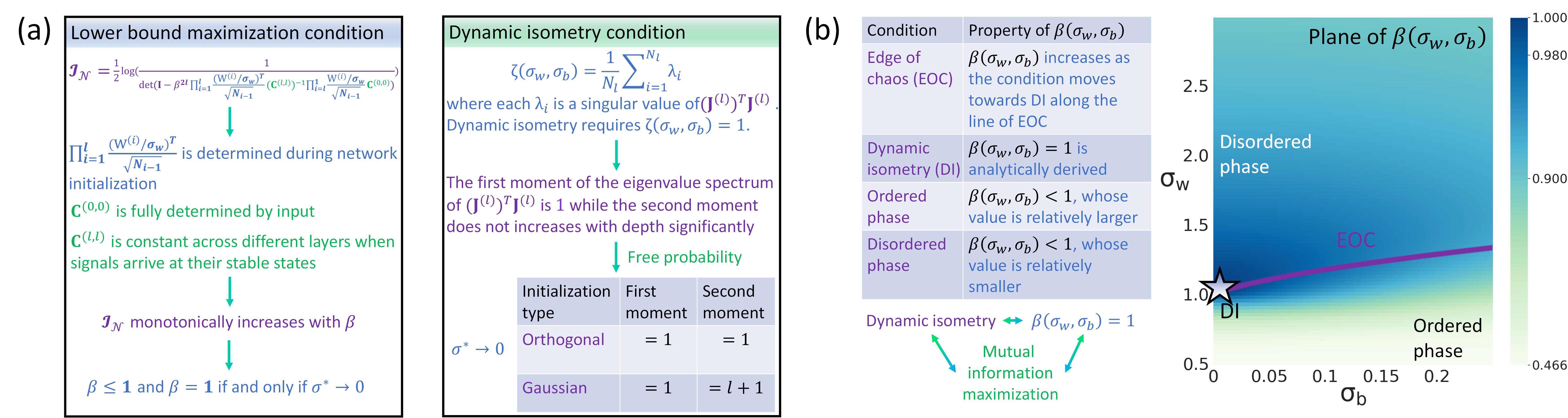}
\caption{Conceptual illustrations of the maximization of the lower bound of mutual information at dynamic isometry. (a) Because $\mathcal{I}_{\mathcal{N}}\left(\mathbf{Z};\mathbf{Z}^{\left(l\right)}\right)$, the lower bound of mutual information, monotonically increases with $\beta\leq 1$, maximizing $\mathcal{I}_{\mathcal{N}}\left(\mathbf{Z};\mathbf{Z}^{\left(l\right)}\right)$ is equivalent to maximizing $\beta\leq 1$, whose condition is analytically derived as $\sigma^{*}\rightarrow 0$. Meanwhile, the condition of dynamic isometry can also be proven as $\sigma^{*}\rightarrow 0$ based on the free probability theory. (b) Consequently, there are equivalent relations among dynamic isometry, lower bound maximization, and $\beta= 1$. As shown in the analytically calculated plane of $\beta\left(\sigma_{w},\sigma_{b}\right)$ (the edge of chaos is represented by a purple line while dynamic isometry point is marked by a star), there is $\beta= 1$ at dynamic isometry (DI), suggesting the maximization of $\mathcal{I}_{\mathcal{N}}\left(\mathbf{Z};\mathbf{Z}^{\left(l\right)}\right)$. Here the plane of $\beta\left(\sigma_{w},\sigma_{b}\right)$ is calculated based on a commonly used non-linear activation function, $\tanh\left(\cdot\right)$, in deep learning.} 
\end{figure*}

\subsection{Mutual information maximization at dynamic isometry}\label{Sec4-3}
As we have proved in Sec. \ref{Sec4-2}, the maximization of the lower bound of mutual information is equivalent to the maximization of $\beta\left(\sigma_{w},\sigma_{b}\right)$. Below, we prove that $\beta\left(\sigma_{w},\sigma_{b}\right)$ is maximized at dynamic isometry. For convenience, we use $\beta$, $\zeta$, and $\sigma^{*}$ as the shorthands of $\beta\left(\sigma_{w},\sigma_{b}\right)$, $\zeta\left(\sigma_{w},\sigma_{b}\right)$, and $\sigma^{*}\left(\sigma_{w},\sigma_{b}\right)$ in our derivations.

Reformulating the one-dimensional integral as two-dimensional integral by Cauchy–Schwarz inequality, we can derive 
\begin{align}
    \left(\int_{\mathbb{R}} \psi\left(\sigma^{*}z\right) z\mathsf{D}z\right)^{2}&=\int_{\mathbb{R}}\int_{\mathbb{R}} \psi\left(\sigma^{*}x\right)\psi\left(\sigma^{*}y\right) x y\mathsf{D}x\mathsf{D}y,\label{EQ50}\\
&\leq \int_{\mathbb{R}}\int_{\mathbb{R}} \psi\left(\sigma^{*}x\right)^{2} y^{2}\mathsf{D}x\mathsf{D}y,\label{EQ51}
\end{align}
where equality holds only if $\psi\left(\sigma^{*}x\right)^{2}\propto x^{2}$. Because $\sigma^{*}$ satisfies
\begin{align}
\left(\sigma^{*}\right)^{2}=\sigma_{w}^{2}\int_{\mathbb{R}}\psi\left(\sigma^{*}z\right)^{2} \mathsf{D}z +\sigma_{b}^{2},\label{EQ52}
\end{align}
we can combine Eq. (\ref{EQ52}) with Eq. (\ref{EQ38}) and Eq. (\ref{EQ51}) to prove $\beta<1$ when $\sigma^{*}\neq 0$
\begin{align}
    \left(\int_{\mathbb{R}} \psi\left(\sigma^{*}z\right) z\mathsf{D}z\right)^{2}&<\int_{\mathbb{R}}\int_{\mathbb{R}} \psi\left(\sigma^{*}x\right)^{2} y^{2}\mathsf{D}x\mathsf{D}y,\label{EQ53}\\
    \sigma_w^2 \left(\int_{\mathbb{R}} \psi\left(\sigma^{*} z\right) z\mathsf{D}z\right)^{2} &< \left(\sigma^{*}\right)^{2},\label{EQ54}\\
    \beta^{2}&< 1.\label{EQ55}
\end{align}

As for the situation where $\sigma^{*}\rightarrow 0$, we can prove that $\mathcal{I}_{\mathcal{N}}\left(\mathbf{X};\mathbf{X}^{\left(l\right)}\right)$ is maximized at dynamic isometry by proving $\beta=1$ under the corresponding condition. As \emph{a priori} knowledge, we can know $\lim_{z\rightarrow +\infty} \psi\left(z\right)=c\in\mathbb{R}$ (i.e., the integral result is a constant) and $\vert\psi^{\prime}\left(0\right)\vert\geq\vert\psi^{\prime}\left(z\right)\vert$ because $\psi\left(\cdot\right)$, an activation function of neural networks, is usually an odd function that is convex in $\left[0,\infty\right]$ and satisfies $\lim_{z\rightarrow +\infty} \psi\left(z\right)=c\in\mathbb{R}$. These properties support us to derive the following proof. First, based on the convex property of $\psi\left(\cdot\right)$ and $1=\int_{\mathbb{R}} z^{2}\mathsf{D}z$ (i.e., $\mathsf{D}z$ is a Gaussian measure), we can derive 
\begin{align}
&\left(\sigma^{\left(l\right)}\right)^{2}=\int_{\mathbb{R}}\left(\sigma^{\left(l\right)}\right)^{2} z^{2}\mathsf{D}z\geq\sigma_{w}^{2} \int_{\mathbb{R}}\psi\left(\sigma^{\left(l\right)} z\right)^{2} \mathsf{D}z \label{EQ56}
\end{align}
when $\sigma_{w}\leq \frac{1}{\psi^{\prime}\left(0\right)}$. Here the equality holds only if $\sigma^{\left(l\right)}=0$. Based on Eq. (\ref{EQ56}), we can see that any $\left(\sigma^{\left(l\right)}\right)^{2}$ will decrease until it arrives at $\left(\sigma^{*}\right)^{2}=0$ when $\sigma_{w}\leq \frac{1}{\psi^{\prime}\left(0\right)}$ and $\sigma_{b}=0$. In other words, point $\sigma^{*}\rightarrow 0$ is a stable fixed point only if $\sigma_{w}\leq \frac{1}{\psi^{\prime}\left(0\right)}$ and $\sigma_{b}=0$. Second, given the condition for $\sigma^{*}\rightarrow 0$ to become a stable fixed point, we can further prove that $\beta=1$ may emerge under this condition. Our derivation utilizes an equality obtained in Appendix \ref{ASec2}.
\begin{align}
    \psi^{\prime}\left(0\right)=\lim_{\sigma^{*}\rightarrow 0} \int_{\mathbb{R}} \frac{\psi\left(\sigma^{*} z\right)}{\sigma^{*}} z\mathsf{D}z .  \label{EQ57}
\end{align}
Substituting Eq. (\ref{EQ57}) into Eq. (\ref{EQ38}), we can see the desired combination of $\left(\sigma_{w},\sigma_{b}\right)$ for $\beta=1$ when $\sigma^{*}\rightarrow 0$
\begin{align}
    \sigma_{w}&=\frac{\beta}{\lim_{\sigma^{*}\rightarrow 0} \int_{\mathbb{R}} \frac{\psi\left(\sigma^{*} z\right)}{\sigma^{*}\left(\sigma_{w},\sigma_{b}\right)} z\mathsf{D}z}=\frac{1}{\psi^{\prime}\left(0\right)},\label{EQ58}\\
    \sigma_{b}&=0.\label{EQ59}
\end{align}
To this point, we have derived Eqs. (\ref{EQ58}-\ref{EQ59}) as the sufficient and necessary condition for $\beta$ to take its maximum, i.e. $\mathcal{I}_{\mathcal{N}}\left(\mathbf{X};\mathbf{X}^{\left(l\right)}\right)$ takes its maximum, which will be shown as exactly the condition of dynamical isometry in our subsequent analysis.


Although previous studies have studied the condition of dynamical isometry in linear \cite{saxe2013exact} and non-linear neural networks \cite{pennington2017resurrecting,pennington2018emergence} as suggested in Sec. \ref{Sec2-3}, it remains unclear if it is possible to relate the condition of dynamical isometry with our theory. Below, we present our detailed derivations of dynamic isometry point $\left(\sigma_{w}^{\diamond},\sigma_{b}^{\diamond}\right)$ based on free probability theory \cite{mingo2017free} to suggest such a possibility.

The key idea in our derviations is to take advantage of the property of $\mathcal{S}$-transform concerning matrix multiplication \cite{speicher1994multiplicative,voiculescu1992free}
\begin{align}
\mathcal{S}_{UV}\left(z\right)=\mathcal{S}_{U}\left(z\right)\mathcal{S}_{V}\left(z\right), \label{EQ60}
\end{align}
where $U$ and $V$ are two freely independent random matrices. Because the Jacobian matrix of the neural network can be defined as 
\begin{align}
    &\mathbf{J}=\frac{\partial \mathbf{X}^{\left(l\right)}}{\partial \mathbf{X}}=\prod_{i=1}^{l} \mathbf{\widetilde {D}}^{\left(i\right)}\mathbf{W}^{\left(i\right)},\label{EQ61}
\end{align}
we can derive the $\mathcal{S}$-transform of the Jacobian matrix 
\begin{align}
   \mathcal{S}_{\mathbf{J}\mathbf{J}^{T}}\left(z\right) = \mathcal{S}^{L}_{\left(\mathbf{\widetilde {D}}^{\left(i\right)}\right)^{2}} \mathcal{S}^{L}_{\left(\mathbf{W}^{\left(i\right)}\right)^{2}\mathbf{W}^{\left(i\right)}}\left(z\right),\label{EQ62}
\end{align}
where $\mathbf{\widetilde {D}}^{\left(i\right)}$ is a diagonal matrix whose diagonal elements are $\mathbf{\widetilde {D}}^{(i)}_{jj}=\psi^{\prime}\left( \mathbf{X}^{\left(i\right)}_{j} \right)$ (please note that the definition is different from the matrix $\mathbf{{D}}^{(i)}$ analyzed before). In the derivation of Eq. (\ref{EQ62}), we have applied that $\mathbf{\widetilde {D}}^{\left(a\right)}=\mathbf{\widetilde {D}}^{\left(b\right)}$ and $\mathbf{W}^{\left(a\right)}=\mathbf{W}^{\left(b\right)}$ for any pair of $\left(a,b\right)$. This property holds because every layer in the neural network shares the same network initialization settings and signals in every layer share the same marginal distribution if they are at the stable fixed point. 

Following the idea in Ref. \cite{pennington2018emergence}, we can derive an implicit equation for eigenvalue spectrum of $\mathbf{H}^{\left(l\right)}=\left(\mathbf{J}^{\left(l\right)}\right)^{T}\mathbf{J}^{\left(l\right)}$ based on Eq. (\ref{EQ62})
\begin{align}
&M_{\mathbf{H}^{\left(l\right)}}(z)=\notag \\ &M_{\mathbf{\widetilde {D}}^2}\left(z^{\frac{1}{l}} \mathcal{S}_{\mathbf{W}^T \mathbf{W}}\left(M_{\mathbf{H}^{\left(l\right)}}(z)\right)\left(1+\frac{1}{M_{ \mathbf{H}^{\left(l\right)} }(z)}\right)^{1-\frac{1}{l}}\right),\label{EQ63}
\end{align}
where $M$ is a moment generating function. After expanding Eq. (\ref{EQ63}) in the powers of $z^{-1}$, the expression of the first two moments of the eigenvalue spectrum of $\mathbf{H}^{\left(l\right)}$ can be obtained as
\begin{align}
&m_1=\left(\sigma_w^2 \tilde{\mu}_1\right)^l, \label{EQ64} \\
&m_2=\left(\sigma_w^2 \tilde{\mu}_1\right)^{2 l} \left(\frac{\tilde{\mu}_2}{\tilde{\mu}_1^2}+\frac{1}{l}-1-s_1\right)l,\label{EQ65}
\end{align}
where we define $\tilde{\mu}_k =\int\psi^{\prime }\left( 
\sigma^{*}z \right)^{2k} \mathsf{D}z$. Meanwhile, one can notice that $m_{1}$ is exactly equivalent to $\zeta$ defined in Eqs. (\ref{EQ16}-\ref{EQ17}). For scaled orthogonal initialization (i.e., weight matrices are initialized as orthogonal random matrices), we have $s_1=0$. For scaled Gaussian initialization (i.e., weight matrices are initialized as Gaussian random matrices), we have $s_1=-1$. More calculation details of $s_1$ can be seen in Ref. \cite{pennington2018emergence}. 

As suggested in Sec. \ref{Sec2-3}, dynamic isometry requires that the first moment of the eigenvalue spectrum of $\mathbf{H}^{\left(l\right)}$ equals $1$ while the second moment approaches to $0$. In deep neural networks, we reasonably relax the restriction on the second moment and require that the second moment does not increases with network depth $l$ significantly. Applying the Cauchy-Schwarz inequality, we can derive
\begin{align}
    \tilde{\mu}_{1}^{2}=\left(\int\psi^{\prime }\left( 
\sigma^{*}z \right)^{2} \mathsf{D}z\right)^{2}\leq \int\psi^{\prime }\left( 
\sigma^{*}z \right)^{4} \mathsf{D}z = \tilde{\mu}_{2}, \label{EQ66}
\end{align}
where the equality holds (i.e., $ \tilde{\mu}_{1}^{2}=\tilde{\mu}_{2}$) if
\begin{align}
     \psi^{\prime }\left( 
\sigma^{*}z \right)^{2}\propto \psi^{\prime }\left( 
\sigma^{*}z \right)^{4} \Leftrightarrow \sigma^{*}\rightarrow 0. \label{EQ67}
\end{align}
In the case where $m_{1}=1$ and $ \tilde{\mu}_{1}^{2}=\tilde{\mu}_{2}$, we can readily obtain $m_{2}=1$ (i.e., the second moment is a small constant) for orthogonal initialization and $m_{2}=l+1$ (i.e., the second moment increases with network depth linearly instead of exhibiting explosive growth) for Gaussian initialization. Moreover, we can know that $\sigma^{*}\rightarrow 0$ enables $m_{1}=1$ (or equivalently $\zeta=1$) to imply $\sigma_{w}=\frac{1}{\psi^{\prime }\left( 
0 \right)}$, which is exactly the condition of the maximization of $\beta$ (or the maximization of the lower bound of mutual information) defined in Eqs. (\ref{EQ58}-\ref{EQ59}) (see \textbf{Fig. 6a} for a summary).

In sum, we have proven that the maximization of the lower bound of mutual information and dynamic isometry shares the same condition (i.e., Eqs. (\ref{EQ58}-\ref{EQ59})). In other words, mutual information $\mathcal{I}\left( \mathbf{Z};\mathbf{Z}^{\left(l\right)} \right)$ and its lower bound $\mathcal{I}_{\mathcal{N}}\left( \mathbf{Z};\mathbf{Z}^{\left(l\right)} \right)$ are maximized at dynamic isometry (see \textbf{Fig. 6b}). Because $\mathbf{Z}$ is a ``Gaussian part" of $\mathbf{X}$ with $\mathcal{I}\left(\mathbf{Z}^{\left(l\right)};\mathbf{Z}\right)\leq\mathcal{I}\left(\mathbf{X}^{\left(l\right)};\mathbf{X}\right)$ (or equivalently $\mathcal{I}\left(\phi\left(\mathbf{Z}\right);\mathbf{Z}\right)\leq\mathcal{I}\left(\phi\left(\mathbf{X}\right);\mathbf{X}\right)$) as suggested in Sec. \ref{Sec4-1}, we know that $\mathcal{I}\left(\mathbf{X}^{\left(l\right)};\mathbf{X}\right)$ is maximized at dynamic isometry in more general cases.



    \begin{figure*}[!t]
\includegraphics[width=1\columnwidth]{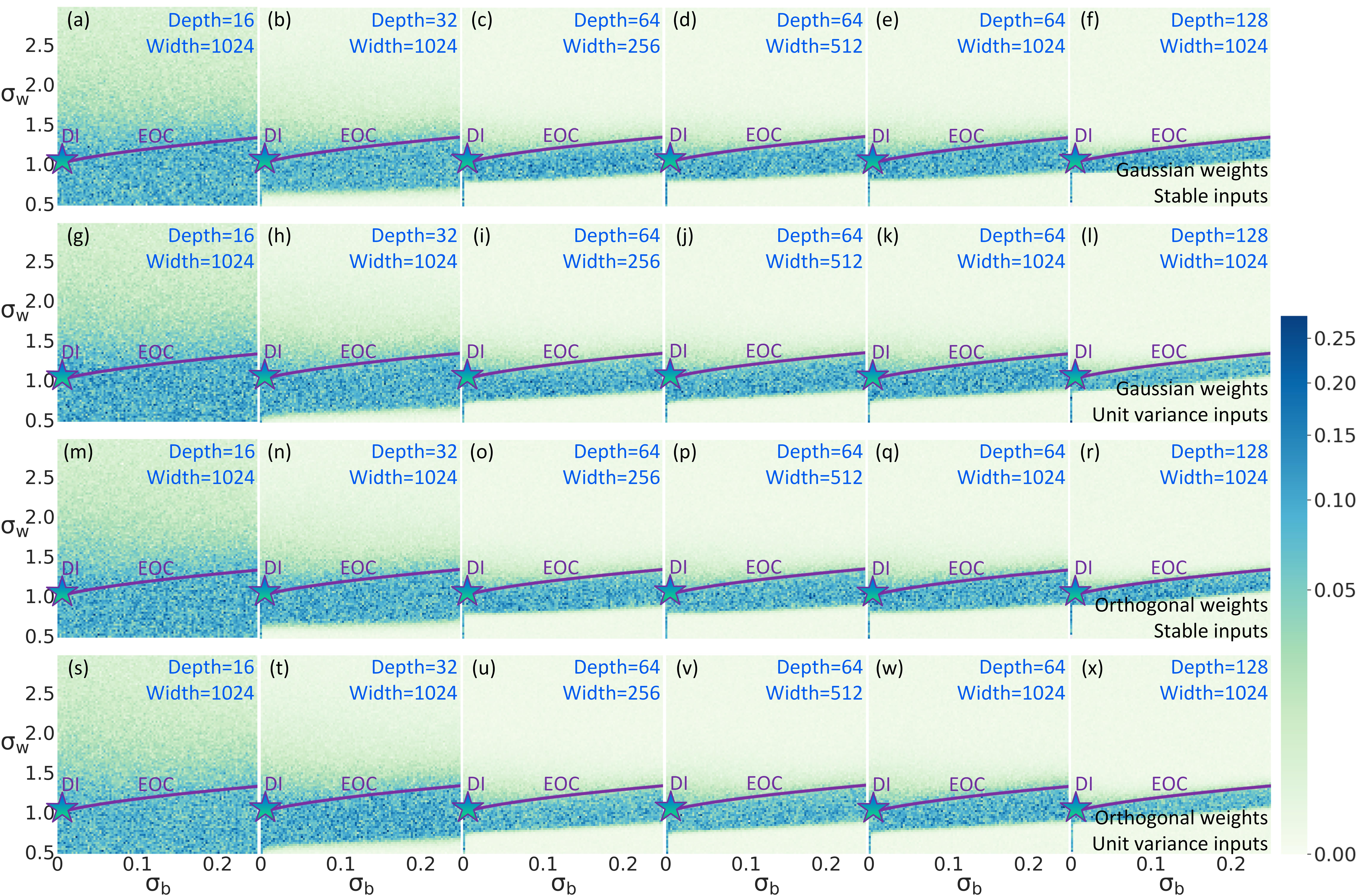}
\caption{The plane of the lower bound of mutual information computationally derived on real neural networks. Experiment settings, such as width, depth, input type, and initialization type, are presented along corresponding planes. Same as \textbf{Fig. 6b}, the edge of chaos (EOC) is marked by a purple line while dynamic isometry is marked by a star.} 
\end{figure*}

\section{Experimental validations}\label{Sec5}
In previous sections, we have present our main theory on the equivalence of dynamic isometry and mutual information maximization, which is developed on infinite-width neural networks. One may reasonably question whether our theory is valid on real finite-width neural networks in deep learning. Below, we validate our theory on real neural networks with various settings (e.g., different widths, depths, inputs, and initialization conditions).

In \textbf{Fig. 7}, we implement our experiments on finite-width neural networks with a widely applied non-linear activation function, $\tanh\left(\cdot\right)$. To comprehensively verify the robustness of our theory against finite size effects, we design these neural networks with different widths and depths. To suggest the applicability of our theory on more general cases where input $\mathbf{Z}$ may not necessarily propagate at its stable state, we distinguish between stable inputs (i.e., propagating at the stable state as our theoretical derivations require) and unit variance inputs (i.e., with a unit covariance matrix that have not been considered in our previous derivations). To show the capacity of our theory to characterize orthogonal and Gaussian initialization, we conduct experiments under both initialization conditions. In our experiments, we measure the lower bound of mutual information between input $\mathbf{Z}$ and output $\phi\left(\mathbf{Z}\right)$ (i.e., $\mathbf{Z}^{\left(l\right)}$ when $l$ stands for the last layer of the neural network) based on Eq. (\ref{EQ35}), in which $\mathbf{\Sigma}^{\left(0\right)}$, $\mathbf{\Sigma}^{\left(l\right)}$, and $\mathbf{\Sigma}^{\left(0,l\right)}$ are computationally estimated from the data. Given a combination of width, depth, input type, and initialization type, we repeat our measurement under each condition of $\left(\sigma_{w},\sigma_{b}\right)$ to obtain a plane of the lower bound of mutual information. To offer a clear vision, we also illustrate the measured lower bound of mutual information along the edge of chaos given orthogonal or Gaussian initialization and stable inputs as two instances (see \textbf{Fig. 8}). As shown in \textbf{Fig. 7} and \textbf{Fig. 8}, our experiment results are consistent with theoretical prediction that the lower bound of mutual information is maximized at the dynamic isometry point $\left(\sigma_{w}^{\diamond},\sigma_{b}^{\diamond}\right)=\left(1,0\right)$ (note that this point is confirmed by whether $\sigma^{*}\rightarrow 0$). Moreover, the distribution of the lower bound of mutual information corroborates the distribution of $\beta$ analytically calculated in \textbf{Fig. 6b}, where both $\beta$ and the lower bound of mutual information on the edge of chaos increase as the condition moves towards dynamic isometry. 

In sum, we have observed consistency between our theory and experiment results, which suggesting the applicability of our theory on real neural networks in deep learning. 

    \begin{figure*}[!t]
\includegraphics[width=1\columnwidth]{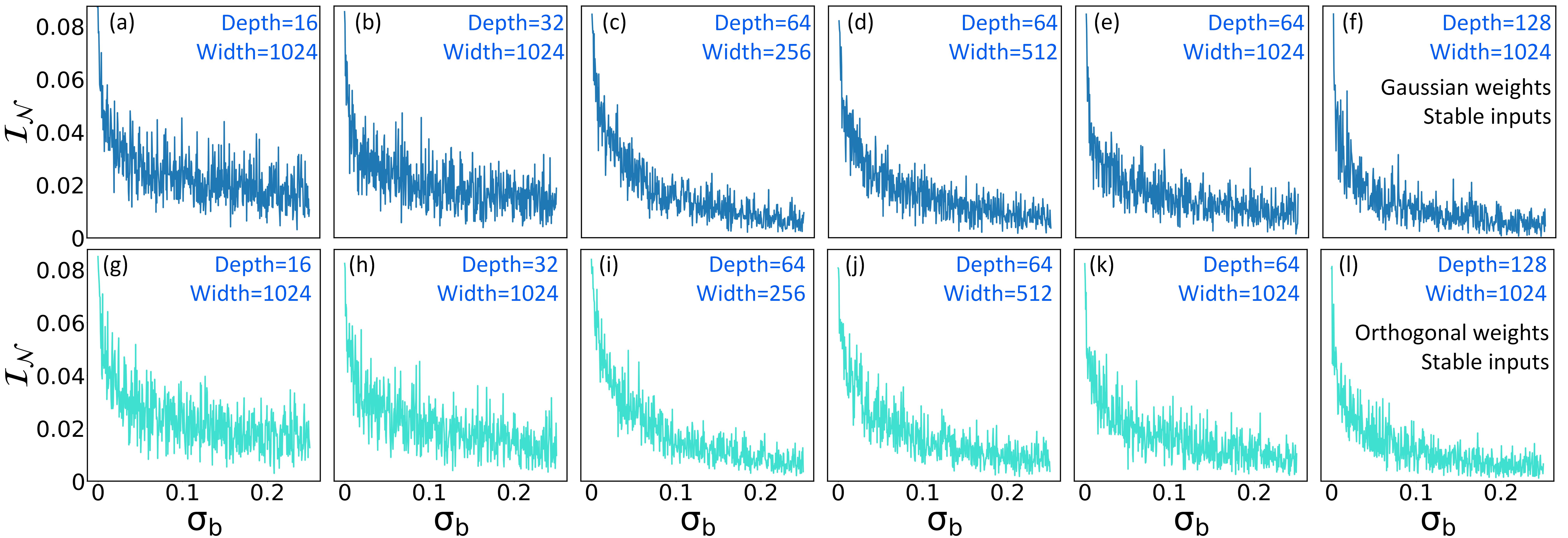}
\caption{Two instances of the computationally measured lower bound of mutual information on the edge of chaos. Experiment settings, such as width, depth, input type, and initialization type, are presented. One can see that $\mathcal{I}_{\mathcal{N}}$, the lower bound of mutual information, is maximized near $\sigma_{b}=0$, which corresponds to the dynamic isometry point.} 
\end{figure*}

\section{Analysis with information bottleneck}\label{Sec6}
To this point, we have analytically developed and computationally validated our theory about mutual information maximization at dynamic isometry. As an interdisciplinary research, our work not only focuses on the statistical physics of neural networks but also aims at offering insights on deep learning techniques. When $\mathcal{I}_{\mathcal{N}}\left( \mathbf{Z};\mathbf{Z}^{\left(l\right)} \right)$, the lower bound of mutual information $\mathcal{I}\left( \mathbf{Z};\mathbf{Z}^{\left(l\right)} \right)$ is maximized because the neural network is initialized at dynamic isometry, we can know $\mathcal{I}\left( \mathbf{X};\mathbf{X}^{\left(l\right)} \right)$ is also maximized according to $\mathcal{I}\left(\mathbf{Z}^{\left(l\right)};\mathbf{Z}\right)\leq\mathcal{I}\left(\mathbf{X}^{\left(l\right)};\mathbf{X}\right)$ in Sec. \ref{Sec4-1}. However, we have suggested that maximizing $\mathcal{I}\left( \mathbf{X};\mathbf{X}^{\left(l\right)} \right)$ is not equivalent to making the neural network an optimal channel in deep learning. Below, we attempt to present a comprehensive analysis on the precise relation between maximizing $\mathcal{I}\left( \mathbf{X};\mathbf{X}^{\left(l\right)} \right)$ and driving neural networks toward optimal channels.

Given the difference between unsupervised and supervised learning, we subdivide our analysis into two cases:
\begin{itemize}
    \item[\textbf{(1)} ] In supervised learning, both sample, $\mathbf{X}$, and target, $\mathbf{Y}$, are accessible for the neural network. Therefore, the optimization objective reduces to the classic information bottleneck \cite{tishby2000information,alemi2016deep,higgins2017betavae}, which is actually a special case of rate distortion theory \cite{berger2003rate} and sufficient statistics theory \cite{kleven2021sufficient}
    \begin{align}
    \max_{\phi}\mathcal{L}_{s}\left(\phi\right):=\underbrace{\mathcal{I}\left(\phi\left(\mathbf{X}\right);\mathbf{Y}\right)}_{\text{Encoding}}\underbrace{-\tau\mathcal{I}\left(\phi\left(\mathbf{X}\right);\mathbf{X}\right)}_{\text{Compression}}.\label{EQ68}
    \end{align}
    In general, Eq. (\ref{EQ68}) defines an objective that the neural network maximizes $\mathcal{I}\left(\phi\left(\mathbf{X}\right);\mathbf{Y}\right)$, the capacity to learn $\mathbf{Y}$, during encoding and minimizes the complexity of representation, $\mathcal{I}\left(\phi\left(\mathbf{X}\right);\mathbf{X}\right)$, during compression. During initialization, we suggest to maximize $\mathcal{I}\left(\phi\left(\mathbf{X}\right);\mathbf{X}\right)$ by initializing the neural network at dynamic isometry because we know $\mathcal{I}\left(\phi\left(\mathbf{X}\right);\mathbf{Y}\right)\leq\mathcal{I}\left(\phi\left(\mathbf{X}\right);\mathbf{X}\right)$ according to the Markov chain in Eq. (\ref{EQ1}). This approach can avoid that $\mathcal{I}\left(\phi\left(\mathbf{X}\right);\mathbf{Y}\right)$ is bounded by a small value of $\mathcal{I}\left(\phi\left(\mathbf{X}\right);\mathbf{X}\right)$ and can not be thoroughly optimized during encoding. During compression, we suggest to minimize $\mathcal{I}\left(\phi\left(\mathbf{X}\right);\mathbf{X}\right)$ while controlling the loss of $\mathcal{I}\left(\phi\left(\mathbf{X}\right);\mathbf{Y}\right)$ (i.e., ensuring $\Delta\vert\mathcal{I}\left(\phi\left(\mathbf{X}\right);\mathbf{Y}\right)\vert<\kappa$ where $\kappa\rightarrow 0$) to avoid neural network over-fitting.
    
    \item[\textbf{(2)} ] In unsupervised learning, the only information accessible to the neural network is sample $\mathbf{X}$. Therefore, the optimization of neural network towards optimal information channel can be implemented following 
    \begin{align}
 \max_{\phi}\mathcal{L}_{u}\left(\phi\right):=\underbrace{\mathcal{I}\left(\phi\left(\mathbf{X}\right);\mathbf{X}\right)}_{\text{Encoding}}\underbrace{-\tau\mathcal{I}\left(\phi\left(\mathbf{X}\right);\mathbf{A}\right)}_{\text{Compression}},\label{EQ69}
    \end{align}
    where $\mathbf{A}\subset\mathbb{N}^{+}$ denotes the index set of samples. Parameter $\tau\in\left(0,\infty\right)$ denotes a Lagrange multiplier. The objective in Eq. (\ref{EQ69}) requires the neural network to maximize the encoded information in its representation, $\mathcal{I}\left(\phi\left(\mathbf{X}\right);\mathbf{X}\right)$, during encoding and reduce the dependence of neural network representation on sample index, $\mathcal{I}\left(\phi\left(\mathbf{X}\right);\mathbf{A}\right)$, during compression. This optimization enables the neural network to learn $\mathbf{Y}=\gamma\left(\mathbf{X}\right)$ under the assumption that sample distribution matches target distribution (e.g., mapping $\gamma$ is invertible such that $\mathcal{I}\left(\phi\left(\mathbf{X}\right);\mathbf{Y}\right)$ can be indirectly optimized through maximizing $\mathcal{I}\left(\phi\left(\mathbf{X}\right);\mathbf{X}\right)$). An intrinsic difference between Eq. (\ref{EQ69}) and classic information bottleneck \cite{tishby2000information,alemi2016deep,higgins2017betavae} lies that there is no strict Markov chain among $\mathbf{A}$, $\mathbf{X}$, $\phi\left(\mathbf{X}\right)$, and $\mathbf{Y}$ because the definition of index set $\mathbf{A}$ is rather flexible in practice. During initialization, we suggest to maximize $\mathcal{I}\left(\phi\left(\mathbf{X}\right);\mathbf{X}\right)$ by initializing the neural network at dynamic isometry. During encoding, we suggest to continue to maximize $\mathcal{I}\left(\phi\left(\mathbf{X}\right);\mathbf{X}\right)$ and apply random data shuffling, a standard trick in real training processes \cite{nguyen2022globally,summers2021nondeterminism}, to make the neural network learn samples rather than over-fit sample index. During compression, we suggest to minimize $\mathcal{I}\left(\phi\left(\mathbf{X}\right);\mathbf{A}\right)$ and ensure $\Delta\vert\mathcal{I}\left(\phi\left(\mathbf{X}\right);\mathbf{X}\right)\vert<\kappa$ for $\kappa\rightarrow 0$.
\end{itemize}

In sum, although neural network initialization can not completely determine the performance in subsequent learning tasks, maximizing $\mathcal{I}\left(\phi\left(\mathbf{X}\right);\mathbf{X}\right)$ based on initialization at dynamic isometry is beneficial for neural network optimization during the encoding phase. The above analysis may be closely related to the empirically observed benefits of dynamic isometry for neural network training (e.g., for convolutional neural networks \cite{xiao2018dynamical} and recurrent neural networks \cite{gilboa2019dynamical}).

\section{Conclusion}\label{Sec7}
In this research, we have explored a frequently neglected possibility that neural networks can be initialized toward optimal information channels in deep learning. 

Compared with some prior studies on related topics (e.g., studies on mean-field dynamics \cite{mei2019mean,nguyen2019mean,poole2016exponential,pennington2017resurrecting,schoenholz2016deep,pennington2018emergence,chen2018dynamical,xiao2018dynamical} and mutual information maximization \cite{ughi2022studies} in neural networks), our work may contribute to both physics and deep learning in the following aspects (see a summary in \textbf{Fig. 2}). First, we present a unified framework to summarize existing works concerning the classic mean-filed approximation of information propagation in neural networks. Second, we indicate the limitation of classic mean-field approximation in characterizing neural networks as information channels (i.e., the implied unreasonable distribution of the correlation measured between inputs and propagated signals). Third, we propose a corrected mean-field approximation of infinite-width neural networks to overcome the limitation of the classic one. Based on the proposed approximation framework and the mechanism underlying Gaussian information bottleneck, we analytically prove that neural networks can realize mutual information maximization between inputs and outputs when they are initialized at dynamic isometry, a case where neural networks serve as norm-preserving random mappings during information propagation. Although initially proposed for infinite-width neural networks, our theory is successfully validated on real finite-width neural networks. Fourth, we have explored an in-depth analysis on the relation between the mutual information maximization emerged at dynamic isometry and driving neural networks towards optimal channels in deep learning tasks.

As a preliminary research, there are diverse intriguing details in our work remained for future exploration (e.g., the accurate quantification of finite-size effects on our theory). The suggested connection between statistical physics and deep learning may be considered as a starting point for more comprehensive interdisciplinary studies bridging between these two fields. 

 \section*{Acknowledgements}
Authors appreciate Wenqing Wei, who studies at the School of science and engineer, Chinese University of Hong kong (Shenzhen), for the inspiration on the proof of that $\mathcal{I}_{\mathcal{N}}\left(\mathbf{X};\mathbf{X}^{\left(l\right)}\right)$ monotonically increases with $\beta(\sigma_{w},\sigma_{b})$. Sirui Huang, who studies at the Qiuzhen College, Tsinghua University, is acknowledged for applying the dominated convergence theorem to derive Eq. (\ref{B3}). Congwen Zhang, who studies at the ZhiLi College, Tsinghua University, is appreciated for proof reading. 

This project is supported by the Artificial and General Intelligence Research Program of Guo Qiang Research Institute at Tsinghua University (2020GQG1017) as well as the Tsinghua University Initiative Scientific Research Program. 

\appendix
\section{Calculation of dynamic isometry}\label{ASec1}
In this section, we present our proof of Eqs. (\ref{EQ34}-\ref{EQ35}). One can also see similar derivations in Ref. \cite{ughi2022studies}.

Let us consider $f(\mathbf{Z})$, the probability density function of an arbitrary variable, and $g(\mathbf{Z})$, the probability density function of a multivariate Gaussian variable
\begin{align}
    g\left(\mathbf{Z}\right) = &\frac{1}{\sqrt{(2\pi)^n \operatorname{det}\left(\mathbf{\Sigma} \right)}}\notag \\ &\times\exp\left[-\frac{1}{2} \left(\mathbf{Z}-\mathbb E\left(\mathbf{Z}\right)\right)^{T}\mathbf{\Sigma}^{(-1)} \left(\mathbf{Z}-\mathbb E\left(\mathbf{Z}\right)\right)  \right],\label{A1}
\end{align}
where $\mathbf{\Sigma}$ and $\mathbb E\left(\mathbf{Z}\right)$ denote the variance matrix and the mean vector shared by $f(\mathbf{Z})$ and $g(\mathbf{Z})$. Then, we can derive
\begin{align}
    &\int f(\mathbf{Z})\log\left[g(\mathbf{Z})\right]\mathrm{d}\mathbf{Z}\notag\\
    =&-\frac{1}{2}\log \left[(2\pi)^n \operatorname{det}\left(\mathbf{\Sigma} \right)\right]-\frac{1}{2}\notag \\  & \times\int f(\mathbf{Z})\log\left[(\mathbf{Z}-\mathbb E\left(\mathbf{Z}\right))^{T}\mathbf{\Sigma}^{-1} (\mathbf{Z}-\mathbb E\left(\mathbf{Z}\right))\right]\mathrm{d}\mathbf{Z},\label{A2}\\
    =&-\frac{1}{2}\log \left[(2\pi)^{n} \operatorname{det}\left(\mathbf{\Sigma} \right)\right]-\frac{\operatorname{tr}\left(\mathbf{\Sigma}^{-1}\mathbf{\Sigma}\right)}{2},\label{A3}\\
    =&\int g(\mathbf{Z})\log\left[g(\mathbf{X})\right]\mathrm{d}\mathbf{Z}.\label{A4}
\end{align}
After replacing $g\left(\mathbf{Z}\right)$ and $f\left(\mathbf{Z}\right)$ by $g\left(\mathbf{Z};\mathbf{Z}^{\left(l\right)}\right)$ and $f\left(\mathbf{Z};\mathbf{Z}^{\left(l\right)}\right)$, constraining the marginal distribution as $f_{\mathbf{Z}}(\mathbf{Z})=g_{\mathbf{Z}}(\mathbf{Z})$, and using $f\left(\mathbf{Z};\mathbf{Z}^{\left(l\right)}\right)$ as the joint distribution of $\left(\mathbf{Z};\mathbf{Z}^{\left(l\right)}\right)$, we can obtain
\begin{align}
    &\mathcal{I}\left( \mathbf{Z};\mathbf{Z}^{\left(l\right)} \right) -
\mathcal{I}_{\mathcal{N}}\left(\mathbf{Z};\mathbf{Z}^{\left(l\right)}\right)\notag\\
    =&\int f\left(\mathbf{Z};\mathbf{Z}^{\left(l\right)}\right)\log\left[ \frac{  f\left(\mathbf{Z};\mathbf{Z}^{\left(l\right)}\right)   }{ f_{\mathbf{Z}}\left(\mathbf{Z}\right)f_{\mathbf{Z}^{\left(l\right)}}\left(\mathbf{Z}^{\left(l\right)}\right)     } \right]\mathrm{d}\mathbf{Z}\mathrm{d}\mathbf{Z}^{\left(l\right)}  \notag \\ 
    &-\int g\left(\mathbf{Z};\mathbf{Z}^{\left(l\right)}\right)\log\left[ \frac{  g\left(\mathbf{Z};\mathbf{Z}^{\left(l\right)}\right)   }{ g_{\mathbf{Z}}\left(\mathbf{Z}\right)g_{\mathbf{Z}^{\left(l\right)}}\left(\mathbf{Z}^{\left(l\right)}\right)     } \right]\mathrm{d}\mathbf{Z}\mathrm{d}\mathbf{Z}^{\left(l\right)},\label{A5}\\
    =&-\int f_{\mathbf{Z}^{\left(l\right)}}\left(\mathbf{Z}^{\left(l\right)}\right)\log\left[  f_{\mathbf{Z}^{\left(l\right)}}\left(\mathbf{Z}^{\left(l\right)}\right)\right]\mathrm{d}\mathbf{Z}^{\left(l\right)}  \notag\\
    &+\int g_{\mathbf{Z}^{\left(l\right)}}\left(\mathbf{Z}^{\left(l\right)}\right)\log\left[  g_{\mathbf{Z}^{\left(l\right)}}\left(\mathbf{Z}^{\left(l\right)}\right)\right]\mathrm{d}\mathbf{Z}^{\left(l\right)} \notag \\
    & +\int f\left(\mathbf{Z};\mathbf{Z}^{\left(l\right)}\right) \log \left[f\left(\mathbf{Z};\mathbf{Z}^{\left(l\right)}\right)\right] \mathrm{d}\mathbf{Z}\mathrm{d}\mathbf{Z}^{\left(l\right)}\notag \\
    & -\int g\left(\mathbf{Z};\mathbf{Z}^{\left(l\right)}\right) \log \left[g\left(\mathbf{Z};\mathbf{Z}^{\left(l\right)}\right)\right] \mathrm{d}\mathbf{Z}\mathrm{d}\mathbf{Z}^{\left(l\right)}, \label{A6}\\
    = &\int f_{\mathbf{Z}^{\left(l\right)}}\left(\mathbf{Z}^{\left(l\right)}\right)\log\left[    \frac{g_{\mathbf{Z}^{\left(l\right)}}\left(\mathbf{Z}^{\left(l\right)}\right) }{f_{\mathbf{Z}^{\left(l\right)}}\left(\mathbf{Z}^{\left(l\right)}\right) }  \right]\mathrm{d}\mathbf{Z}^{\left(l\right)}\notag \\
    &- \int f\left(\mathbf{Z};\mathbf{Z}^{\left(l\right)}\right) \log \left[ 
 \frac{g\left(\mathbf{Z};\mathbf{Z}^{\left(l\right)}\right)}{ f\left(\mathbf{Z};\mathbf{Z}^{\left(l\right)}\right)}  \right] \mathrm{d}\mathbf{Z}\mathrm{d}\mathbf{Z}^{\left(l\right)},\label{A7}\\
 = &\int f\left(\mathbf{Z};\mathbf{Z}^{\left(l\right)}\right)\log \left[ 
 \frac{g_{\mathbf{Z}^{\left(l\right)}} \left(\mathbf{Z}^{\left(l\right)}\right)f\left(\mathbf{Z};\mathbf{Z}^{\left(l\right)}\right)}{f_{\mathbf{Z}^{\left(l\right)}}\left(\mathbf{Z}^{\left(l\right)}\right) g\left(\mathbf{Z};\mathbf{Z}^{\left(l\right)}\right)}  \right] 
     \mathrm{d}\mathbf{Z}\mathrm{d}\mathbf{Z}^{\left(l\right)},\label{A8}\\
     =&\int f_{\mathbf{Z}^{\left(l\right)}}\left(\mathbf{Z}^{\left(l\right)}\right) D_{KL}\left(f_{\mathbf{Z}|\mathbf{Z}^{\left(l\right)}}||g_{\mathbf{Z}|\mathbf{Z}^{\left(l\right)}}\right)\mathrm{d}\mathbf{Z}^{\left(l\right)},\label{A9}\\
     \geq& 0, \label{A10}
\end{align}
where $f_{\mathbf{Z}}(\mathbf{Z})=g_{\mathbf{Z}}(\mathbf{Z})$ is used in Eq. (\ref{A6}) and Eq. (\ref{A4}) is used to derive Eq.(\ref{A7}). Based on Eq.(\ref{A10}), Eqs. (\ref{EQ34}-\ref{EQ35}) in the main text can be proven.

\section{Necessary derivations of Eq. (\ref{EQ57})}\label{ASec2}
In this section, we present our derivations of Eq. (\ref{EQ57}). Let us reformulate the right side of Eq. (\ref{EQ57}) as
\begin{align}
    \int_{\mathbb{R}} \frac{\psi\left(\sigma^{*} z\right)}{\sigma^{*}} z\mathsf{D}z = \int_{\mathbb{R}} \frac{\psi\left(\sigma^{*} z\right)-\psi\left(0\right)}{\sigma^{*}z} z^{2}\mathsf{D}z,\label{B1}
\end{align}
where we have used the fact that $\psi\left(0\right)=0$. Because $\psi\left(\cdot\right)$ is a odd function that is convex in $[0,+\infty]$, we can know $\vert\psi^{\prime}\left(0\right)\vert>\vert\psi^{\prime}\left(z\right)\vert$ for any $z\neq 0$. Then, we have
\begin{align}
    \lim_{\sigma^{*}\rightarrow 0} \frac{\psi\left(\sigma^{*} z\right)-\psi\left(0\right)}{\sigma^{*}z} z^{2}\mathsf{D}z = \psi^{\prime}\left(0\right).\label{B2}
\end{align}
Based on the dominated convergence theorem \cite{stein2009real}, we can obtain
\begin{align}
    \lim_{\sigma^{*}\rightarrow 0} \int_{\mathbb{R}} \frac{\psi\left(\sigma^{*} z\right)-\psi\left(0\right)}{\sigma^{*}z} z^{2}\mathsf{D}z=\psi^{\prime}\left(0\right)\int_{\mathbb{R}}  z^{2}\mathsf{D}z=\psi^{\prime}\left(0\right)\label{B3}
\end{align}
based on Eq. (\ref{B2}), which finishes our derivations on Eq. (\ref{EQ57}) in the main text.
\bibliography{apssamp}

\begin{thebibliography}{67}%
\makeatletter
\providecommand \@ifxundefined [1]{%
 \@ifx{#1\undefined}
}%
\providecommand \@ifnum [1]{%
 \ifnum #1\expandafter \@firstoftwo
 \else \expandafter \@secondoftwo
 \fi
}%
\providecommand \@ifx [1]{%
 \ifx #1\expandafter \@firstoftwo
 \else \expandafter \@secondoftwo
 \fi
}%
\providecommand \natexlab [1]{#1}%
\providecommand \enquote  [1]{``#1''}%
\providecommand \bibnamefont  [1]{#1}%
\providecommand \bibfnamefont [1]{#1}%
\providecommand \citenamefont [1]{#1}%
\providecommand \href@noop [0]{\@secondoftwo}%
\providecommand \href [0]{\begingroup \@sanitize@url \@href}%
\providecommand \@href[1]{\@@startlink{#1}\@@href}%
\providecommand \@@href[1]{\endgroup#1\@@endlink}%
\providecommand \@sanitize@url [0]{\catcode `\\12\catcode `\$12\catcode
  `\&12\catcode `\#12\catcode `\^12\catcode `\_12\catcode `\%12\relax}%
\providecommand \@@startlink[1]{}%
\providecommand \@@endlink[0]{}%
\providecommand \url  [0]{\begingroup\@sanitize@url \@url }%
\providecommand \@url [1]{\endgroup\@href {#1}{\urlprefix }}%
\providecommand \urlprefix  [0]{URL }%
\providecommand \Eprint [0]{\href }%
\providecommand \doibase [0]{https://doi.org/}%
\providecommand \selectlanguage [0]{\@gobble}%
\providecommand \bibinfo  [0]{\@secondoftwo}%
\providecommand \bibfield  [0]{\@secondoftwo}%
\providecommand \translation [1]{[#1]}%
\providecommand \BibitemOpen [0]{}%
\providecommand \bibitemStop [0]{}%
\providecommand \bibitemNoStop [0]{.\EOS\space}%
\providecommand \EOS [0]{\spacefactor3000\relax}%
\providecommand \BibitemShut  [1]{\csname bibitem#1\endcsname}%
\let\auto@bib@innerbib\@empty
\bibitem [{\citenamefont {Bengio}\ \emph {et~al.}(2013)\citenamefont {Bengio},
  \citenamefont {Courville},\ and\ \citenamefont
  {Vincent}}]{bengio2013representation}%
  \BibitemOpen
  \bibfield  {author} {\bibinfo {author} {\bibfnamefont {Y.}~\bibnamefont
  {Bengio}}, \bibinfo {author} {\bibfnamefont {A.}~\bibnamefont {Courville}},\
  and\ \bibinfo {author} {\bibfnamefont {P.}~\bibnamefont {Vincent}},\
  }\bibfield  {title} {\bibinfo {title} {Representation learning: A review and
  new perspectives},\ }\href@noop {} {\bibfield  {journal} {\bibinfo  {journal}
  {IEEE transactions on pattern analysis and machine intelligence}\ }\textbf
  {\bibinfo {volume} {35}},\ \bibinfo {pages} {1798} (\bibinfo {year}
  {2013})}\BibitemShut {NoStop}%
\bibitem [{\citenamefont {Lesort}\ \emph {et~al.}(2018)\citenamefont {Lesort},
  \citenamefont {D{\'\i}az-Rodr{\'\i}guez}, \citenamefont {Goudou},\ and\
  \citenamefont {Filliat}}]{lesort2018state}%
  \BibitemOpen
  \bibfield  {author} {\bibinfo {author} {\bibfnamefont {T.}~\bibnamefont
  {Lesort}}, \bibinfo {author} {\bibfnamefont {N.}~\bibnamefont
  {D{\'\i}az-Rodr{\'\i}guez}}, \bibinfo {author} {\bibfnamefont {J.-F.}\
  \bibnamefont {Goudou}},\ and\ \bibinfo {author} {\bibfnamefont
  {D.}~\bibnamefont {Filliat}},\ }\bibfield  {title} {\bibinfo {title} {State
  representation learning for control: An overview},\ }\href@noop {} {\bibfield
   {journal} {\bibinfo  {journal} {Neural Networks}\ }\textbf {\bibinfo
  {volume} {108}},\ \bibinfo {pages} {379} (\bibinfo {year}
  {2018})}\BibitemShut {NoStop}%
\bibitem [{\citenamefont {Hamilton}\ \emph {et~al.}(2017)\citenamefont
  {Hamilton}, \citenamefont {Ying},\ and\ \citenamefont
  {Leskovec}}]{hamilton2017representation}%
  \BibitemOpen
  \bibfield  {author} {\bibinfo {author} {\bibfnamefont {W.~L.}\ \bibnamefont
  {Hamilton}}, \bibinfo {author} {\bibfnamefont {R.}~\bibnamefont {Ying}},\
  and\ \bibinfo {author} {\bibfnamefont {J.}~\bibnamefont {Leskovec}},\
  }\bibfield  {title} {\bibinfo {title} {Representation learning on graphs:
  Methods and applications},\ }\href@noop {} {\bibfield  {journal} {\bibinfo
  {journal} {arXiv preprint arXiv:1709.05584}\ } (\bibinfo {year}
  {2017})}\BibitemShut {NoStop}%
\bibitem [{\citenamefont {Sch{\"o}lkopf}\ \emph {et~al.}(2021)\citenamefont
  {Sch{\"o}lkopf}, \citenamefont {Locatello}, \citenamefont {Bauer},
  \citenamefont {Ke}, \citenamefont {Kalchbrenner}, \citenamefont {Goyal},\
  and\ \citenamefont {Bengio}}]{scholkopf2021toward}%
  \BibitemOpen
  \bibfield  {author} {\bibinfo {author} {\bibfnamefont {B.}~\bibnamefont
  {Sch{\"o}lkopf}}, \bibinfo {author} {\bibfnamefont {F.}~\bibnamefont
  {Locatello}}, \bibinfo {author} {\bibfnamefont {S.}~\bibnamefont {Bauer}},
  \bibinfo {author} {\bibfnamefont {N.~R.}\ \bibnamefont {Ke}}, \bibinfo
  {author} {\bibfnamefont {N.}~\bibnamefont {Kalchbrenner}}, \bibinfo {author}
  {\bibfnamefont {A.}~\bibnamefont {Goyal}},\ and\ \bibinfo {author}
  {\bibfnamefont {Y.}~\bibnamefont {Bengio}},\ }\bibfield  {title} {\bibinfo
  {title} {Toward causal representation learning},\ }\href@noop {} {\bibfield
  {journal} {\bibinfo  {journal} {Proceedings of the IEEE}\ }\textbf {\bibinfo
  {volume} {109}},\ \bibinfo {pages} {612} (\bibinfo {year}
  {2021})}\BibitemShut {NoStop}%
\bibitem [{\citenamefont {Dike}\ \emph {et~al.}(2018)\citenamefont {Dike},
  \citenamefont {Zhou}, \citenamefont {Deveerasetty},\ and\ \citenamefont
  {Wu}}]{dike2018unsupervised}%
  \BibitemOpen
  \bibfield  {author} {\bibinfo {author} {\bibfnamefont {H.~U.}\ \bibnamefont
  {Dike}}, \bibinfo {author} {\bibfnamefont {Y.}~\bibnamefont {Zhou}}, \bibinfo
  {author} {\bibfnamefont {K.~K.}\ \bibnamefont {Deveerasetty}},\ and\ \bibinfo
  {author} {\bibfnamefont {Q.}~\bibnamefont {Wu}},\ }\bibfield  {title}
  {\bibinfo {title} {Unsupervised learning based on artificial neural network:
  A review},\ }in\ \href@noop {} {\emph {\bibinfo {booktitle} {2018 IEEE
  International Conference on Cyborg and Bionic Systems (CBS)}}}\ (\bibinfo
  {organization} {IEEE},\ \bibinfo {year} {2018})\ pp.\ \bibinfo {pages}
  {322--327}\BibitemShut {NoStop}%
\bibitem [{\citenamefont {Xu}\ and\ \citenamefont
  {Wunsch}(2005)}]{xu2005survey}%
  \BibitemOpen
  \bibfield  {author} {\bibinfo {author} {\bibfnamefont {R.}~\bibnamefont
  {Xu}}\ and\ \bibinfo {author} {\bibfnamefont {D.}~\bibnamefont {Wunsch}},\
  }\bibfield  {title} {\bibinfo {title} {Survey of clustering algorithms},\
  }\href@noop {} {\bibfield  {journal} {\bibinfo  {journal} {IEEE Transactions
  on neural networks}\ }\textbf {\bibinfo {volume} {16}},\ \bibinfo {pages}
  {645} (\bibinfo {year} {2005})}\BibitemShut {NoStop}%
\bibitem [{\citenamefont {Xu}\ and\ \citenamefont
  {Tian}(2015)}]{xu2015comprehensive}%
  \BibitemOpen
  \bibfield  {author} {\bibinfo {author} {\bibfnamefont {D.}~\bibnamefont
  {Xu}}\ and\ \bibinfo {author} {\bibfnamefont {Y.}~\bibnamefont {Tian}},\
  }\bibfield  {title} {\bibinfo {title} {A comprehensive survey of clustering
  algorithms},\ }\href@noop {} {\bibfield  {journal} {\bibinfo  {journal}
  {Annals of Data Science}\ }\textbf {\bibinfo {volume} {2}},\ \bibinfo {pages}
  {165} (\bibinfo {year} {2015})}\BibitemShut {NoStop}%
\bibitem [{\citenamefont {Radford}\ \emph {et~al.}(2015)\citenamefont
  {Radford}, \citenamefont {Metz},\ and\ \citenamefont
  {Chintala}}]{radford2015unsupervised}%
  \BibitemOpen
  \bibfield  {author} {\bibinfo {author} {\bibfnamefont {A.}~\bibnamefont
  {Radford}}, \bibinfo {author} {\bibfnamefont {L.}~\bibnamefont {Metz}},\ and\
  \bibinfo {author} {\bibfnamefont {S.}~\bibnamefont {Chintala}},\ }\bibfield
  {title} {\bibinfo {title} {Unsupervised representation learning with deep
  convolutional generative adversarial networks},\ }\href@noop {} {\bibfield
  {journal} {\bibinfo  {journal} {arXiv preprint arXiv:1511.06434}\ } (\bibinfo
  {year} {2015})}\BibitemShut {NoStop}%
\bibitem [{\citenamefont {Arora}\ \emph {et~al.}(2019)\citenamefont {Arora},
  \citenamefont {Khandeparkar}, \citenamefont {Khodak}, \citenamefont
  {Plevrakis},\ and\ \citenamefont {Saunshi}}]{arora2019theoretical}%
  \BibitemOpen
  \bibfield  {author} {\bibinfo {author} {\bibfnamefont {S.}~\bibnamefont
  {Arora}}, \bibinfo {author} {\bibfnamefont {H.}~\bibnamefont {Khandeparkar}},
  \bibinfo {author} {\bibfnamefont {M.}~\bibnamefont {Khodak}}, \bibinfo
  {author} {\bibfnamefont {O.}~\bibnamefont {Plevrakis}},\ and\ \bibinfo
  {author} {\bibfnamefont {N.}~\bibnamefont {Saunshi}},\ }\bibfield  {title}
  {\bibinfo {title} {A theoretical analysis of contrastive unsupervised
  representation learning},\ }\href@noop {} {\bibfield  {journal} {\bibinfo
  {journal} {arXiv preprint arXiv:1902.09229}\ } (\bibinfo {year}
  {2019})}\BibitemShut {NoStop}%
\bibitem [{\citenamefont {Kotsiantis}\ \emph {et~al.}(2007)\citenamefont
  {Kotsiantis}, \citenamefont {Zaharakis}, \citenamefont {Pintelas} \emph
  {et~al.}}]{kotsiantis2007supervised}%
  \BibitemOpen
  \bibfield  {author} {\bibinfo {author} {\bibfnamefont {S.~B.}\ \bibnamefont
  {Kotsiantis}}, \bibinfo {author} {\bibfnamefont {I.}~\bibnamefont
  {Zaharakis}}, \bibinfo {author} {\bibfnamefont {P.}~\bibnamefont {Pintelas}},
  \emph {et~al.},\ }\bibfield  {title} {\bibinfo {title} {Supervised machine
  learning: A review of classification techniques},\ }\href@noop {} {\bibfield
  {journal} {\bibinfo  {journal} {Emerging artificial intelligence applications
  in computer engineering}\ }\textbf {\bibinfo {volume} {160}},\ \bibinfo
  {pages} {3} (\bibinfo {year} {2007})}\BibitemShut {NoStop}%
\bibitem [{\citenamefont {Hastie}\ \emph {et~al.}(2009)\citenamefont {Hastie},
  \citenamefont {Tibshirani}, \citenamefont {Friedman},\ and\ \citenamefont
  {Friedman}}]{hastie2009elements}%
  \BibitemOpen
  \bibfield  {author} {\bibinfo {author} {\bibfnamefont {T.}~\bibnamefont
  {Hastie}}, \bibinfo {author} {\bibfnamefont {R.}~\bibnamefont {Tibshirani}},
  \bibinfo {author} {\bibfnamefont {J.~H.}\ \bibnamefont {Friedman}},\ and\
  \bibinfo {author} {\bibfnamefont {J.~H.}\ \bibnamefont {Friedman}},\
  }\href@noop {} {\emph {\bibinfo {title} {The elements of statistical
  learning: data mining, inference, and prediction}}},\ Vol.~\bibinfo {volume}
  {2}\ (\bibinfo  {publisher} {Springer},\ \bibinfo {year} {2009})\BibitemShut
  {NoStop}%
\bibitem [{\citenamefont {Deisenroth}\ \emph {et~al.}(2020)\citenamefont
  {Deisenroth}, \citenamefont {Faisal},\ and\ \citenamefont
  {Ong}}]{deisenroth2020mathematics}%
  \BibitemOpen
  \bibfield  {author} {\bibinfo {author} {\bibfnamefont {M.~P.}\ \bibnamefont
  {Deisenroth}}, \bibinfo {author} {\bibfnamefont {A.~A.}\ \bibnamefont
  {Faisal}},\ and\ \bibinfo {author} {\bibfnamefont {C.~S.}\ \bibnamefont
  {Ong}},\ }\href@noop {} {\emph {\bibinfo {title} {Mathematics for machine
  learning}}}\ (\bibinfo  {publisher} {Cambridge University Press},\ \bibinfo
  {year} {2020})\BibitemShut {NoStop}%
\bibitem [{\citenamefont {Song}\ \emph {et~al.}(2022)\citenamefont {Song},
  \citenamefont {Kim}, \citenamefont {Park}, \citenamefont {Shin},\ and\
  \citenamefont {Lee}}]{song2022learning}%
  \BibitemOpen
  \bibfield  {author} {\bibinfo {author} {\bibfnamefont {H.}~\bibnamefont
  {Song}}, \bibinfo {author} {\bibfnamefont {M.}~\bibnamefont {Kim}}, \bibinfo
  {author} {\bibfnamefont {D.}~\bibnamefont {Park}}, \bibinfo {author}
  {\bibfnamefont {Y.}~\bibnamefont {Shin}},\ and\ \bibinfo {author}
  {\bibfnamefont {J.-G.}\ \bibnamefont {Lee}},\ }\bibfield  {title} {\bibinfo
  {title} {Learning from noisy labels with deep neural networks: A survey},\
  }\href@noop {} {\bibfield  {journal} {\bibinfo  {journal} {IEEE Transactions
  on Neural Networks and Learning Systems}\ } (\bibinfo {year}
  {2022})}\BibitemShut {NoStop}%
\bibitem [{\citenamefont {Fr{\'e}nay}\ and\ \citenamefont
  {Verleysen}(2013)}]{frenay2013classification}%
  \BibitemOpen
  \bibfield  {author} {\bibinfo {author} {\bibfnamefont {B.}~\bibnamefont
  {Fr{\'e}nay}}\ and\ \bibinfo {author} {\bibfnamefont {M.}~\bibnamefont
  {Verleysen}},\ }\bibfield  {title} {\bibinfo {title} {Classification in the
  presence of label noise: a survey},\ }\href@noop {} {\bibfield  {journal}
  {\bibinfo  {journal} {IEEE transactions on neural networks and learning
  systems}\ }\textbf {\bibinfo {volume} {25}},\ \bibinfo {pages} {845}
  (\bibinfo {year} {2013})}\BibitemShut {NoStop}%
\bibitem [{\citenamefont {Cover}(1999)}]{cover1999elements}%
  \BibitemOpen
  \bibfield  {author} {\bibinfo {author} {\bibfnamefont {T.~M.}\ \bibnamefont
  {Cover}},\ }\href@noop {} {\emph {\bibinfo {title} {Elements of information
  theory}}}\ (\bibinfo  {publisher} {John Wiley \& Sons},\ \bibinfo {year}
  {1999})\BibitemShut {NoStop}%
\bibitem [{\citenamefont {Shamir}\ \emph {et~al.}(2010)\citenamefont {Shamir},
  \citenamefont {Sabato},\ and\ \citenamefont {Tishby}}]{shamir2010learning}%
  \BibitemOpen
  \bibfield  {author} {\bibinfo {author} {\bibfnamefont {O.}~\bibnamefont
  {Shamir}}, \bibinfo {author} {\bibfnamefont {S.}~\bibnamefont {Sabato}},\
  and\ \bibinfo {author} {\bibfnamefont {N.}~\bibnamefont {Tishby}},\
  }\bibfield  {title} {\bibinfo {title} {Learning and generalization with the
  information bottleneck},\ }\href@noop {} {\bibfield  {journal} {\bibinfo
  {journal} {Theoretical Computer Science}\ }\textbf {\bibinfo {volume}
  {411}},\ \bibinfo {pages} {2696} (\bibinfo {year} {2010})}\BibitemShut
  {NoStop}%
\bibitem [{\citenamefont {Kang}\ and\ \citenamefont
  {Ulukus}(2010)}]{kang2010new}%
  \BibitemOpen
  \bibfield  {author} {\bibinfo {author} {\bibfnamefont {W.}~\bibnamefont
  {Kang}}\ and\ \bibinfo {author} {\bibfnamefont {S.}~\bibnamefont {Ulukus}},\
  }\bibfield  {title} {\bibinfo {title} {A new data processing inequality and
  its applications in distributed source and channel coding},\ }\href@noop {}
  {\bibfield  {journal} {\bibinfo  {journal} {IEEE Transactions on Information
  Theory}\ }\textbf {\bibinfo {volume} {57}},\ \bibinfo {pages} {56} (\bibinfo
  {year} {2010})}\BibitemShut {NoStop}%
\bibitem [{\citenamefont {Zhou}\ \emph {et~al.}(2021)\citenamefont {Zhou},
  \citenamefont {Zhuang}, \citenamefont {Mattina},\ and\ \citenamefont
  {Whatmough}}]{zhou2021strong}%
  \BibitemOpen
  \bibfield  {author} {\bibinfo {author} {\bibfnamefont {C.}~\bibnamefont
  {Zhou}}, \bibinfo {author} {\bibfnamefont {Q.}~\bibnamefont {Zhuang}},
  \bibinfo {author} {\bibfnamefont {M.}~\bibnamefont {Mattina}},\ and\ \bibinfo
  {author} {\bibfnamefont {P.~N.}\ \bibnamefont {Whatmough}},\ }\bibfield
  {title} {\bibinfo {title} {Strong data processing inequality in neural
  networks with noisy neurons and its implications},\ }in\ \href@noop {} {\emph
  {\bibinfo {booktitle} {2021 IEEE International Symposium on Information
  Theory (ISIT)}}}\ (\bibinfo {organization} {IEEE},\ \bibinfo {year} {2021})\
  pp.\ \bibinfo {pages} {1170--1175}\BibitemShut {NoStop}%
\bibitem [{\citenamefont {Oord}\ \emph {et~al.}(2018)\citenamefont {Oord},
  \citenamefont {Li},\ and\ \citenamefont {Vinyals}}]{oord2018representation}%
  \BibitemOpen
  \bibfield  {author} {\bibinfo {author} {\bibfnamefont {A.~v.~d.}\
  \bibnamefont {Oord}}, \bibinfo {author} {\bibfnamefont {Y.}~\bibnamefont
  {Li}},\ and\ \bibinfo {author} {\bibfnamefont {O.}~\bibnamefont {Vinyals}},\
  }\bibfield  {title} {\bibinfo {title} {Representation learning with
  contrastive predictive coding},\ }\href@noop {} {\bibfield  {journal}
  {\bibinfo  {journal} {arXiv preprint arXiv:1807.03748}\ } (\bibinfo {year}
  {2018})}\BibitemShut {NoStop}%
\bibitem [{\citenamefont {Henaff}(2020)}]{henaff2020data}%
  \BibitemOpen
  \bibfield  {author} {\bibinfo {author} {\bibfnamefont {O.}~\bibnamefont
  {Henaff}},\ }\bibfield  {title} {\bibinfo {title} {Data-efficient image
  recognition with contrastive predictive coding},\ }in\ \href@noop {} {\emph
  {\bibinfo {booktitle} {International conference on machine learning}}}\
  (\bibinfo {organization} {PMLR},\ \bibinfo {year} {2020})\ pp.\ \bibinfo
  {pages} {4182--4192}\BibitemShut {NoStop}%
\bibitem [{\citenamefont {Tian}\ \emph {et~al.}(2020)\citenamefont {Tian},
  \citenamefont {Krishnan},\ and\ \citenamefont {Isola}}]{tian2020contrastive}%
  \BibitemOpen
  \bibfield  {author} {\bibinfo {author} {\bibfnamefont {Y.}~\bibnamefont
  {Tian}}, \bibinfo {author} {\bibfnamefont {D.}~\bibnamefont {Krishnan}},\
  and\ \bibinfo {author} {\bibfnamefont {P.}~\bibnamefont {Isola}},\ }\bibfield
   {title} {\bibinfo {title} {Contrastive multiview coding},\ }in\ \href@noop
  {} {\emph {\bibinfo {booktitle} {European conference on computer vision}}}\
  (\bibinfo {organization} {Springer},\ \bibinfo {year} {2020})\ pp.\ \bibinfo
  {pages} {776--794}\BibitemShut {NoStop}%
\bibitem [{\citenamefont {Hjelm}\ \emph {et~al.}(2018)\citenamefont {Hjelm},
  \citenamefont {Fedorov}, \citenamefont {Lavoie-Marchildon}, \citenamefont
  {Grewal}, \citenamefont {Bachman}, \citenamefont {Trischler},\ and\
  \citenamefont {Bengio}}]{hjelm2018learning}%
  \BibitemOpen
  \bibfield  {author} {\bibinfo {author} {\bibfnamefont {R.~D.}\ \bibnamefont
  {Hjelm}}, \bibinfo {author} {\bibfnamefont {A.}~\bibnamefont {Fedorov}},
  \bibinfo {author} {\bibfnamefont {S.}~\bibnamefont {Lavoie-Marchildon}},
  \bibinfo {author} {\bibfnamefont {K.}~\bibnamefont {Grewal}}, \bibinfo
  {author} {\bibfnamefont {P.}~\bibnamefont {Bachman}}, \bibinfo {author}
  {\bibfnamefont {A.}~\bibnamefont {Trischler}},\ and\ \bibinfo {author}
  {\bibfnamefont {Y.}~\bibnamefont {Bengio}},\ }\bibfield  {title} {\bibinfo
  {title} {Learning deep representations by mutual information estimation and
  maximization},\ }\href@noop {} {\bibfield  {journal} {\bibinfo  {journal}
  {arXiv preprint arXiv:1808.06670}\ } (\bibinfo {year} {2018})}\BibitemShut
  {NoStop}%
\bibitem [{\citenamefont {Poole}\ \emph {et~al.}(2019)\citenamefont {Poole},
  \citenamefont {Ozair}, \citenamefont {Van Den~Oord}, \citenamefont {Alemi},\
  and\ \citenamefont {Tucker}}]{poole2019variational}%
  \BibitemOpen
  \bibfield  {author} {\bibinfo {author} {\bibfnamefont {B.}~\bibnamefont
  {Poole}}, \bibinfo {author} {\bibfnamefont {S.}~\bibnamefont {Ozair}},
  \bibinfo {author} {\bibfnamefont {A.}~\bibnamefont {Van Den~Oord}}, \bibinfo
  {author} {\bibfnamefont {A.}~\bibnamefont {Alemi}},\ and\ \bibinfo {author}
  {\bibfnamefont {G.}~\bibnamefont {Tucker}},\ }\bibfield  {title} {\bibinfo
  {title} {On variational bounds of mutual information},\ }in\ \href@noop {}
  {\emph {\bibinfo {booktitle} {International Conference on Machine
  Learning}}}\ (\bibinfo {organization} {PMLR},\ \bibinfo {year} {2019})\ pp.\
  \bibinfo {pages} {5171--5180}\BibitemShut {NoStop}%
\bibitem [{\citenamefont {McAllester}\ and\ \citenamefont
  {Stratos}(2020)}]{mcallester2020formal}%
  \BibitemOpen
  \bibfield  {author} {\bibinfo {author} {\bibfnamefont {D.}~\bibnamefont
  {McAllester}}\ and\ \bibinfo {author} {\bibfnamefont {K.}~\bibnamefont
  {Stratos}},\ }\bibfield  {title} {\bibinfo {title} {Formal limitations on the
  measurement of mutual information},\ }in\ \href@noop {} {\emph {\bibinfo
  {booktitle} {International Conference on Artificial Intelligence and
  Statistics}}}\ (\bibinfo {organization} {PMLR},\ \bibinfo {year} {2020})\
  pp.\ \bibinfo {pages} {875--884}\BibitemShut {NoStop}%
\bibitem [{\citenamefont {Tschannen}\ \emph {et~al.}(2019)\citenamefont
  {Tschannen}, \citenamefont {Djolonga}, \citenamefont {Rubenstein},
  \citenamefont {Gelly},\ and\ \citenamefont {Lucic}}]{tschannen2019mutual}%
  \BibitemOpen
  \bibfield  {author} {\bibinfo {author} {\bibfnamefont {M.}~\bibnamefont
  {Tschannen}}, \bibinfo {author} {\bibfnamefont {J.}~\bibnamefont {Djolonga}},
  \bibinfo {author} {\bibfnamefont {P.~K.}\ \bibnamefont {Rubenstein}},
  \bibinfo {author} {\bibfnamefont {S.}~\bibnamefont {Gelly}},\ and\ \bibinfo
  {author} {\bibfnamefont {M.}~\bibnamefont {Lucic}},\ }\bibfield  {title}
  {\bibinfo {title} {On mutual information maximization for representation
  learning},\ }\href@noop {} {\bibfield  {journal} {\bibinfo  {journal} {arXiv
  preprint arXiv:1907.13625}\ } (\bibinfo {year} {2019})}\BibitemShut {NoStop}%
\bibitem [{\citenamefont {Becker}\ and\ \citenamefont
  {Hinton}(1992)}]{becker1992self}%
  \BibitemOpen
  \bibfield  {author} {\bibinfo {author} {\bibfnamefont {S.}~\bibnamefont
  {Becker}}\ and\ \bibinfo {author} {\bibfnamefont {G.~E.}\ \bibnamefont
  {Hinton}},\ }\bibfield  {title} {\bibinfo {title} {Self-organizing neural
  network that discovers surfaces in random-dot stereograms},\ }\href@noop {}
  {\bibfield  {journal} {\bibinfo  {journal} {Nature}\ }\textbf {\bibinfo
  {volume} {355}},\ \bibinfo {pages} {161} (\bibinfo {year}
  {1992})}\BibitemShut {NoStop}%
\bibitem [{\citenamefont {Linsker}(1988)}]{linsker1988self}%
  \BibitemOpen
  \bibfield  {author} {\bibinfo {author} {\bibfnamefont {R.}~\bibnamefont
  {Linsker}},\ }\bibfield  {title} {\bibinfo {title} {Self-organization in a
  perceptual network},\ }\href@noop {} {\bibfield  {journal} {\bibinfo
  {journal} {Computer}\ }\textbf {\bibinfo {volume} {21}},\ \bibinfo {pages}
  {105} (\bibinfo {year} {1988})}\BibitemShut {NoStop}%
\bibitem [{\citenamefont {Ge}(2018)}]{ge2018deep}%
  \BibitemOpen
  \bibfield  {author} {\bibinfo {author} {\bibfnamefont {W.}~\bibnamefont
  {Ge}},\ }\bibfield  {title} {\bibinfo {title} {Deep metric learning with
  hierarchical triplet loss},\ }in\ \href@noop {} {\emph {\bibinfo {booktitle}
  {Proceedings of the European Conference on Computer Vision (ECCV)}}}\
  (\bibinfo {year} {2018})\ pp.\ \bibinfo {pages} {269--285}\BibitemShut
  {NoStop}%
\bibitem [{\citenamefont {Yu}\ and\ \citenamefont {Tao}(2019)}]{yu2019deep}%
  \BibitemOpen
  \bibfield  {author} {\bibinfo {author} {\bibfnamefont {B.}~\bibnamefont
  {Yu}}\ and\ \bibinfo {author} {\bibfnamefont {D.}~\bibnamefont {Tao}},\
  }\bibfield  {title} {\bibinfo {title} {Deep metric learning with tuplet
  margin loss},\ }in\ \href@noop {} {\emph {\bibinfo {booktitle} {Proceedings
  of the IEEE/CVF International Conference on Computer Vision}}}\ (\bibinfo
  {year} {2019})\ pp.\ \bibinfo {pages} {6490--6499}\BibitemShut {NoStop}%
\bibitem [{\citenamefont {Sohn}(2016)}]{sohn2016improved}%
  \BibitemOpen
  \bibfield  {author} {\bibinfo {author} {\bibfnamefont {K.}~\bibnamefont
  {Sohn}},\ }\bibfield  {title} {\bibinfo {title} {Improved deep metric
  learning with multi-class n-pair loss objective},\ }\href@noop {} {\bibfield
  {journal} {\bibinfo  {journal} {Advances in neural information processing
  systems}\ }\textbf {\bibinfo {volume} {29}} (\bibinfo {year}
  {2016})}\BibitemShut {NoStop}%
\bibitem [{\citenamefont {Li}\ and\ \citenamefont
  {Sompolinsky}(2021)}]{li2021statistical}%
  \BibitemOpen
  \bibfield  {author} {\bibinfo {author} {\bibfnamefont {Q.}~\bibnamefont
  {Li}}\ and\ \bibinfo {author} {\bibfnamefont {H.}~\bibnamefont
  {Sompolinsky}},\ }\bibfield  {title} {\bibinfo {title} {Statistical mechanics
  of deep linear neural networks: The backpropagating kernel renormalization},\
  }\href@noop {} {\bibfield  {journal} {\bibinfo  {journal} {Physical Review
  X}\ }\textbf {\bibinfo {volume} {11}},\ \bibinfo {pages} {031059} (\bibinfo
  {year} {2021})}\BibitemShut {NoStop}%
\bibitem [{\citenamefont {Bahri}\ \emph {et~al.}(2020)\citenamefont {Bahri},
  \citenamefont {Kadmon}, \citenamefont {Pennington}, \citenamefont
  {Schoenholz}, \citenamefont {Sohl-Dickstein},\ and\ \citenamefont
  {Ganguli}}]{bahri2020statistical}%
  \BibitemOpen
  \bibfield  {author} {\bibinfo {author} {\bibfnamefont {Y.}~\bibnamefont
  {Bahri}}, \bibinfo {author} {\bibfnamefont {J.}~\bibnamefont {Kadmon}},
  \bibinfo {author} {\bibfnamefont {J.}~\bibnamefont {Pennington}}, \bibinfo
  {author} {\bibfnamefont {S.~S.}\ \bibnamefont {Schoenholz}}, \bibinfo
  {author} {\bibfnamefont {J.}~\bibnamefont {Sohl-Dickstein}},\ and\ \bibinfo
  {author} {\bibfnamefont {S.}~\bibnamefont {Ganguli}},\ }\bibfield  {title}
  {\bibinfo {title} {Statistical mechanics of deep learning},\ }\href@noop {}
  {\bibfield  {journal} {\bibinfo  {journal} {Annual Review of Condensed Matter
  Physics}\ }\textbf {\bibinfo {volume} {11}} (\bibinfo {year}
  {2020})}\BibitemShut {NoStop}%
\bibitem [{\citenamefont {Jacot}\ \emph {et~al.}(2018)\citenamefont {Jacot},
  \citenamefont {Gabriel},\ and\ \citenamefont {Hongler}}]{jacot2018neural}%
  \BibitemOpen
  \bibfield  {author} {\bibinfo {author} {\bibfnamefont {A.}~\bibnamefont
  {Jacot}}, \bibinfo {author} {\bibfnamefont {F.}~\bibnamefont {Gabriel}},\
  and\ \bibinfo {author} {\bibfnamefont {C.}~\bibnamefont {Hongler}},\
  }\bibfield  {title} {\bibinfo {title} {Neural tangent kernel: Convergence and
  generalization in neural networks},\ }\href@noop {} {\bibfield  {journal}
  {\bibinfo  {journal} {Advances in neural information processing systems}\
  }\textbf {\bibinfo {volume} {31}} (\bibinfo {year} {2018})}\BibitemShut
  {NoStop}%
\bibitem [{\citenamefont {Golikov}\ \emph {et~al.}(2022)\citenamefont
  {Golikov}, \citenamefont {Pokonechnyy},\ and\ \citenamefont
  {Korviakov}}]{golikov2022neural}%
  \BibitemOpen
  \bibfield  {author} {\bibinfo {author} {\bibfnamefont {E.}~\bibnamefont
  {Golikov}}, \bibinfo {author} {\bibfnamefont {E.}~\bibnamefont
  {Pokonechnyy}},\ and\ \bibinfo {author} {\bibfnamefont {V.}~\bibnamefont
  {Korviakov}},\ }\bibfield  {title} {\bibinfo {title} {Neural tangent kernel:
  A survey},\ }\href@noop {} {\bibfield  {journal} {\bibinfo  {journal} {arXiv
  preprint arXiv:2208.13614}\ } (\bibinfo {year} {2022})}\BibitemShut {NoStop}%
\bibitem [{\citenamefont {Tishby}\ \emph {et~al.}(2000)\citenamefont {Tishby},
  \citenamefont {Pereira},\ and\ \citenamefont
  {Bialek}}]{tishby2000information}%
  \BibitemOpen
  \bibfield  {author} {\bibinfo {author} {\bibfnamefont {N.}~\bibnamefont
  {Tishby}}, \bibinfo {author} {\bibfnamefont {F.~C.}\ \bibnamefont
  {Pereira}},\ and\ \bibinfo {author} {\bibfnamefont {W.}~\bibnamefont
  {Bialek}},\ }\bibfield  {title} {\bibinfo {title} {The information bottleneck
  method},\ }\href@noop {} {\bibfield  {journal} {\bibinfo  {journal} {arXiv
  preprint physics/0004057}\ } (\bibinfo {year} {2000})}\BibitemShut {NoStop}%
\bibitem [{\citenamefont {Alemi}\ \emph {et~al.}(2016)\citenamefont {Alemi},
  \citenamefont {Fischer}, \citenamefont {Dillon},\ and\ \citenamefont
  {Murphy}}]{alemi2016deep}%
  \BibitemOpen
  \bibfield  {author} {\bibinfo {author} {\bibfnamefont {A.~A.}\ \bibnamefont
  {Alemi}}, \bibinfo {author} {\bibfnamefont {I.}~\bibnamefont {Fischer}},
  \bibinfo {author} {\bibfnamefont {J.~V.}\ \bibnamefont {Dillon}},\ and\
  \bibinfo {author} {\bibfnamefont {K.}~\bibnamefont {Murphy}},\ }\bibfield
  {title} {\bibinfo {title} {Deep variational information bottleneck},\
  }\href@noop {} {\bibfield  {journal} {\bibinfo  {journal} {arXiv preprint
  arXiv:1612.00410}\ } (\bibinfo {year} {2016})}\BibitemShut {NoStop}%
\bibitem [{\citenamefont {Higgins}\ \emph {et~al.}(2017)\citenamefont
  {Higgins}, \citenamefont {Matthey}, \citenamefont {Pal}, \citenamefont
  {Burgess}, \citenamefont {Glorot}, \citenamefont {Botvinick}, \citenamefont
  {Mohamed},\ and\ \citenamefont {Lerchner}}]{higgins2017betavae}%
  \BibitemOpen
  \bibfield  {author} {\bibinfo {author} {\bibfnamefont {I.}~\bibnamefont
  {Higgins}}, \bibinfo {author} {\bibfnamefont {L.}~\bibnamefont {Matthey}},
  \bibinfo {author} {\bibfnamefont {A.}~\bibnamefont {Pal}}, \bibinfo {author}
  {\bibfnamefont {C.}~\bibnamefont {Burgess}}, \bibinfo {author} {\bibfnamefont
  {X.}~\bibnamefont {Glorot}}, \bibinfo {author} {\bibfnamefont
  {M.}~\bibnamefont {Botvinick}}, \bibinfo {author} {\bibfnamefont
  {S.}~\bibnamefont {Mohamed}},\ and\ \bibinfo {author} {\bibfnamefont
  {A.}~\bibnamefont {Lerchner}},\ }\bibfield  {title} {\bibinfo {title}
  {beta-{VAE}: Learning basic visual concepts with a constrained variational
  framework},\ }in\ \href {https://openreview.net/forum?id=Sy2fzU9gl} {\emph
  {\bibinfo {booktitle} {International Conference on Learning
  Representations}}}\ (\bibinfo {year} {2017})\BibitemShut {NoStop}%
\bibitem [{\citenamefont {Lee}\ \emph {et~al.}(2017)\citenamefont {Lee},
  \citenamefont {Bahri}, \citenamefont {Novak}, \citenamefont {Schoenholz},
  \citenamefont {Pennington},\ and\ \citenamefont
  {Sohl-Dickstein}}]{lee2017deep}%
  \BibitemOpen
  \bibfield  {author} {\bibinfo {author} {\bibfnamefont {J.}~\bibnamefont
  {Lee}}, \bibinfo {author} {\bibfnamefont {Y.}~\bibnamefont {Bahri}}, \bibinfo
  {author} {\bibfnamefont {R.}~\bibnamefont {Novak}}, \bibinfo {author}
  {\bibfnamefont {S.~S.}\ \bibnamefont {Schoenholz}}, \bibinfo {author}
  {\bibfnamefont {J.}~\bibnamefont {Pennington}},\ and\ \bibinfo {author}
  {\bibfnamefont {J.}~\bibnamefont {Sohl-Dickstein}},\ }\bibfield  {title}
  {\bibinfo {title} {Deep neural networks as gaussian processes},\ }\href@noop
  {} {\bibfield  {journal} {\bibinfo  {journal} {arXiv preprint
  arXiv:1711.00165}\ } (\bibinfo {year} {2017})}\BibitemShut {NoStop}%
\bibitem [{\citenamefont {Matthews}\ \emph {et~al.}(2018)\citenamefont
  {Matthews}, \citenamefont {Rowland}, \citenamefont {Hron}, \citenamefont
  {Turner},\ and\ \citenamefont {Ghahramani}}]{matthews2018gaussian}%
  \BibitemOpen
  \bibfield  {author} {\bibinfo {author} {\bibfnamefont {A.~G. d.~G.}\
  \bibnamefont {Matthews}}, \bibinfo {author} {\bibfnamefont {M.}~\bibnamefont
  {Rowland}}, \bibinfo {author} {\bibfnamefont {J.}~\bibnamefont {Hron}},
  \bibinfo {author} {\bibfnamefont {R.~E.}\ \bibnamefont {Turner}},\ and\
  \bibinfo {author} {\bibfnamefont {Z.}~\bibnamefont {Ghahramani}},\ }\bibfield
   {title} {\bibinfo {title} {Gaussian process behaviour in wide deep neural
  networks},\ }\href@noop {} {\bibfield  {journal} {\bibinfo  {journal} {arXiv
  preprint arXiv:1804.11271}\ } (\bibinfo {year} {2018})}\BibitemShut {NoStop}%
\bibitem [{\citenamefont {Garriga-Alonso}\ \emph {et~al.}(2018)\citenamefont
  {Garriga-Alonso}, \citenamefont {Rasmussen},\ and\ \citenamefont
  {Aitchison}}]{garriga2018deep}%
  \BibitemOpen
  \bibfield  {author} {\bibinfo {author} {\bibfnamefont {A.}~\bibnamefont
  {Garriga-Alonso}}, \bibinfo {author} {\bibfnamefont {C.~E.}\ \bibnamefont
  {Rasmussen}},\ and\ \bibinfo {author} {\bibfnamefont {L.}~\bibnamefont
  {Aitchison}},\ }\bibfield  {title} {\bibinfo {title} {Deep convolutional
  networks as shallow gaussian processes},\ }\href@noop {} {\bibfield
  {journal} {\bibinfo  {journal} {arXiv preprint arXiv:1808.05587}\ } (\bibinfo
  {year} {2018})}\BibitemShut {NoStop}%
\bibitem [{\citenamefont {Lee}\ \emph {et~al.}(2019)\citenamefont {Lee},
  \citenamefont {Xiao}, \citenamefont {Schoenholz}, \citenamefont {Bahri},
  \citenamefont {Novak}, \citenamefont {Sohl-Dickstein},\ and\ \citenamefont
  {Pennington}}]{lee2019wide}%
  \BibitemOpen
  \bibfield  {author} {\bibinfo {author} {\bibfnamefont {J.}~\bibnamefont
  {Lee}}, \bibinfo {author} {\bibfnamefont {L.}~\bibnamefont {Xiao}}, \bibinfo
  {author} {\bibfnamefont {S.}~\bibnamefont {Schoenholz}}, \bibinfo {author}
  {\bibfnamefont {Y.}~\bibnamefont {Bahri}}, \bibinfo {author} {\bibfnamefont
  {R.}~\bibnamefont {Novak}}, \bibinfo {author} {\bibfnamefont
  {J.}~\bibnamefont {Sohl-Dickstein}},\ and\ \bibinfo {author} {\bibfnamefont
  {J.}~\bibnamefont {Pennington}},\ }\bibfield  {title} {\bibinfo {title} {Wide
  neural networks of any depth evolve as linear models under gradient
  descent},\ }\href@noop {} {\bibfield  {journal} {\bibinfo  {journal}
  {Advances in neural information processing systems}\ }\textbf {\bibinfo
  {volume} {32}} (\bibinfo {year} {2019})}\BibitemShut {NoStop}%
\bibitem [{\citenamefont {Chizat}\ \emph {et~al.}(2019)\citenamefont {Chizat},
  \citenamefont {Oyallon},\ and\ \citenamefont {Bach}}]{chizat2019lazy}%
  \BibitemOpen
  \bibfield  {author} {\bibinfo {author} {\bibfnamefont {L.}~\bibnamefont
  {Chizat}}, \bibinfo {author} {\bibfnamefont {E.}~\bibnamefont {Oyallon}},\
  and\ \bibinfo {author} {\bibfnamefont {F.}~\bibnamefont {Bach}},\ }\bibfield
  {title} {\bibinfo {title} {On lazy training in differentiable programming},\
  }\href@noop {} {\bibfield  {journal} {\bibinfo  {journal} {Advances in Neural
  Information Processing Systems}\ }\textbf {\bibinfo {volume} {32}} (\bibinfo
  {year} {2019})}\BibitemShut {NoStop}%
\bibitem [{\citenamefont {Novak}\ \emph {et~al.}(2018)\citenamefont {Novak},
  \citenamefont {Xiao}, \citenamefont {Lee}, \citenamefont {Bahri},
  \citenamefont {Yang}, \citenamefont {Hron}, \citenamefont {Abolafia},
  \citenamefont {Pennington},\ and\ \citenamefont
  {Sohl-Dickstein}}]{novak2018bayesian}%
  \BibitemOpen
  \bibfield  {author} {\bibinfo {author} {\bibfnamefont {R.}~\bibnamefont
  {Novak}}, \bibinfo {author} {\bibfnamefont {L.}~\bibnamefont {Xiao}},
  \bibinfo {author} {\bibfnamefont {J.}~\bibnamefont {Lee}}, \bibinfo {author}
  {\bibfnamefont {Y.}~\bibnamefont {Bahri}}, \bibinfo {author} {\bibfnamefont
  {G.}~\bibnamefont {Yang}}, \bibinfo {author} {\bibfnamefont {J.}~\bibnamefont
  {Hron}}, \bibinfo {author} {\bibfnamefont {D.~A.}\ \bibnamefont {Abolafia}},
  \bibinfo {author} {\bibfnamefont {J.}~\bibnamefont {Pennington}},\ and\
  \bibinfo {author} {\bibfnamefont {J.}~\bibnamefont {Sohl-Dickstein}},\
  }\bibfield  {title} {\bibinfo {title} {Bayesian deep convolutional networks
  with many channels are gaussian processes},\ }\href@noop {} {\bibfield
  {journal} {\bibinfo  {journal} {arXiv preprint arXiv:1810.05148}\ } (\bibinfo
  {year} {2018})}\BibitemShut {NoStop}%
\bibitem [{\citenamefont {Yang}(2019)}]{yang2019wide}%
  \BibitemOpen
  \bibfield  {author} {\bibinfo {author} {\bibfnamefont {G.}~\bibnamefont
  {Yang}},\ }\bibfield  {title} {\bibinfo {title} {Wide feedforward or
  recurrent neural networks of any architecture are gaussian processes},\
  }\href@noop {} {\bibfield  {journal} {\bibinfo  {journal} {Advances in Neural
  Information Processing Systems}\ }\textbf {\bibinfo {volume} {32}} (\bibinfo
  {year} {2019})}\BibitemShut {NoStop}%
\bibitem [{\citenamefont {Sirignano}\ and\ \citenamefont
  {Spiliopoulos}(2022)}]{sirignano2022mean}%
  \BibitemOpen
  \bibfield  {author} {\bibinfo {author} {\bibfnamefont {J.}~\bibnamefont
  {Sirignano}}\ and\ \bibinfo {author} {\bibfnamefont {K.}~\bibnamefont
  {Spiliopoulos}},\ }\bibfield  {title} {\bibinfo {title} {Mean field analysis
  of deep neural networks},\ }\href@noop {} {\bibfield  {journal} {\bibinfo
  {journal} {Mathematics of Operations Research}\ }\textbf {\bibinfo {volume}
  {47}},\ \bibinfo {pages} {120} (\bibinfo {year} {2022})}\BibitemShut
  {NoStop}%
\bibitem [{\citenamefont {Schoenholz}\ \emph {et~al.}(2016)\citenamefont
  {Schoenholz}, \citenamefont {Gilmer}, \citenamefont {Ganguli},\ and\
  \citenamefont {Sohl-Dickstein}}]{schoenholz2016deep}%
  \BibitemOpen
  \bibfield  {author} {\bibinfo {author} {\bibfnamefont {S.~S.}\ \bibnamefont
  {Schoenholz}}, \bibinfo {author} {\bibfnamefont {J.}~\bibnamefont {Gilmer}},
  \bibinfo {author} {\bibfnamefont {S.}~\bibnamefont {Ganguli}},\ and\ \bibinfo
  {author} {\bibfnamefont {J.}~\bibnamefont {Sohl-Dickstein}},\ }\bibfield
  {title} {\bibinfo {title} {Deep information propagation},\ }\href@noop {}
  {\bibfield  {journal} {\bibinfo  {journal} {arXiv preprint arXiv:1611.01232}\
  } (\bibinfo {year} {2016})}\BibitemShut {NoStop}%
\bibitem [{\citenamefont {Pennington}\ \emph {et~al.}(2017)\citenamefont
  {Pennington}, \citenamefont {Schoenholz},\ and\ \citenamefont
  {Ganguli}}]{pennington2017resurrecting}%
  \BibitemOpen
  \bibfield  {author} {\bibinfo {author} {\bibfnamefont {J.}~\bibnamefont
  {Pennington}}, \bibinfo {author} {\bibfnamefont {S.}~\bibnamefont
  {Schoenholz}},\ and\ \bibinfo {author} {\bibfnamefont {S.}~\bibnamefont
  {Ganguli}},\ }\bibfield  {title} {\bibinfo {title} {Resurrecting the sigmoid
  in deep learning through dynamical isometry: theory and practice},\
  }\href@noop {} {\bibfield  {journal} {\bibinfo  {journal} {Advances in neural
  information processing systems}\ }\textbf {\bibinfo {volume} {30}} (\bibinfo
  {year} {2017})}\BibitemShut {NoStop}%
\bibitem [{\citenamefont {Yang}\ and\ \citenamefont
  {Schoenholz}(2017)}]{yang2017mean}%
  \BibitemOpen
  \bibfield  {author} {\bibinfo {author} {\bibfnamefont {G.}~\bibnamefont
  {Yang}}\ and\ \bibinfo {author} {\bibfnamefont {S.}~\bibnamefont
  {Schoenholz}},\ }\bibfield  {title} {\bibinfo {title} {Mean field residual
  networks: On the edge of chaos},\ }\href@noop {} {\bibfield  {journal}
  {\bibinfo  {journal} {Advances in neural information processing systems}\
  }\textbf {\bibinfo {volume} {30}} (\bibinfo {year} {2017})}\BibitemShut
  {NoStop}%
\bibitem [{\citenamefont {Pennington}\ \emph {et~al.}(2018)\citenamefont
  {Pennington}, \citenamefont {Schoenholz},\ and\ \citenamefont
  {Ganguli}}]{pennington2018emergence}%
  \BibitemOpen
  \bibfield  {author} {\bibinfo {author} {\bibfnamefont {J.}~\bibnamefont
  {Pennington}}, \bibinfo {author} {\bibfnamefont {S.}~\bibnamefont
  {Schoenholz}},\ and\ \bibinfo {author} {\bibfnamefont {S.}~\bibnamefont
  {Ganguli}},\ }\bibfield  {title} {\bibinfo {title} {The emergence of spectral
  universality in deep networks},\ }in\ \href@noop {} {\emph {\bibinfo
  {booktitle} {International Conference on Artificial Intelligence and
  Statistics}}}\ (\bibinfo {organization} {PMLR},\ \bibinfo {year} {2018})\
  pp.\ \bibinfo {pages} {1924--1932}\BibitemShut {NoStop}%
\bibitem [{\citenamefont {Chen}\ \emph {et~al.}(2018)\citenamefont {Chen},
  \citenamefont {Pennington},\ and\ \citenamefont
  {Schoenholz}}]{chen2018dynamical}%
  \BibitemOpen
  \bibfield  {author} {\bibinfo {author} {\bibfnamefont {M.}~\bibnamefont
  {Chen}}, \bibinfo {author} {\bibfnamefont {J.}~\bibnamefont {Pennington}},\
  and\ \bibinfo {author} {\bibfnamefont {S.}~\bibnamefont {Schoenholz}},\
  }\bibfield  {title} {\bibinfo {title} {Dynamical isometry and a mean field
  theory of rnns: Gating enables signal propagation in recurrent neural
  networks},\ }in\ \href@noop {} {\emph {\bibinfo {booktitle} {International
  Conference on Machine Learning}}}\ (\bibinfo {organization} {PMLR},\ \bibinfo
  {year} {2018})\ pp.\ \bibinfo {pages} {873--882}\BibitemShut {NoStop}%
\bibitem [{\citenamefont {Xiao}\ \emph {et~al.}(2018)\citenamefont {Xiao},
  \citenamefont {Bahri}, \citenamefont {Sohl-Dickstein}, \citenamefont
  {Schoenholz},\ and\ \citenamefont {Pennington}}]{xiao2018dynamical}%
  \BibitemOpen
  \bibfield  {author} {\bibinfo {author} {\bibfnamefont {L.}~\bibnamefont
  {Xiao}}, \bibinfo {author} {\bibfnamefont {Y.}~\bibnamefont {Bahri}},
  \bibinfo {author} {\bibfnamefont {J.}~\bibnamefont {Sohl-Dickstein}},
  \bibinfo {author} {\bibfnamefont {S.}~\bibnamefont {Schoenholz}},\ and\
  \bibinfo {author} {\bibfnamefont {J.}~\bibnamefont {Pennington}},\ }\bibfield
   {title} {\bibinfo {title} {Dynamical isometry and a mean field theory of
  cnns: How to train 10,000-layer vanilla convolutional neural networks},\ }in\
  \href@noop {} {\emph {\bibinfo {booktitle} {International Conference on
  Machine Learning}}}\ (\bibinfo {organization} {PMLR},\ \bibinfo {year}
  {2018})\ pp.\ \bibinfo {pages} {5393--5402}\BibitemShut {NoStop}%
\bibitem [{\citenamefont {Berger}(2003)}]{berger2003rate}%
  \BibitemOpen
  \bibfield  {author} {\bibinfo {author} {\bibfnamefont {T.}~\bibnamefont
  {Berger}},\ }\bibfield  {title} {\bibinfo {title} {Rate-distortion theory},\
  }\href@noop {} {\bibfield  {journal} {\bibinfo  {journal} {Wiley Encyclopedia
  of Telecommunications}\ } (\bibinfo {year} {2003})}\BibitemShut {NoStop}%
\bibitem [{\citenamefont {Kleven}(2021)}]{kleven2021sufficient}%
  \BibitemOpen
  \bibfield  {author} {\bibinfo {author} {\bibfnamefont {H.~J.}\ \bibnamefont
  {Kleven}},\ }\bibfield  {title} {\bibinfo {title} {Sufficient statistics
  revisited},\ }\href@noop {} {\bibfield  {journal} {\bibinfo  {journal}
  {Annual Review of Economics}\ }\textbf {\bibinfo {volume} {13}},\ \bibinfo
  {pages} {515} (\bibinfo {year} {2021})}\BibitemShut {NoStop}%
\bibitem [{\citenamefont {Nguyen}\ \emph {et~al.}(2022)\citenamefont {Nguyen},
  \citenamefont {Trahay}, \citenamefont {Domke}, \citenamefont {Drozd},
  \citenamefont {Vatai}, \citenamefont {Liao}, \citenamefont {Wahib},\ and\
  \citenamefont {Gerofi}}]{nguyen2022globally}%
  \BibitemOpen
  \bibfield  {author} {\bibinfo {author} {\bibfnamefont {T.~T.}\ \bibnamefont
  {Nguyen}}, \bibinfo {author} {\bibfnamefont {F.}~\bibnamefont {Trahay}},
  \bibinfo {author} {\bibfnamefont {J.}~\bibnamefont {Domke}}, \bibinfo
  {author} {\bibfnamefont {A.}~\bibnamefont {Drozd}}, \bibinfo {author}
  {\bibfnamefont {E.}~\bibnamefont {Vatai}}, \bibinfo {author} {\bibfnamefont
  {J.}~\bibnamefont {Liao}}, \bibinfo {author} {\bibfnamefont {M.}~\bibnamefont
  {Wahib}},\ and\ \bibinfo {author} {\bibfnamefont {B.}~\bibnamefont
  {Gerofi}},\ }\bibfield  {title} {\bibinfo {title} {Why globally re-shuffle?
  revisiting data shuffling in large scale deep learning},\ }in\ \href@noop {}
  {\emph {\bibinfo {booktitle} {2022 IEEE International Parallel and
  Distributed Processing Symposium (IPDPS)}}}\ (\bibinfo {organization}
  {IEEE},\ \bibinfo {year} {2022})\ pp.\ \bibinfo {pages}
  {1085--1096}\BibitemShut {NoStop}%
\bibitem [{\citenamefont {Summers}\ and\ \citenamefont
  {Dinneen}(2021)}]{summers2021nondeterminism}%
  \BibitemOpen
  \bibfield  {author} {\bibinfo {author} {\bibfnamefont {C.}~\bibnamefont
  {Summers}}\ and\ \bibinfo {author} {\bibfnamefont {M.~J.}\ \bibnamefont
  {Dinneen}},\ }\bibfield  {title} {\bibinfo {title} {Nondeterminism and
  instability in neural network optimization},\ }in\ \href@noop {} {\emph
  {\bibinfo {booktitle} {International Conference on Machine Learning}}}\
  (\bibinfo {organization} {PMLR},\ \bibinfo {year} {2021})\ pp.\ \bibinfo
  {pages} {9913--9922}\BibitemShut {NoStop}%
\bibitem [{\citenamefont {Mei}\ \emph {et~al.}(2019)\citenamefont {Mei},
  \citenamefont {Misiakiewicz},\ and\ \citenamefont {Montanari}}]{mei2019mean}%
  \BibitemOpen
  \bibfield  {author} {\bibinfo {author} {\bibfnamefont {S.}~\bibnamefont
  {Mei}}, \bibinfo {author} {\bibfnamefont {T.}~\bibnamefont {Misiakiewicz}},\
  and\ \bibinfo {author} {\bibfnamefont {A.}~\bibnamefont {Montanari}},\
  }\bibfield  {title} {\bibinfo {title} {Mean-field theory of two-layers neural
  networks: dimension-free bounds and kernel limit},\ }in\ \href@noop {} {\emph
  {\bibinfo {booktitle} {Conference on Learning Theory}}}\ (\bibinfo
  {organization} {PMLR},\ \bibinfo {year} {2019})\ pp.\ \bibinfo {pages}
  {2388--2464}\BibitemShut {NoStop}%
\bibitem [{\citenamefont {Nguyen}(2019)}]{nguyen2019mean}%
  \BibitemOpen
  \bibfield  {author} {\bibinfo {author} {\bibfnamefont {P.-M.}\ \bibnamefont
  {Nguyen}},\ }\bibfield  {title} {\bibinfo {title} {Mean field limit of the
  learning dynamics of multilayer neural networks},\ }\href@noop {} {\bibfield
  {journal} {\bibinfo  {journal} {arXiv preprint arXiv:1902.02880}\ } (\bibinfo
  {year} {2019})}\BibitemShut {NoStop}%
\bibitem [{\citenamefont {Poole}\ \emph {et~al.}(2016)\citenamefont {Poole},
  \citenamefont {Lahiri}, \citenamefont {Raghu}, \citenamefont
  {Sohl-Dickstein},\ and\ \citenamefont {Ganguli}}]{poole2016exponential}%
  \BibitemOpen
  \bibfield  {author} {\bibinfo {author} {\bibfnamefont {B.}~\bibnamefont
  {Poole}}, \bibinfo {author} {\bibfnamefont {S.}~\bibnamefont {Lahiri}},
  \bibinfo {author} {\bibfnamefont {M.}~\bibnamefont {Raghu}}, \bibinfo
  {author} {\bibfnamefont {J.}~\bibnamefont {Sohl-Dickstein}},\ and\ \bibinfo
  {author} {\bibfnamefont {S.}~\bibnamefont {Ganguli}},\ }\bibfield  {title}
  {\bibinfo {title} {Exponential expressivity in deep neural networks through
  transient chaos},\ }\href@noop {} {\bibfield  {journal} {\bibinfo  {journal}
  {Advances in neural information processing systems}\ }\textbf {\bibinfo
  {volume} {29}} (\bibinfo {year} {2016})}\BibitemShut {NoStop}%
\bibitem [{\citenamefont {Mingo}\ and\ \citenamefont
  {Speicher}(2017)}]{mingo2017free}%
  \BibitemOpen
  \bibfield  {author} {\bibinfo {author} {\bibfnamefont {J.~A.}\ \bibnamefont
  {Mingo}}\ and\ \bibinfo {author} {\bibfnamefont {R.}~\bibnamefont
  {Speicher}},\ }\href@noop {} {\emph {\bibinfo {title} {Free probability and
  random matrices}}},\ Vol.~\bibinfo {volume} {35}\ (\bibinfo  {publisher}
  {Springer},\ \bibinfo {year} {2017})\BibitemShut {NoStop}%
\bibitem [{\citenamefont {Painsky}\ and\ \citenamefont
  {Tishby}(2017)}]{painsky2017gaussian}%
  \BibitemOpen
  \bibfield  {author} {\bibinfo {author} {\bibfnamefont {A.}~\bibnamefont
  {Painsky}}\ and\ \bibinfo {author} {\bibfnamefont {N.}~\bibnamefont
  {Tishby}},\ }\bibfield  {title} {\bibinfo {title} {Gaussian lower bound for
  the information bottleneck limit.},\ }\href@noop {} {\bibfield  {journal}
  {\bibinfo  {journal} {J. Mach. Learn. Res.}\ }\textbf {\bibinfo {volume}
  {18}},\ \bibinfo {pages} {213} (\bibinfo {year} {2017})}\BibitemShut
  {NoStop}%
\bibitem [{\citenamefont {Chechik}\ \emph {et~al.}(2003)\citenamefont
  {Chechik}, \citenamefont {Globerson}, \citenamefont {Tishby},\ and\
  \citenamefont {Weiss}}]{chechik2003information}%
  \BibitemOpen
  \bibfield  {author} {\bibinfo {author} {\bibfnamefont {G.}~\bibnamefont
  {Chechik}}, \bibinfo {author} {\bibfnamefont {A.}~\bibnamefont {Globerson}},
  \bibinfo {author} {\bibfnamefont {N.}~\bibnamefont {Tishby}},\ and\ \bibinfo
  {author} {\bibfnamefont {Y.}~\bibnamefont {Weiss}},\ }\bibfield  {title}
  {\bibinfo {title} {Information bottleneck for gaussian variables},\
  }\href@noop {} {\bibfield  {journal} {\bibinfo  {journal} {Advances in Neural
  Information Processing Systems}\ }\textbf {\bibinfo {volume} {16}} (\bibinfo
  {year} {2003})}\BibitemShut {NoStop}%
\bibitem [{\citenamefont {Ughi}(2022)}]{ughi2022studies}%
  \BibitemOpen
  \bibfield  {author} {\bibinfo {author} {\bibfnamefont {G.}~\bibnamefont
  {Ughi}},\ }\emph {\bibinfo {title} {Studies on neural networks: Information
  propagation at initialisation and robustness to adversarial examples}},\
  \href@noop {} {Ph.D. thesis},\ \bibinfo  {school} {University of Oxford}
  (\bibinfo {year} {2022})\BibitemShut {NoStop}%
\bibitem [{\citenamefont {Saxe}\ \emph {et~al.}(2013)\citenamefont {Saxe},
  \citenamefont {McClelland},\ and\ \citenamefont {Ganguli}}]{saxe2013exact}%
  \BibitemOpen
  \bibfield  {author} {\bibinfo {author} {\bibfnamefont {A.~M.}\ \bibnamefont
  {Saxe}}, \bibinfo {author} {\bibfnamefont {J.~L.}\ \bibnamefont
  {McClelland}},\ and\ \bibinfo {author} {\bibfnamefont {S.}~\bibnamefont
  {Ganguli}},\ }\bibfield  {title} {\bibinfo {title} {Exact solutions to the
  nonlinear dynamics of learning in deep linear neural networks},\ }\href@noop
  {} {\bibfield  {journal} {\bibinfo  {journal} {arXiv preprint
  arXiv:1312.6120}\ } (\bibinfo {year} {2013})}\BibitemShut {NoStop}%
\bibitem [{\citenamefont {Speicher}(1994)}]{speicher1994multiplicative}%
  \BibitemOpen
  \bibfield  {author} {\bibinfo {author} {\bibfnamefont {R.}~\bibnamefont
  {Speicher}},\ }\bibfield  {title} {\bibinfo {title} {Multiplicative functions
  on the lattice of non-crossing partitions and free convolution},\ }\href@noop
  {} {\bibfield  {journal} {\bibinfo  {journal} {Mathematische Annalen}\
  }\textbf {\bibinfo {volume} {298}},\ \bibinfo {pages} {611} (\bibinfo {year}
  {1994})}\BibitemShut {NoStop}%
\bibitem [{\citenamefont {Voiculescu}\ \emph {et~al.}(1992)\citenamefont
  {Voiculescu}, \citenamefont {Dykema},\ and\ \citenamefont
  {Nica}}]{voiculescu1992free}%
  \BibitemOpen
  \bibfield  {author} {\bibinfo {author} {\bibfnamefont {D.~V.}\ \bibnamefont
  {Voiculescu}}, \bibinfo {author} {\bibfnamefont {K.~J.}\ \bibnamefont
  {Dykema}},\ and\ \bibinfo {author} {\bibfnamefont {A.}~\bibnamefont {Nica}},\
  }\href@noop {} {\emph {\bibinfo {title} {Free random variables}}},\ \bibinfo
  {number} {1}\ (\bibinfo  {publisher} {American Mathematical},\ \bibinfo
  {year} {1992})\BibitemShut {NoStop}%
\bibitem [{\citenamefont {Gilboa}\ \emph {et~al.}(2019)\citenamefont {Gilboa},
  \citenamefont {Chang}, \citenamefont {Chen}, \citenamefont {Yang},
  \citenamefont {Schoenholz}, \citenamefont {Chi},\ and\ \citenamefont
  {Pennington}}]{gilboa2019dynamical}%
  \BibitemOpen
  \bibfield  {author} {\bibinfo {author} {\bibfnamefont {D.}~\bibnamefont
  {Gilboa}}, \bibinfo {author} {\bibfnamefont {B.}~\bibnamefont {Chang}},
  \bibinfo {author} {\bibfnamefont {M.}~\bibnamefont {Chen}}, \bibinfo {author}
  {\bibfnamefont {G.}~\bibnamefont {Yang}}, \bibinfo {author} {\bibfnamefont
  {S.~S.}\ \bibnamefont {Schoenholz}}, \bibinfo {author} {\bibfnamefont
  {E.~H.}\ \bibnamefont {Chi}},\ and\ \bibinfo {author} {\bibfnamefont
  {J.}~\bibnamefont {Pennington}},\ }\bibfield  {title} {\bibinfo {title}
  {Dynamical isometry and a mean field theory of lstms and grus},\ }\href@noop
  {} {\bibfield  {journal} {\bibinfo  {journal} {arXiv preprint
  arXiv:1901.08987}\ } (\bibinfo {year} {2019})}\BibitemShut {NoStop}%
\bibitem [{\citenamefont {Stein}\ and\ \citenamefont
  {Shakarchi}(2009)}]{stein2009real}%
  \BibitemOpen
  \bibfield  {author} {\bibinfo {author} {\bibfnamefont {E.~M.}\ \bibnamefont
  {Stein}}\ and\ \bibinfo {author} {\bibfnamefont {R.}~\bibnamefont
  {Shakarchi}},\ }\href@noop {} {\emph {\bibinfo {title} {Real analysis:
  measure theory, integration, and Hilbert spaces}}}\ (\bibinfo  {publisher}
  {Princeton University Press},\ \bibinfo {year} {2009})\BibitemShut {NoStop}%
\end{thebibliography}%
\end{document}